\documentclass[final,1p,times,numbers,sort&compress]{elsarticle}

\usepackage{graphicx}
\usepackage{mathbbold}
\usepackage{subfigure}
\usepackage{amsmath}
\usepackage{amssymb}
\usepackage{chemsym}
\usepackage{slashbox}
\usepackage{algorithm}
\usepackage{algorithmic}

\newtheorem{theorem}{Theorem}[section]

\newtheorem{lemma}[theorem]{Lemma}
\newtheorem{corollary}[theorem]{Corollary}
\newproof{pf}{Proof}

\journal{}

\begin{document}

\begin{frontmatter}



\title{Fast Linearized Alternating Direction Minimization Algorithm with Adaptive Parameter Selection for Multiplicative Noise Removal}


\author[dqc]{Dai-Qiang Chen\corref{cor1}}

\address[dqc]{Department of Mathematics, Third Military Medical University, Chongqing 400038,
Chongqing, People's Republic of China}
\cortext[cor1]{Corresponding author, telephone number: $86-13628484021$}
\ead{chener050@sina.com}

\author[lic]{Li-Zhi Cheng}
\address[lic]{Department of Mathematics and System, School of Sciences,
National University of Defense Technology, Changsha 410073, Hunan,
People's Republic of China}

\begin{abstract}
Owing to the edge preserving ability and low computational cost of the total variation (TV), variational models with the TV regularization have been widely investigated in the field of multiplicative noise removal. The key points of the successful application of these models lie in: the optimal selection of the regularization parameter which balances the data-fidelity term with the TV regularizer; the efficient algorithm to compute the solution. In this paper, we propose two fast algorithms based on the linearized technique, which are able to estimate the regularization parameter and recover the image simultaneously. In the iteration step of the proposed algorithms, the regularization parameter is adjusted by a special discrepancy function defined for multiplicative noise. The convergence properties of the proposed algorithms are proved under certain conditions, and numerical experiments demonstrate that the proposed algorithms overall outperform some state-of-the-art methods in the PSNR values and computational time.
\end{abstract}


\end{frontmatter}

\section{Introduction}\label{sec1}

Images contaminated by the additive Gaussian noise are widely investigated in the field of image restoration. However, for coherent imaging systems such as laser or ultrasound imaging, synthetic aperture radar (SAR) and optical coherence tomography, image acquisition processes are different from the usual optical imaging, and thus the multiplicative noise (speckle), rather than the additive noise, provides an appropriate description of these imaging systems. In this paper, we mainly consider the problem of the multiplicative noise removal.

Let $f\in \mathbb{R}^{m\times n}$ be the observed image corrupted by the multiplicative noise. Our goal is to recover
the original image $u\in \mathbb{R}^{m\times n}$ from the observed image. The corresponding relationship can be expressed as follows:
\begin{equation}\label{equ1.1}
f = u\cdot \eta,
\end{equation}
where $\eta$ is the multiplicative noise that follows the Gamma distribution, i.e.,
\begin{equation}\label{equ1.2}
P_\eta(x)=\frac{M^M}{\Gamma(M)}x^{M-1}e^{-Mx}1_{\{x\geq0\}}.
\end{equation}

In the last several years, various variational models and iterative algorithms \cite{SIAMJAM:AA, SIAMImage:HNW, IEEETIP:MIDAL, JMIV:ST, Submitted:DZ, IEEETIP:Shift, IEEETIP:STV} based on the TV regularization have been proposed for the multiplicative noise removal. Aubert and Aujol \cite{SIAMJAM:AA} used a maximum a posteriori (MAP) estimator and established the following TV regularized variational model:
\begin{equation}\label{equ1.3}
\min_{u}\tau \langle \log u+ fu^{-1}, \textbf{1}\rangle + \|\nabla u\|_{1} ,
\end{equation}
where $\tau>0$ is a fixed regularization parameter, and $\|\nabla u\|_{1}$ is the (isotropic) total variation of $u$, i.e.,
\[
\|\nabla u\|_{1} = \sum_{i=1}^{m}\sum_{j=1}^{n}\sqrt{\left(\nabla_{ij}^{1}u\right)^{2}+\left(\nabla_{ij}^{2}u\right)^{2}}.
\]
Note that $\nabla_{ij}^{1}u$ and $\nabla_{ij}^{2}u$ denote the horizontal and vertical first order differences at pixel $(i,j)$ respectively, $\textbf{1}$ denotes a matrix of ones with the same size as $u$, the multiplication and division are performed in componentwise, and Neumann boundary conditions are used for the computation of the gradient operator $\nabla$ and its adjoint (just the negative divergence operator) $-\textrm{div}$; see \cite{IEEETIP:STV, JSC:TVStokes} for more details.

The objective function in the AA model (\ref{equ1.3}) is nonconvex, and hence it is difficult to find a global optimal solution. Recently, in many existing literatures \cite{SIAMImage:HNW, IEEETIP:MIDAL}, the log transformation was used to resolve the non-convexity. The variational model based on the log transformed image can be reformulated as follows:
\begin{equation}\label{equ1.4}
\min_{u}\tau \langle u+ fe^{-u}, \textbf{1}\rangle + \|\nabla u\|_{1}.
\end{equation}
The above model overcomes the drawback of the AA model, and numerical experiments verify its efficiency \cite{SIAMImage:HNW}. In \cite{JMIV:ST}, the exponential of the solution of the exponential model (\ref{equ1.4}) is proved to be equal to the solution of the classical I-divergence model
\begin{equation}\label{equ1.5}
\min_{u}\tau \langle u- f\log u, \textbf{1}\rangle + \|\nabla u\|_{1}.
\end{equation}
Very recently, new convex methods such as shifting technique \cite{IEEETIP:Shift} and m-th root transformation \cite{IEEETIP:MRoot} were also investigated for this problem.

The augmented Lagrangian framework has recently been proposed to solve the exponential model (\ref{equ1.4}) and the I-divergence model (\ref{equ1.5}). Although this iterative technique is useful, inner iteration \cite{IEEETIP:MIDAL} or inverses involving the Laplacian operator \cite{JMIV:ST} are required at each iteration. The computational cost of the inner iteration or the inverse operation is still expensive. Very recently, linearized techniques \cite{IEEETMI:Linearized, COA:Linearized, PR:Linearized, CAM:Subspace} were widely used for accelerating the alternating minimization algorithm for solving the variational model in image processing. In \cite{SIAMJSC:Linearized}, a fast proximal linearized alternating direction (PLAD) algorithm, which linearized both the fidelity term and the quadratic term of the augmented Lagrangian function, were proposed to solve the TV models (\ref{equ1.4}) and (\ref{equ1.5}). Numerical results there demonstrate that the PLAD algorithm overall outperforms the augmented Lagrangian method for multiplicative noise removal.

The regularization parameter $\tau$ in (\ref{equ1.4}) controls the trade-off between the goodness-of-fit of $f$ and a smoothness requirement due to the TV regularization. The regularized solution highly depends on the selection of $\tau$. Specifically, large $\tau$ leads to little smoothing, and thus, noise will remain almost unchanged in the denoised image or the regularized solution, whereas small $\tau$ leads to oversmoothing solution, so that fine details in the image are destroyed. Therefore, choosing a suitable $\tau$ is a key issue for the variational model (\ref{equ1.4}). In the existing literatures, the main approaches for automatic parameter setting include the generalized cross validation \cite{Technometrics:GCV}, the L-curve \cite{SIAMJSC:LCurve, CAM:LCurve}, the Stein unbiased risk estimator \cite{Statistics:SURE} and the discrepancy principle \cite{NL:DP}. These methods are all based on the assumption of Gaussian noise and not suitable for the special denoised problem here. In \cite{IEEETIP:STV}, a new discrepancy principle based on the statistical characteristics of the multiplicative noise was proposed for selecting a proper value of $\tau$. However, it needs to solve (\ref{equ1.4}) or the corresponding I-divergence model several times for a sequence of $\tau$'s. Hence the computational cost is expensive.

In this paper, we use the special discrepancy principle for the multiplicative noise to compute $\tau$. A linearized alternating direction minimization algorithm with auto-adjusting of the regularization parameter is proposed to solve (\ref{equ1.4}). In each iteration of the proposed algorithm, the regularization parameter is updated in order to guarantee that the denoised image satisfies the discrepancy principle. Due to the use of the linearization technique,
the solution of the subproblem in the iteration step has a closed-form expression, and therefore the zero point of the discrepancy function with respect to $\tau$ can be computed by the Newton method efficiently. Numerical experiments show that our algorithm is very effective in finding a proper $\tau$. Moreover, the proposed algorithm can be extended to the case of the spatially adapted regularization parameter without extra computational burden almost.

We will give the convergence proof of the proposed algorithms with adaptive parameter estimation. With a fixed regularization parameter, our algorithms are reduced to the PLAD algorithm proposed in \cite{SIAMJSC:Linearized}. As we known, the convergence of the original PLAD method is unsolved at present. Therefore, one contribution of our work is to supply a complete convergence proof for the PLAD method.

The rest of this paper is organized as follows. In section \ref{sec2} we briefly review the PLAD algorithms proposed in \cite{SIAMJSC:Linearized}. In section \ref{sec3}, we utilize the statistical characteristic of the Gamma noise to define a special discrepancy function, and then propose a new linearized alternating direction minimization algorithm, which automatically estimate the value of the regularization parameter by computing the zero point of the corresponding discrepancy function. Finally the convergence properties of the proposed iterative algorithm are proved. In section \ref{sec4} we extend the proposed algorithm to the case of the spatially-adaptive regularization parameter. In section \ref{sec5} the numerical results are reported to compare the proposed algorithms with some of the recent state-of-the-art methods.

\section{The proximal linearized alternating direction method}\label{sec2}
\setcounter{equation}{0}

In this section, we briefly review the PLAD algorithm proposed in \cite{SIAMJSC:Linearized}, which will serve as the foundation for the algorithm presented in the next section. Consider the following TV regularized minimization problem:
\begin{equation}\label{equ2.1}
\min_{u} H(u)+ \|\nabla u\|_{1},
\end{equation}
where $H$ is a real-valued, convex and smooth function. The corresponding constrained optimization problem can be reformulated as:
\begin{equation}\label{equ2.2}
\min \limits_{u, z} \{H(u)+ \|z\|_{1} | \nabla u = z\}.
\end{equation}

The augmented Lagrangian function for (\ref{equ2.2}) is defined as
\begin{equation}\label{equ2.3}
\mathcal{L}_{\rho}(u,z,b) = H(u)+ \|z\|_{1} + \langle b, z-\nabla u\rangle + \frac{\rho}{2}\|z-\nabla u\|_{2}^{2}.
\end{equation}

Then the well-known ADMM (alternating direction method of multipliers) for solving (\ref{equ2.1}) can be expressed as follows:
\begin{equation}\label{equ2.4}
\left\{\begin{array}{lll}u^{k+1} = \textrm{arg}\min \limits_{u}\left\{H(u)+\langle b^{k}, z^{k}-\nabla u\rangle + \frac{\rho}{2}\|z^{k}-\nabla u\|_{2}^{2} \right\}, &~ &~ ~\\
z^{k+1} = \textrm{arg}\min \limits_{z}\left\{\|z\|_{1}+ \langle b^{k}, z-\nabla u^{k+1}\rangle + \frac{\rho}{2}\|z-\nabla u^{k+1}\|_{2}^{2} \right\},
&~ &~ ~\\ b^{k+1}=b^{k} + \rho(z^{k+1} - \nabla u^{k+1}).& ~
&~
\end{array} \right.
\end{equation}
The first subproblem in (\ref{equ2.4}) is difficult to solve. Therefore, the authors in \cite{SIAMJSC:Linearized} used the second-order Taylor expansion of $\mathcal{F}^{z^{k}}(u) = H(u) + \frac{\rho}{2}\|z^{k}-\nabla u\|_{2}^{2}$ at $u^{k}$, and meanwhile replaced the Hessian matrix $\nabla^{2}_{u}\mathcal{F}^{z^{k}}(u)$ with $\frac{1}{\delta}\textbf{I}$ ($\delta>0$ is a constant). Then the first subproblem in (\ref{equ2.4}) can be simplified as follows:
\begin{equation}\label{equ2.5}
u^{k+1} = \textrm{arg}\min_{u} \left\{\langle \nabla H(u^{k}) + \rho \textrm{div}(z^{k}-\nabla u^{k}), u-u^{k}\rangle + \langle b^{k}, z^{k}-\nabla u\rangle + \frac{1}{2\delta}\|u - u^{k}\|_{2}^{2} \right\}.
\end{equation}
Replacing the first subproblem in (\ref{equ2.4}) with (\ref{equ2.5}) we obtain the PLAD algorithm for the minimization problem (\ref{equ2.1}).

Now consider the variational models (\ref{equ1.4}) and (\ref{equ1.5}). They can be reformulated as
\begin{equation}\label{equ2.6}
\min_{u} H(u) + \lambda \|\nabla u\|_{1},
\end{equation}
where $\lambda = \frac{1}{\tau}$, and $H(u)= \langle u+ fe^{-u}, \textbf{1}\rangle$ or $\langle u- f\log u, \textbf{1}\rangle$. While the PLAD algorithm is applied to solve the minimization problem in (\ref{equ2.6}), it can be expressed as the following simple form:
\begin{equation}\label{equ2.7}
\left\{\begin{array}{lll}u^{k+1} = u^{k}-\delta \left(1-\frac{f}{\varphi^{-1}(u^{k})}+\rho \textrm{div}(z^{k}-\nabla u^{k})+\textrm{div} b^{k}\right), &~ &~ ~\\
z^{k+1} = shrink\left(\nabla u^{k+1} - \frac{b^{k}}{\rho},\frac{\lambda}{\rho}\right),
&~ &~ ~\\ b^{k+1}=b^{k} + \rho(z^{k+1} - \nabla u^{k+1}),& ~
&~
\end{array} \right.
\end{equation}
where $\varphi(u)=\log u$ for the exponential model and $\varphi(u)=u$ for the I-divergence model. The shrinkage operator is defined as:
\[
z = shrink(u, c) = \max(\|u\|_{2}-c,0)\frac{u}{\|u\|_{2}}.
\]
If we further assume that $u\in U = [c_{\inf}, c_{\sup}]^{m\times n}$, and $0< c_{\inf}<c_{\sup}<+\infty$, then the first formula in (\ref{equ2.7}) should be replaced by
\begin{equation}\label{equ2.8}
u^{k+1} = P_{U}\left(u^{k}-\delta \left(1-\frac{f}{\varphi^{-1}(u^{k})}+\rho \textrm{div}(z^{k}-\nabla u^{k})+\textrm{div} b^{k}\right)\right),
\end{equation}
where $P_{U}$ denotes the projection onto the set $U$.

\section{Linearized alternating minimization algorithm with adaptive parameter selection}\label{sec3}
\setcounter{equation}{0}

In this section, we describe the proposed linearized alternating minimization algorithm with adaptive parameter selection. To this end, we first introduce the statistical characteristic of some random variable with respect to Gamma noise $\eta$. It will play an important role in developing our algorithm.

\begin{lemma}\label{lem1}
Let $\eta$ be a Gamma random variable (r.v) with mean 1 and standard deviation $\frac{1}{\sqrt{M}}$. Consider the following discrepancy function with respect to $\eta$
\begin{equation}\label{equ3.1}
I(\eta)=\eta-\log \eta.
\end{equation}
Then the following estimate of the expected value of $I(\eta)$ holds
true for large $M$:
\begin{equation}\label{equ3.2}
E\{I(\eta)\}=1+\frac{1}{2M}+\frac{1}{12M^{2}}-\frac{5}{2M^{3}}+
O\left(\frac{1}{M^{3}}\right).
\end{equation}
\end{lemma}
The above conclusion has appeared in the existing literature \cite{IEEETIP:STV}, and therefore we omit the proof here.

Next, we consider the PLAD algorithm for solving the exponential model (\ref{equ1.4}). Choose $H(u)=\tau \langle u+ fe^{-u}, \textbf{1}\rangle$. The solution $u^{k+1}$ in (\ref{equ2.5}) has a closed expression as follows:
\begin{equation}\label{equ3.3}
u^{k+1} = u^{k} - \delta\left(\tau(\textbf{1}-fe^{-u^{k}})+\rho \textrm{div}(z^{k}-\nabla u^{k})+ \textrm{div} b^{k}\right),
\end{equation}
which can be seen as a linear function with respect to $\tau$. Since $u^{k+1}$ is an approximation of the log transformed original image, we have
\begin{equation}\label{equ3.4}
u^{k+1} + fe^{-u^{k+1}} -\log f = \frac{f}{e^{u^{k+1}}} - \log \frac{f}{e^{u^{k+1}}} \approx \eta - \log \eta.
\end{equation}
Denote
\[
A_{1} = - \delta(\textbf{1}-fe^{-u^{k}}),
\]
\[
A_{2} = u^{k} - \delta \left(\rho \textrm{div}(z^{k}-\nabla u^{k})+ \textrm{div} b^{k}\right).
\]
Then $u^{k+1} = A_{1} \tau + A_{2}$. Let $C=1+\frac{1}{2M}+\frac{1}{12M^{2}}-\frac{5}{2M^{3}}$. Then according to Lemma \ref{lem1} and the relation (\ref{equ3.4}), for large $M$ we have
\begin{equation}\label{equ3.5}
\frac{1}{mn}\sum_{i=1}^{m}\sum_{j=1}^{n}\left(A_{1} \tau + A_{2} + f e^{-(A_{1} \tau + A_{2})}- \log f\right)_{i,j} - C \approx 0.
\end{equation}

On the other hand, considering the function $q(x) = x + f_{ij} e^{-x} - \log f_{ij}$, it is monotonically increasing for $x>\log f_{ij}$ and decreasing for $x<\log f_{ij}$. Besides, $q(\log f_{ij}) = 0$. The denoised image $u^{k+1}$ is a smoothed version of the noisy image $\log f$, and its value will be far from $\log f$ under the over-smoothness condition (with small $\tau$). Therefore, we demand that $u^{k+1}$ obtained by (\ref{equ3.3}) satisfies the following inequality in order to avoid over-smoothness:
\begin{equation}\label{equ3.6}
\mathcal{K}(\tau; w^{k}) = \frac{1}{mn}\sum_{i=1}^{m}\sum_{j=1}^{n}\left(A_{1} \tau + A_{2} + f e^{-(A_{1} \tau + A_{2})}- \log f\right)_{i,j} - \bar{C} \leq 0,
\end{equation}
where $w^{k}=(u^{k}; z^{k}; b^{k})$ and $\bar{C}>0$ is a constant for each iteration. According to (\ref{equ3.5}) we infer that $\bar{C} = C$ is proper for large $M$. In fact, we use this setting for most experiments in section \ref{sec5}.
The function $\mathcal{K}$ in (\ref{equ3.6}) is called as the discrepancy function defined for Gamma noise.

The relation in (\ref{equ3.6}) inspires us to update the value of $\tau$ in the iteration step $k$ as follows:
if $\mathcal{K}(\tau^{k}; w^{k}) \leq 0$, then $\tau^{k+1} = \tau^{k}$. Otherwise update $\tau^{k+1}$ just as
the zero point of the function $\mathcal{K}(\tau; w^{k})$, i.e., $\mathcal{K}(\tau^{k+1}; w^{k})=0$.
The following lemma guarantees the existence of the solution of $\mathcal{K}(\tau; w^{k})=0$.
\begin{lemma}\label{lem2}
Assume that $\sum_{i=1}^{m}\sum_{j=1}^{n}(A_{1})_{i,j}\neq 0$, $f>0$ and $\bar{C}\geq \frac{1}{mn}\sum_{i=1}^{m}\sum_{j=1}^{n}\left(A_{2} + f e^{-A_{2}}- \log f\right)_{i,j}$, then the equation $\mathcal{K}(\tau; w^{k})=0$ has at least one solution.
\end{lemma}

\begin{pf}
Obviously, $\mathcal{K}(\tau; w^{k})$ is continuous with respect to $\tau$. If $\sum_{i=1}^{m}\sum_{j=1}^{n}(A_{1})_{i,j}> 0$, then $\lim \limits_{\tau \rightarrow +\infty} \sum_{i=1}^{m}\sum_{j=1}^{n}(A_{1})_{i,j} \tau = +\infty$. Hence there exists $\tau_{1}>0$ such that $\mathcal{K}(\tau; w^{k})>0$ for any $\tau\geq \tau_{1}$. If $\sum_{i=1}^{m}\sum_{j=1}^{n}(A_{1})_{i,j}< 0$, then there exists $(A_{1})_{i_{0},j_{0}}<0$ for some $(i_{0},j_{0})$, and thus $\lim \limits_{\tau \rightarrow +\infty} \sum_{i=1}^{m}\sum_{j=1}^{n}(A_{1})_{i,j} \tau + f e^{-(A_{1})_{i_{0},j_{0}} \tau} = +\infty$. This shows that there exists $\tau_{2}>0$ such that $\mathcal{K}(\tau; w^{k})>0$
for any $\tau\geq \tau_{2}$. Moreover, due to $\bar{C}\geq \frac{1}{mn}\sum_{i=1}^{m}\sum_{j=1}^{n}\left(A_{2} + f e^{-A_{2}}- \log f\right)_{i,j}$, there exists $\tau_{3}\leq 0$
satisfies $\mathcal{K}(\tau_{3}; w^{k})>0$. Therefore, the equation $\mathcal{K}(\tau; w^{k})=0$ has at least one solution.


\end{pf}

Based on the above analysis, we obtain the linearized alternating direction minimization algorithm with adaptive parameter selection, which is summarized in Algorithm 1.

\begin{algorithm}[htb]
\caption{ Discrepancy Principle Based Linearized Alternating Direction Minimization Algorithm for Multiplicative Noise Removal (DP-LADM)}
\begin{algorithmic}[1]
\REQUIRE  noisy image $f$; choose $\rho >0$, $\delta >0$.\\
\textbf{Output}: denoised log transformed image $u$; regularization parameter $\tau$. \\
\textbf{Initialization}: $k=0; b^{0}=0; u^{0}=\log f; z^{0} = \nabla u^{0}$; $\tau^{0} = \tau_{0}$.\\
\textbf{Iteration}: \\
~~~if $\mathcal{K}(\tau^{k}; w^{k}) \leq 0$; then \\
~~~~~~$\tau^{k+1} = \tau^{k}$; \\
~~~else \\
~~~~~~compute the solution of $\mathcal{K}(\tau; w^{k})=0$ by the Newton iteration; \\
~~~end if \\
~~~$u^{k+1} = u^{k} - \delta\left(\tau^{k+1}(\textbf{1}-fe^{-u^{k}})+\rho \textrm{div}(z^{k}-\nabla u^{k})+ \textrm{div} b^{k}\right)$; \\
~~~$z^{k+1} = \textrm{arg}\min \limits_{z}\left\{\|z\|_{1}+ \langle b^{k}, z-\nabla u^{k+1}\rangle + \frac{\rho}{2}\|z-\nabla u^{k+1}\|_{2}^{2} \right\}$; \\
~~~$b^{k+1}=b^{k} + \rho(z^{k+1} - \nabla u^{k+1})$; \\
~~~$k = k+1$; \\
until some stopping criterion is satisfied. \\
return $u = u^{k+1}$ and $\tau = \tau^{k+1}$.
\end{algorithmic}
\end{algorithm}

\subsection{Convergence analysis of the proposed algorithm}\label{subsec3.1}

Here we investigate the convergence properties of the sequence $(u^{k}, z^{k}, b^{k}, \tau^{k})$ generated by Algorithm 1. Assume that $\bar{\tau} = \sup_{k} \tau^{k}< +\infty$, and $(\bar{u}, \bar{z})$ is one solution of the variational model
\begin{equation}\label{equ3.7}
\min_{u, z}\bar{\tau} \langle u+ fe^{-u}, \textbf{1}\rangle + \|z\|_{1} ~~~ s.t. ~~ \nabla u = z.
\end{equation}
Choose $\bar{C} = \frac{1}{mn}\langle \bar{u}+ fe^{-\bar{u}}-\log f, \textbf{1}\rangle$ in Algorithm 1. Denote $D(u) = \langle u+ fe^{-u}, \textbf{1}\rangle$, and $L_{D}$ as a Lipschitz constant that satisfies
\[
\|\nabla D(u_{1}) - \nabla D(u_{2})\|\leq L_{D}\|u_{1} - u_{2}\|
\]
for any $u_{1}, u_{2}$ ($\|\cdot\|$ denotes $\|\cdot\|_{2}$ unless otherwise specified in the following). Then we have the following convergence result.

\begin{theorem}\label{the1}
Let $(u^{k}, z^{k}, b^{k})$ be the sequence generated by Algorithm 1 with $\delta < \frac{1}{\bar{\tau} L_{D} + \rho \|\triangle\|_{2}}$. Then it converges to a point $(\bar{u}, \bar{z}, \bar{b})$ where the first-order optimality conditions for (\ref{equ3.7}) are satisfied.
\end{theorem}

\begin{pf}
The first-order optimality conditions for the sequence $(u^{k}, z^{k}, b^{k}, \tau^{k})$ generated by Algorithm 1 are
\begin{equation}\label{equ3.8}
\left\{\begin{array}{lll}0 = \tau^{k+1} \nabla D(u^{k})+\frac{1}{\delta}(u^{k+1}-u^{k})+\rho \textrm{div}(z^{k}-\nabla u^{k}) + \textrm{div} b^{k}, &~ &~ ~\\
0=t^{k+1}+\rho(z^{k+1}-\nabla u^{k+1}+\rho^{-1}b^{k}),
&~ &~ ~\\ b^{k+1}=b^{k} + \rho(z^{k+1} - \nabla u^{k+1}),& ~ &~
\end{array} \right.
\end{equation}
where $t^{k+1} \in \partial \|z^{k+1}\|_{1}$. Since $(\bar{u}, \bar{z})$ is one solution of (\ref{equ3.7}), the first-order optimality conditions can be rewritten as
\begin{equation}\label{equ3.9}
0=\bar{t} + \bar{b},~~~ 0=\bar{\tau} \nabla D(\bar{u})+\textrm{div}\bar{b}, ~~~0=\nabla \bar{u}-\bar{z}
\end{equation}
for some $\bar{t} \in \partial \|\bar{z}\|_{1}$. Rearrange (\ref{equ3.9}) to get
\begin{equation}\label{equ3.10}
\left\{\begin{array}{lll}0 = \bar{\tau} \nabla D(\bar{u})+\frac{1}{\delta}(\bar{u}-\bar{u})+\rho \textrm{div}(\bar{z}-\nabla \bar{u}) + \textrm{div} \bar{b}, &~ &~ ~\\
0=\bar{t}+\rho(\bar{z}-\nabla \bar{u}+\rho^{-1}\bar{b}),
&~ &~ ~\\ \bar{b}=\bar{b} + \rho(\bar{z}-\nabla \bar{u}).& ~
&~
\end{array} \right.
\end{equation}

Denote the errors by $u^{k+1}_{e} = u^{k+1}-\bar{u}$, $z^{k+1}_{e} = z^{k+1}-\bar{z}$, $b^{k+1}_{e} = b^{k+1}-\bar{b}$ and $t^{k+1}_{e} = t^{k+1}-\bar{t}$. Subtracting (\ref{equ3.10}) from (\ref{equ3.8}) we obtain
\begin{equation}\label{equ3.11}
\left\{\begin{array}{lll}0 = \tau^{k+1} \nabla D(u^{k}) - \bar{\tau} \nabla D(\bar{u}) +\frac{1}{\delta}(u^{k+1}_{e}-u^{k}_{e})+\rho \textrm{div}(z^{k}_{e}-\nabla u^{k}_{e}) + \textrm{div} b^{k}_{e}, &~ &~ ~\\
0=t^{k+1}_{e}+\rho(z^{k+1}_{e}-\nabla u^{k+1}_{e}+\rho^{-1}b^{k}_{e}),
&~ &~ ~\\ b^{k+1}_{e}=b^{k}_{e} + \rho(z^{k+1}_{e} - \nabla u^{k+1}_{e}).& ~ &~
\end{array} \right.
\end{equation}
Taking the inner product with $u^{k+1}_{e}$, $z^{k+1}_{e}$ and $b^{k}_{e}$ on both sides of the three equations of (\ref{equ3.11}) respectively, we obtain
\begin{equation}\label{equ3.12}
\left\{\begin{array}{lll}0 = \langle \tau^{k+1} \nabla D(u^{k}) - \bar{\tau} \nabla D(\bar{u}), u^{k+1}_{e}\rangle +\frac{1}{\delta}\langle u^{k+1}_{e}-u^{k}_{e}, u^{k+1}_{e}\rangle + \langle \rho \textrm{div}(z^{k}_{e}-\nabla u^{k}_{e}) + \textrm{div} b^{k}_{e}, u^{k+1}_{e}\rangle, &~ &~ ~\\
0=\langle t^{k+1}_{e}, z^{k+1}_{e}\rangle + \rho\langle z^{k+1}_{e}-\nabla u^{k+1}_{e}+\rho^{-1}b^{k}_{e}, z^{k+1}_{e}\rangle,
&~ &~ ~\\ \langle b^{k+1}_{e}, b^{k}_{e}\rangle=\langle b^{k}_{e}, b^{k}_{e}\rangle + \rho\langle z^{k+1}_{e} - \nabla u^{k+1}_{e}, b^{k}_{e}\rangle.& ~ &~
\end{array} \right.
\end{equation}
Due to $\langle x-y, x\rangle = \frac{1}{2}(\|x\|^{2} + \|x-y\|^{2}- \|y\|^{2})$, according to (\ref{equ3.12}) we get
\begin{equation}\label{equ3.13}
\left\{\begin{array}{lll}\langle \tau^{k+1} \nabla D(u^{k}) - \bar{\tau} \nabla D(\bar{u}), u^{k+1}_{e}\rangle +\frac{1}{2\delta}(\|u^{k+1}_{e}\|^{2}+\|u^{k+1}-u^{k}\|^{2}-\|u^{k}_{e}\|^{2})= \rho \langle \nabla u^{k+1}_{e}, z^{k}_{e}+ \rho^{-1} b^{k}_{e} -\nabla u^{k}_{e} \rangle, &~ &~ ~\\
\langle t^{k+1}_{e}, z^{k+1}_{e}\rangle = \langle -b^{k}_{e}, z^{k+1}_{e}\rangle + \rho\langle \nabla u^{k+1}_{e}-z^{k+1}_{e}, z^{k+1}_{e}\rangle ,
&~ &~ ~\\ \frac{1}{2\rho}(\|b^{k+1}_{e}\|^{2}-\|b^{k}_{e}\|^{2})=\frac{\rho}{2}\|\nabla u^{k+1}_{e}-z^{k+1}_{e}\|^{2} + \langle b^{k}_{e}, z^{k+1}_{e}-\nabla u^{k+1}_{e}\rangle,& ~ &~
\end{array} \right.
\end{equation}
The last equation in (\ref{equ3.13}) uses the relation $b^{k+1} - b^{k} = \rho(z^{k+1}-\nabla u^{k+1})$.
Adding the equations in (\ref{equ3.13}) we obtain
\begin{equation}\label{equ3.14}
\begin{split}
\frac{1}{2\delta}(\|u^{k+1}_{e}\|^{2}+\|u^{k+1}-u^{k}\|^{2}-\|u^{k}_{e}\|^{2}) + \langle \tau^{k+1} \nabla D(u^{k}) - \bar{\tau} \nabla D(\bar{u}), u^{k+1}_{e}\rangle + \langle t^{k+1}_{e}, z^{k+1}_{e}\rangle +  \\ \frac{1}{2\rho}(\|b^{k+1}_{e}\|^{2}-\|b^{k}_{e}\|^{2})
= \rho \langle \nabla u^{k+1}_{e}, z^{k}_{e} -\nabla u^{k}_{e} \rangle + \rho\langle \nabla u^{k+1}_{e}-z^{k+1}_{e}, z^{k+1}_{e}\rangle + \frac{\rho}{2}\|\nabla u^{k+1}_{e}-z^{k+1}_{e}\|^{2} \\
= -\frac{\rho}{2}\|\nabla u^{k+1}_{e}-z^{k}_{e}\|^{2} + \frac{\rho}{2}(\|\nabla u^{k+1}_{e}\|^{2} + \|\nabla (u^{k+1}-u^{k})\|^{2} - \|\nabla u^{k}_{e}\|^{2}) - \frac{\rho}{2}(\|z^{k+1}_{e}\|^{2}-\|z^{k}_{e}\|^{2}).
\end{split}
\end{equation}
Besides,
\begin{eqnarray*}
\langle \tau^{k+1} \nabla D(u^{k}) - \bar{\tau} \nabla D(\bar{u}), u^{k+1}_{e}\rangle & = & \tau^{k+1} \langle \nabla D(u^{k}) - \nabla D(\bar{u}), u^{k+1}_{e}\rangle + (\tau^{k+1}-\bar{\tau}) \langle  \nabla D(\bar{u}), u^{k+1}_{e}\rangle  \\
& = & \tau^{k+1} \langle Q u^{k}_{e}, Q u^{k+1}_{e}\rangle + (\tau^{k+1}-\bar{\tau}) \langle  \nabla D(\bar{u}), u^{k+1}_{e}\rangle  \\
& = & \frac{\tau^{k+1}}{2}(\|Q u^{k}_{e}\|^{2}+\|Q u^{k+1}_{e}\|^{2}-\|Q (u^{k+1}-u^{k})\|^{2}) + (\tau^{k+1}-\bar{\tau}) \langle  \nabla D(\bar{u}), u^{k+1}_{e}\rangle,
\end{eqnarray*}
where $\nabla D(u^{k}) - \nabla D(\bar{u}) = \nabla^{2} D(\xi) (u^{k}-\bar{u})$ and $\nabla^{2} D(\xi)=Q^{T}Q$($\nabla^{2} D(\xi)$ is a positive-definite matrix due to $f>0$).

Substituting it in (\ref{equ3.14}) we obtain
\begin{equation}\label{equ3.15}
\begin{split}
\frac{1}{\delta}\|u^{k+1}_{e}\|^{2} - \rho\|\nabla u^{k+1}_{e}\|^{2} +  \rho\|z^{k+1}_{e}\|^{2}+\frac{1}{\rho}\|b^{k+1}_{e}\|^{2} + \tau^{k+1}(\|Q u^{k}_{e}\|^{2}+\|Q u^{k+1}_{e}\|^{2})\\
+\left(\frac{1}{\delta}\|u^{k+1} - u^{k}\|^{2}-\rho \|\nabla (u^{k+1} - u^{k})\|^{2}-\tau^{k+1}\|Q(u^{k+1} - u^{k})\|^{2}\right)\\
+2(\tau^{k+1}-\bar{\tau})\langle \nabla D(\bar{u}), u^{k+1}_{e}\rangle + 2\langle t^{k+1}_{e}, z^{k+1}_{e}\rangle + \rho\|\nabla u^{k+1} - z^{k}\|^{2} \\
= \frac{1}{\delta}\|u^{k}_{e}\|^{2} - \rho\|\nabla u^{k}_{e}\|^{2} + \rho\|z^{k}_{e}\|^{2}+\frac{1}{\rho}\|b^{k}_{e}\|^{2}.
\end{split}
\end{equation}
Since $\delta < \frac{1}{\bar{\tau} L_{D} + \rho \|\triangle\|_{2}}$, we have
\[
\left(\frac{1}{\delta}\|u^{k+1} - u^{k}\|^{2}-\rho \|\nabla (u^{k+1} - u^{k})\|^{2}-\tau^{k+1}\|Q(u^{k+1} - u^{k})\|^{2}\right)\geq 0.
\]
Moreover, by the convexity of $D(u)$ we have $\langle \nabla D(\bar{u}), u^{k+1}_{e}\rangle \leq D(u^{k+1})-D(\bar{u})$. By the update criterion of $\tau^{k+1}$ in Algorithm 1 we infer that  $\mathcal{K}(\tau^{k+1}; w^{k}) \leq 0$, and therefore
\[
\frac{1}{mn}\langle u^{k+1}+ fe^{-u^{k+1}}-\log f, \textbf{1}\rangle \leq \bar{C},
\]
which implies that $D(u^{k+1})-D(\bar{u})\leq 0$ by the definition of $\bar{C}$. Due to $\tau^{k+1}\leq \bar{\tau}$, we infer that $(\tau^{k}-\bar{\tau})\langle \nabla D(\bar{u}), u^{k+1}_{e}\rangle\geq 0$.
By the convexity of $\|z\|_{1}$ we also have $\langle t^{k+1}_{e}, z^{k+1}_{e}\rangle\geq 0$.

After removing the nonnegative terms except the first four terms in the left side of the equation (\ref{equ3.15}), we obtain
\begin{equation}\label{equ3.16}
\frac{1}{\delta}\|u^{k+1}_{e}\|^{2} - \rho\|\nabla u^{k+1}_{e}\|^{2} + \rho\|z^{k+1}_{e}\|^{2}+\frac{1}{\rho}\|b^{k+1}_{e}\|^{2}
\leq \frac{1}{\delta}\|u^{k}_{e}\|^{2} - \rho\|\nabla u^{k}_{e}\|^{2} + \rho\|z^{k}_{e}\|^{2}+\frac{1}{\rho}\|b^{k}_{e}\|^{2}.
\end{equation}
Since $\frac{1}{\delta}>\rho \|\triangle\|_{2}$, we know that $\frac{1}{\delta}\|u^{k}_{e}\|^{2} - \rho\|\nabla u^{k}_{e}\|^{2}\geq 0$ for any $k$. According to (\ref{equ3.16}) we conclude that $(u^{k}_{e}, z^{k}_{e}, b^{k}_{e})$ is bounded and hence $(u^{k}, z^{k}, b^{k})$ is also bounded.

By summing (\ref{equ3.15}) from some $j_{0}$ to $+\infty$, we can derive

\begin{equation}\label{equ3.17}
\begin{split}
\sum_{k=j_{0}}^{+\infty}\left(\frac{1}{\delta}\|u^{k+1} - u^{k}\|^{2}-\rho \|\nabla (u^{k+1} - u^{k})\|^{2}-\tau^{k+1}\|Q(u^{k+1} - u^{k})\|^{2}\right)\\
+\sum_{k=j_{0}}^{+\infty}\left(2(\tau^{k+1}-\bar{\tau})\langle \nabla D(\bar{u}), u^{k+1}_{e}\rangle + 2\langle t^{k+1}_{e}, z^{k+1}_{e}\rangle + \rho\|\nabla u^{k+1} - z^{k}\|^{2}\right) \\
= \frac{1}{\delta}\|u^{j_{0}}_{e}\|^{2} - \rho\|\nabla u^{j_{0}}_{e}\|^{2} + \rho\|z^{j_{0}}_{e}\|^{2}+\frac{1}{\rho}\|b^{j_{0}}_{e}\|^{2},
\end{split}
\end{equation}
and hence we obtain
\begin{equation}\label{equ3.18}
\sum_{k=j_{0}}^{+\infty}\left(\frac{1}{\delta}-\rho \|\triangle\|_{2}-\bar{\tau} L_{D}\right)\|u^{k+1} - u^{k}\|^{2}+\sum_{k=j_{0}}^{+\infty}\rho\|\nabla u^{k+1} - z^{k}\|^{2}<+\infty.
\end{equation}
This implies that
\begin{equation}\label{equ3.19}
\lim_{k\rightarrow +\infty}\|u^{k+1} - u^{k}\|=0, ~~~\lim_{k\rightarrow +\infty}\|\nabla u^{k+1} - z^{k}\|=0.
\end{equation}
Therefore,
\begin{equation}\label{equ3.20}
\|\nabla u^{k} - z^{k}\|\leq \|\nabla (u^{k+1} - u^{k})\| + \|\nabla u^{k+1} - z^{k}\| \rightarrow 0, ~~ \textrm{as} ~~k\rightarrow +\infty.
\end{equation}
According to (\ref{equ3.19}) and (\ref{equ3.20}) we also have
\begin{equation}\label{equ3.21}
\|z^{k+1} - z^{k}\|\leq \|\nabla u^{k+1} - z^{k+1}\| + \|\nabla u^{k+1} - z^{k}\| \rightarrow 0, ~~ \textrm{as} ~~k\rightarrow +\infty.
\end{equation}
Moreover, by the relation of $b^{k+1} - b^{k} = \rho(z^{k+1}-\nabla u^{k+1})$ and (\ref{equ3.20}) we have
\begin{equation}\label{equ3.22}
\lim_{k\rightarrow +\infty}\|b^{k+1} - b^{k}\|=0.
\end{equation}

In (\ref{equ3.16}) we have shown that $(u^{k}, z^{k}, b^{k})$ is bounded. Then there exists a convergent subsequence, also denoted by $(u^{k}, z^{k}, b^{k})$ for convenience, converging to a limit $(u^{\infty}, z^{\infty}, b^{\infty})$. Due to $t^{k}\in \partial \|z^{k}\|_{1}$, it follows that $
\|t^{k}\|\in [-1, 1]$ and hence there exists a subsequence of $(t^{k})$ converging to a limit $t^{\infty}$. Besides, as $\bar{\tau} = \sup_{k} \tau^{k}$, there exists a subsequence of $(\tau^{k+1})$ converging to $\bar{\tau}$.

For convenience, let $(u^{k}, z^{k}, b^{k}, t^{k}, \tau^{k+1})$ be a subsequence converging to $(u^{\infty}, z^{\infty}, b^{\infty}, t^{\infty}, \bar{\tau})$. Since $\|z\|_{1}$ is a closed proper convex function, according to Theorem 24.4 in \cite{Convex} we have $t^{\infty}\in \partial\|z^{\infty}\|_{1}$.

Let $k\rightarrow +\infty$ in (\ref{equ3.8}) while utilizing the convergence results in (\ref{equ3.19})-(\ref{equ3.22}) to obtain

\begin{equation}\label{equ3.23}
\left\{\begin{array}{lll}0 = \bar{\tau} \nabla D(u^{\infty})+\rho \textrm{div}(z^{\infty}-\nabla u^{\infty}) + \textrm{div} b^{\infty}, &~ &~ ~\\
0=t^{\infty}+\rho(z^{\infty}-\nabla u^{\infty}+\rho^{-1}b^{\infty}),
&~ &~ ~\\ 0 = z^{\infty}-\nabla u^{\infty},& ~
&~
\end{array} \right.
\end{equation}
which is equivalent to
\[
0=\bar{\tau} \nabla D(u^{\infty})+\textrm{div}b^{\infty}, ~~~  0=t^{\infty} + b^{\infty}, ~~~0=\nabla u^{\infty}-z^{\infty}.
\]
Hence the limit point $(u^{\infty}, z^{\infty}, b^{\infty})$ satisfies the first-order optimality conditions for
(\ref{equ3.7}).

The proof of this theorem started with any saddle point $(\bar{u}, \bar{z}, \bar{b})$. Here we consider the special case of $\bar{u}=u^{\infty}, \bar{z}=z^{\infty}, \bar{b}=b^{\infty}$ which is the limit of a convergent subsequence $(u^{k^{j}}, z^{k^{j}}, b^{k^{j}})$. According to (\ref{equ3.16}) we know that, for any $k>k^{j}$,
\[
\frac{1}{\delta}\|u^{k}_{e}\|^{2} - \rho\|\nabla u^{k}_{e}\|^{2} + \rho\|z^{k}_{e}\|^{2}+\frac{1}{\rho}\|b^{k}_{e}\|^{2}
\leq \frac{1}{\delta}\|u^{k^{j}}_{e}\|^{2} - \rho\|\nabla u^{k^{j}}_{e}\|^{2} + \rho\|z^{k^{j}}_{e}\|^{2}+\frac{1}{\rho}\|b^{k^{j}}_{e}\|^{2}.
\]
Let $j$ tend to infinity to deduce that
\[
\lim_{k\rightarrow +\infty}u^{k}_{e} = \lim_{k\rightarrow +\infty}z^{k}_{e} = \lim_{k\rightarrow +\infty}b^{k}_{e} =0,
\]
which demonstrates that $(u^{k}, z^{k}, b^{k})$ converges to $(\bar{u}, \bar{z}, \bar{b})$. This completes the proof.
\end{pf}

If we fix $\tau^{k} \equiv \tau_{0}$, then Algorithm 1 reduces to the PLAD method proposed in \cite{SIAMJSC:Linearized}. The proof above can be used for the special case with little modification. Therefore, we can obtain the following convergence result.

\begin{corollary}\label{cor1}
(the convergence property of the PLAD method) Let $(u^{k}, z^{k}, b^{k})$ be the sequence generated by the PLAD method with $\delta < \frac{1}{\tau_{0} L_{D} + \rho \|\triangle\|_{2}}$. Then it converges to a point $(\bar{u}, \bar{z}, \bar{b})$ where the first-order optimality conditions for (\ref{equ3.7}) ($\bar{\tau}=\tau_{0}$ in this case) are satisfied.
\end{corollary}

\begin{pf}
In this proof we consider the constraint $u\in U = [c_{\inf}, c_{\sup}]^{m\times n}$. We can obtain the first-order optimality conditions similar to those in (\ref{equ3.8}) and (\ref{equ3.10}). The differences lie in $l^{k+1}\in \mathcal{N}_{U}(u^{k+1})$ and $\bar{l}\in \mathcal{N}_{U}(\bar{u})$ are added to the first formulas of (\ref{equ3.8}) and (\ref{equ3.10}) respectively, and $\tau^{k} \equiv \tau_{0} \equiv \bar{\tau}$. Here $\mathcal{N}_{U}$ denotes the normal cone \cite{VariationalAnalysis}. Then similarly to (\ref{equ3.14}) we can obtain that
\begin{equation}\label{equ3.26}
\begin{split}
\frac{1}{2\delta}(\|u^{k+1}_{e}\|^{2}+\|u^{k+1}-u^{k}\|^{2}-\|u^{k}_{e}\|^{2}) + \tau_{0} \langle  \nabla D(u^{k}) - \nabla D(\bar{u}), u^{k+1}_{e}\rangle + \\ \langle t^{k+1}_{e}, z^{k+1}_{e}\rangle + \langle l^{k+1}_{e}, u^{k+1}_{e}\rangle +  \frac{1}{2\rho}(\|b^{k+1}_{e}\|^{2}-\|b^{k}_{e}\|^{2})
= -\frac{\rho}{2}\|\nabla u^{k+1}_{e}-z^{k}_{e}\|^{2} + \\ \frac{\rho}{2}(\|\nabla u^{k+1}_{e}\|^{2} + \|\nabla (u^{k+1}-u^{k})\|^{2} - \|\nabla u^{k}_{e}\|^{2}) - \frac{\rho}{2}(\|z^{k+1}_{e}\|^{2}-z^{k}_{e}\|^{2}),
\end{split}
\end{equation}
where $l^{k+1}_{e} = l^{k+1}-\bar{l}$. Similarly to (\ref{equ3.15}) we can also get
\begin{equation}\label{equ3.27}
\begin{split}
\frac{1}{\delta}\|u^{k+1}_{e}\|^{2} - \rho\|\nabla u^{k+1}_{e}\|^{2} + \rho\|z^{k+1}_{e}\|^{2}+\frac{1}{\rho}\|b^{k+1}_{e}\|^{2} + \tau_{0}(\|Q u^{k}_{e}\|^{2}+\|Q u^{k+1}_{e}\|^{2})\\
+\left(\frac{1}{\delta}\|u^{k+1} - u^{k}\|^{2}-\rho \|\nabla (u^{k+1} - u^{k})\|^{2}-\tau_{0}\|Q(u^{k+1} - u^{k})\|^{2}\right)\\
+ 2\langle t^{k+1}_{e}, z^{k+1}_{e}\rangle + 2\langle l^{k+1}_{e}, u^{k+1}_{e}\rangle + \rho\|\nabla u^{k+1} - z^{k}\|^{2} \\
= \frac{1}{\delta}\|u^{k}_{e}\|^{2} - \rho\|\nabla u^{k}_{e}\|^{2} + \rho\|z^{k}_{e}\|^{2}+\frac{1}{\rho}\|b^{k}_{e}\|^{2}.
\end{split}
\end{equation}
Due to $\bar{u}, u^{k+1}\in U$, by the definition of normal cone we infer that
\[
\langle l^{k+1}, \bar{u} - u^{k+1}\rangle \leq 0,
\]
and
\[
\langle \bar{l}, u^{k+1} -\bar{u}\rangle \leq 0.
\]
Add the two inequalities we obtain that $\langle l^{k+1}_{e}, u^{k+1}_{e}\rangle\geq 0$.
Then similarly to the proof below (\ref{equ3.15}) we can complete the proof here.
\end{pf}

\textbf{Remark}: If we consider the constraint $u\in U$ in Algorithm 1, the iterative formula with respect to $u^{k}$ is rewritten as
\[
u^{k+1} = P_{U}\left(u^{k} - \delta\left(\tau^{k+1}(\textbf{1}-fe^{-u^{k}})+\rho \textrm{div}(z^{k}-\nabla u^{k})+ \textrm{div} b^{k}\right)\right).
\]
Assume that $u^{k+1}$ still satisfies that the discrepancy function $\mathcal{K}\leq 0$, then the convergence properties can be still guaranteed. In section \ref{sec5}, we choose $U=[\log f_{\min}, \log f_{\max}]^{m\times n}$ for the exponential model, and find that the convergence of Algorithm 1 still exists.

In Algorithm 1, a fixed step $\delta$ is used. However, through further investigation we observe that a variable step $\delta^{k}$ can be used to accelerate our algorithm. This can be stated by the following conclusion.

\begin{corollary}\label{cor2}
(the convergence property of the proposed method with variable step)
Let $(u^{k}, z^{k}, b^{k})$ be the sequence generated by Algorithm 1 with a variable step $\delta^{k}$ that satisfies
$\delta^{k}\leq \frac{1}{\tau_{k} L_{D} + \rho \|\triangle\|_{2}+\epsilon_{0}}$ ($\epsilon_{0}>0$ is a small constant) and $\delta^{k}$ is monotonically increasing for sufficiently large $k$. Then it converges to a point $(\bar{u}, \bar{z}, \bar{b})$ where the first-order optimality conditions for (\ref{equ3.7}) are satisfied.
\end{corollary}

\begin{pf}
The proof is analogous to that presented in Theorem \ref{the1}, and we only illustrate the difference due to limited space. Substitute $\delta^{k}$ for $\delta$ in the formulas (\ref{equ3.8})-(\ref{equ3.16}). Since $\delta^{k}$ is monotonically increasing for sufficiently large $k$, there exists $k_{0}>0$ satisfies $\delta^{k+1}\geq \delta^{k}$ for any $k\geq k_{0}$. Hence we obtain that
\begin{equation}\label{equ3.24}
\frac{1}{\delta^{k+1}}\|u^{k+1}_{e}\|^{2} - \rho\|\nabla u^{k+1}_{e}\|^{2} + \rho\|z^{k+1}_{e}\|^{2}+\frac{1}{\rho}\|b^{k+1}_{e}\|^{2}
\leq \frac{1}{\delta^{k}}\|u^{k}_{e}\|^{2} - \rho\|\nabla u^{k}_{e}\|^{2} + \rho\|z^{k}_{e}\|^{2}+\frac{1}{\rho}\|b^{k}_{e}\|^{2}
\end{equation}
Due to $\delta^{k}\leq \frac{1}{\tau_{k} L_{D} + \rho \|\triangle\|_{2}+\epsilon_{0}}$, we have $\frac{1}{\delta^{k}}\|u^{k}_{e}\|^{2} - \rho\|\nabla u^{k}_{e}\|^{2}\geq \epsilon_{0}\|u^{k}_{e}\|^{2}$, and then according to (\ref{equ3.24}) we conclude that $(u^{k}_{e}, z^{k}_{e}, b^{k}_{e})$ is bounded and hence $(u^{k}, z^{k}, b^{k})$ is also bounded.

Choose $j_{0}\geq k_{0}$. Similarly to (\ref{equ3.17})-(\ref{equ3.18}) we obtain that
\begin{equation}\label{equ3.25}
\sum_{k=j_{0}}^{+\infty}\left(\frac{1}{\delta^{k+1}}-\rho \|\triangle\|_{2}-\tau^{k+1} L_{D}\right)\|u^{k+1} - u^{k}\|^{2}+\sum_{k=j_{0}}^{+\infty}\rho\|\nabla u^{k+1} - z^{k}\|^{2}<+\infty.
\end{equation}
Then similarly to the proof below (\ref{equ3.18}) we can complete the proof here.
\end{pf}

\section{Extension to the case of spatially-adaptive parameter selection}\label{sec4}
\setcounter{equation}{0}

The parameter $\tau$ in (\ref{equ1.4}) controls the smoothness of the denoised image generated by the TV regularizer.
As we know, small $\tau$ is more suitable for the homogeneous regions of the image to remove the noise sufficiently, while large $\tau$ is more appropriate in the regions richly contain the textures and details. Therefore, TV models with the spatially-adaptive regularization parameter were widely researched recently, see \cite{IEEETIP:STV, JSC:STV, IP:STV} for instance. However, the parameter is difficult to compute and needs to solve the corresponding TV regularized variational model for many times.

In this section, we use the Newton iteration to update the regularization parameter and extend Algorithm 1 to the case of the TV model with spatially-adaptive parameter. To this end, we first define the local discrepancy function as follows:
\begin{equation}\label{equ4.1}
\mathcal{K}(\tau_{i,j}) = \frac{1}{r^{2}}\sum_{i_{1}=i-r}^{i+r}\sum_{j_{1}=j-r}^{j+r}\left(A_{1} \tau + A_{2} + f e^{-(A_{1} \tau + A_{2})}- \log f\right)_{i_{1},j_{1}} - \bar{C},
\end{equation}
where $r$ denotes the size of the local region centered at the pixel $(i, j)$. The formula (\ref{equ4.1}) is just the local version of the discrepancy function $\mathcal{K}(\tau; w^{k})$ defined in section \ref{sec3}, and the essential difference lies in a regularization parameter $\tau \in \mathbb{R}^{m\times n}$, rather than the scale parameter $\tau>0$, is used here.

For the convenience of description, we column-wise stack $\tau$ into a vector, and denote $\mathcal{\bar{K}}(\tau) = [\mathcal{K}(\tau_{s})]_{s}: \mathbb{R}^{mn}\rightarrow \mathbb{R}^{mn}$. Here the index $s=(i-1)\times n + j$ corresponds to the pixel $(i, j)$ in the image. Similarly to the discussion in section \ref{sec3}, $\tau$ is computed by solve the system of equations $\mathcal{\bar{K}}(\tau)=\textbf{0}$, which can be solved by the Newton iteration as follows:
\begin{equation}\label{equ4.2}
\tau^{k+1} = \tau^{k} -  \left(\nabla \mathcal{\bar{K}}(\tau^{k})\right)^{-1}\mathcal{\bar{K}}(\tau^{k}),
\end{equation}
where $\nabla \mathcal{\bar{K}}(\tau^{k}) = \left(\partial_{s_{2}}\mathcal{K}(\tau_{s_{1}}^{k})\right)_{1\leq s_{1}, s_{2}\leq mn}$. Therefore, the computation of the inverse of a $mn\times mn$ matrix is required in each iteration step of the Newton method and the computational cost is rather expensive.

In this paper, we simplify the update of $\tau$ by the following strategy. Due to the properties of local regions in the image are similar, we assume that the value of $\tau$ is constant in the local regions. Specifically, for the index $(i,j)$, assume that $\tau_{i_{1},j_{1}}=\tau_{i,j}$ for any $i-r\leq i_{1}\leq i+r, j-r\leq j_{1}\leq j+r$.
Then the local discrepancy function in (\ref{equ4.1}) is only a function with respect to $\tau_{i,j}$, and hence $\tau_{i,j}$ can be updated by
\begin{equation}\label{equ4.3}
\tau^{k+1}_{s} = \tau^{k}_{s} -  \left(\partial \mathcal{K}(\tau^{k}_{s})\right)^{-1}\mathcal{K}(\tau^{k}_{s}).
\end{equation}
Let $h$ denote the mean operator with size $r$ in convolution-form, and $R(\tau)=A_{1} \tau + A_{2} + f e^{-(A_{1} \tau + A_{2})}$. Then based on the patch-wise constant assumption of $\tau$, the update formula of $\tau$ can be further approximated by
\begin{equation}\label{equ4.4}
\tau^{k+1} = \tau^{k} -  \max\left\{(h\ast R)(\tau^{k}) - \bar{C}, 0\right\}./\left(h\ast \nabla R(\tau^{k})\right),
\end{equation}
where $'\ast'$ denotes the convolution operation, and $'./'$ denotes componentwise division. Here we use the function $'\max'$, which means that $\tau^{k+1} = \tau^{k}$ while $(h\ast R)(\tau^{k}) \leq \bar{C}$.

Due to the parameter $\tau$ obtained by (\ref{equ4.4}) is based on the patch-wise constant assumption, the estimator of $\tau_{i,j}$ should be computed by the average of the estimation values of $\tau$ around the index $(i,j)$. Therefore, the final $\tau$ is obtained by
\begin{equation}\label{equ4.5}
\tau = h\ast \tau.
\end{equation}

The whole process can be described by Algorithm 2 as follows.

\begin{algorithm}[htb]
\caption{ Local Discrepancy Principle Based Linearized Alternating Direction Minimization Algorithm for Multiplicative Noise Removal (LDP-LADM)}
\begin{algorithmic}[2]
\REQUIRE  noisy image $f$; choose $\rho >0$, $\delta >0$; maximum inner iteration number $Q$;\\
\textbf{Output}: denoised log transformed image $u$; regularization parameter $\tau$. \\
\textbf{Initialization}: $k=0; b^{0}=0; u^{0}=f; z^{0} = \nabla u^{0}$; $\tau^{0} = \tau_{0}$.\\
\textbf{Iteration}: \\
~~~$q =0; \tau^{k, q} = \tau^{k};$ \\
~~~while $q<Q$  \\
~~~~~~$\tau^{k, q+1} = \tau^{k, q} -  \max\left\{(h\ast R)(\tau^{k, q}) - \bar{C}, 0\right\}./\left(h\ast \nabla R(\tau^{k, q})\right)$; \\
~~~~~~$q=q+1;$ \\
~~~end while \\
~~~$\tau^{k+1} = h\ast \tau^{k, Q}$; \\
~~~$u^{k+1} = u^{k} - \delta\left(\tau^{k+1}(\textbf{1}-fe^{-u^{k}})+\rho \textrm{div}(z^{k}-\nabla u^{k})+ \textrm{div} b^{k}\right)$; \\
~~~$z^{k+1} = \textrm{arg}\min_{z}\left\{\|z\|_{1}+ \langle b^{k}, z-\nabla u^{k+1}\rangle + \frac{\rho}{2}\|z-\nabla u^{k+1}\|_{2}^{2} \right\}$; \\
~~~$b^{k+1}=b^{k} + \rho(z^{k+1} - \nabla u^{k+1})$; \\
~~~$k = k+1$; \\
until some stopping criterion is satisfied. \\
return $u = u^{k+1}$ and $\tau = \tau^{k+1}$.
\end{algorithmic}
\end{algorithm}

\section{Numerical experiments}\label{sec5}
\setcounter{equation}{0}

In this section, we evaluate the performance of the proposed algorithms by various experiments. First, the ability of auto-selection of the regularization parameter $\tau$ in the proposed DP-LADM algorithm is investigated. Second, the proposed algorithms are compared with those of the current state-of-the-art methods: one is the PLAD algorithm proposed in \cite{SIAMJSC:Linearized}, which was proved to be more efficient than the widely used augmented Lagrangian method \cite{IEEETIP:MIDAL, JMIV:ST} for multiplicative noise removal at present; the other is the spatially adapted total variation model proposed in \cite{IEEETIP:STV}, which solved the TV model with spatially-adapted regularization parameter by the augmented Lagrangian method for several times.

\begin{figure}
  \centering
  \subfigure[]{
    \label{fig5.1:subfig:a} 
    \includegraphics[width=1.3in,clip]{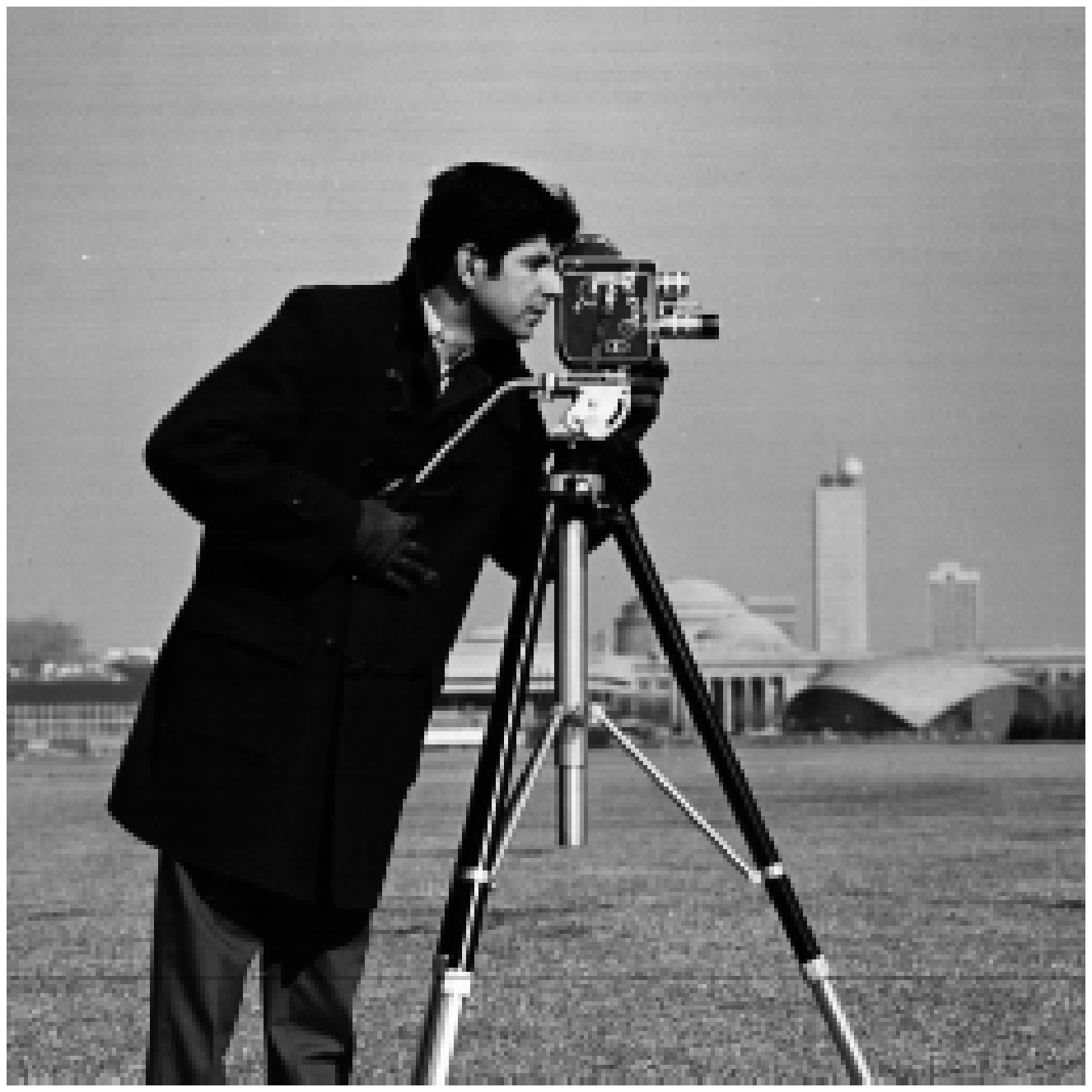}}
  \hspace{0pt}
  \subfigure[]{
    \label{fig5.1:subfig:b} 
    \includegraphics[width=1.3in,clip]{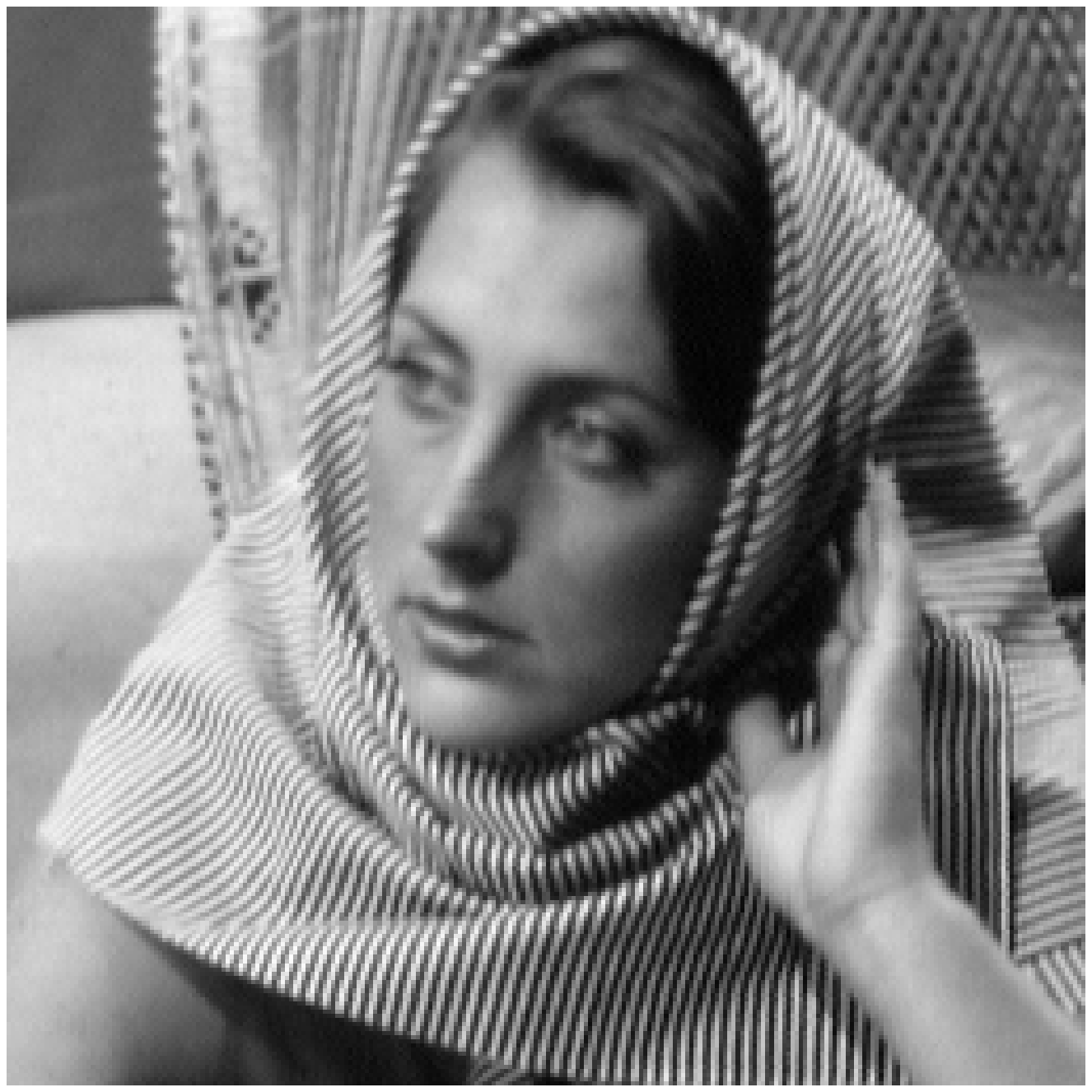}}
  \subfigure[]{
    \label{fig5.1:subfig:c} 
    \includegraphics[width=1.3in,clip]{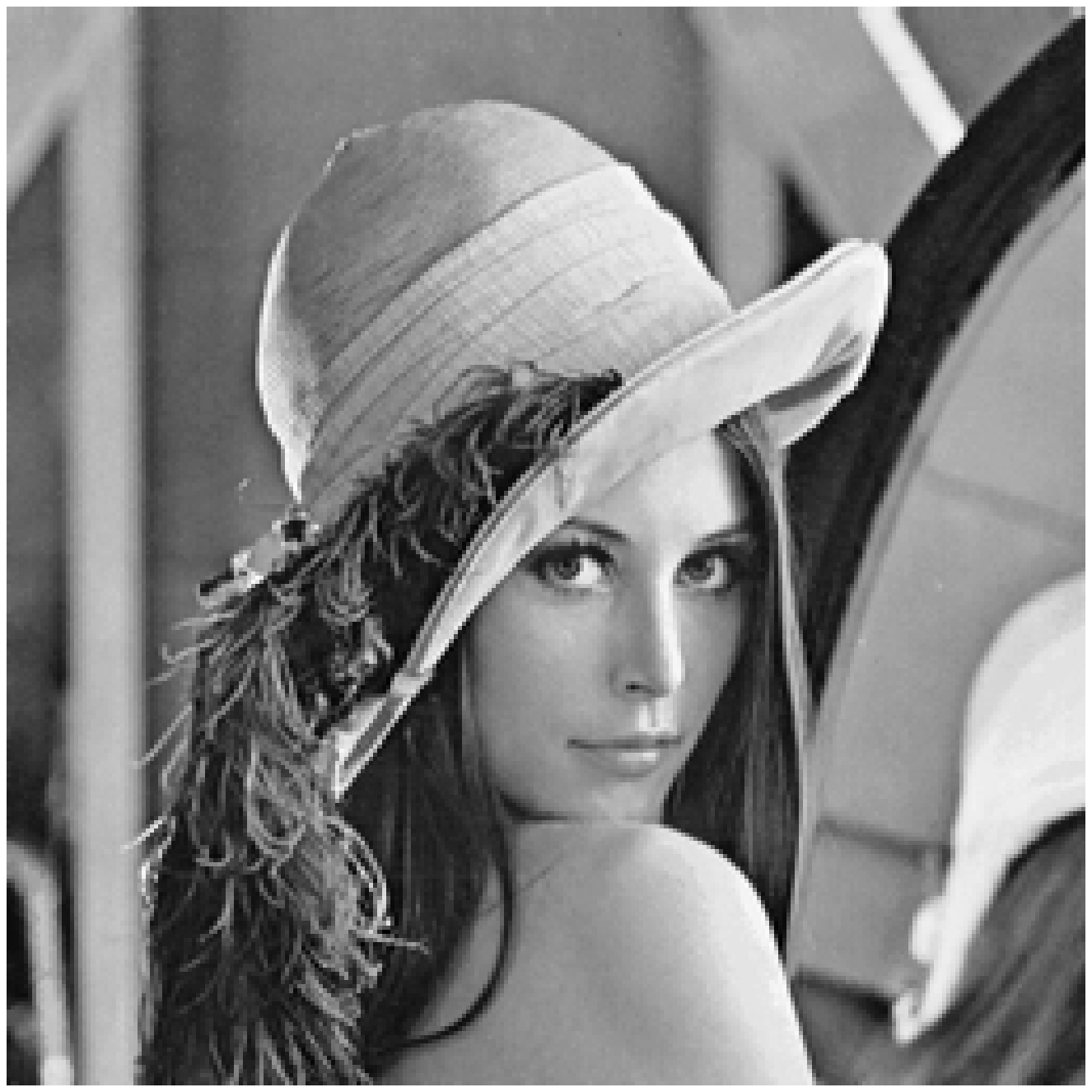}}  \\
  \hspace{0pt}
  \subfigure[]{
    \label{fig5.1:subfig:d} 
    \includegraphics[width=1.3in,clip]{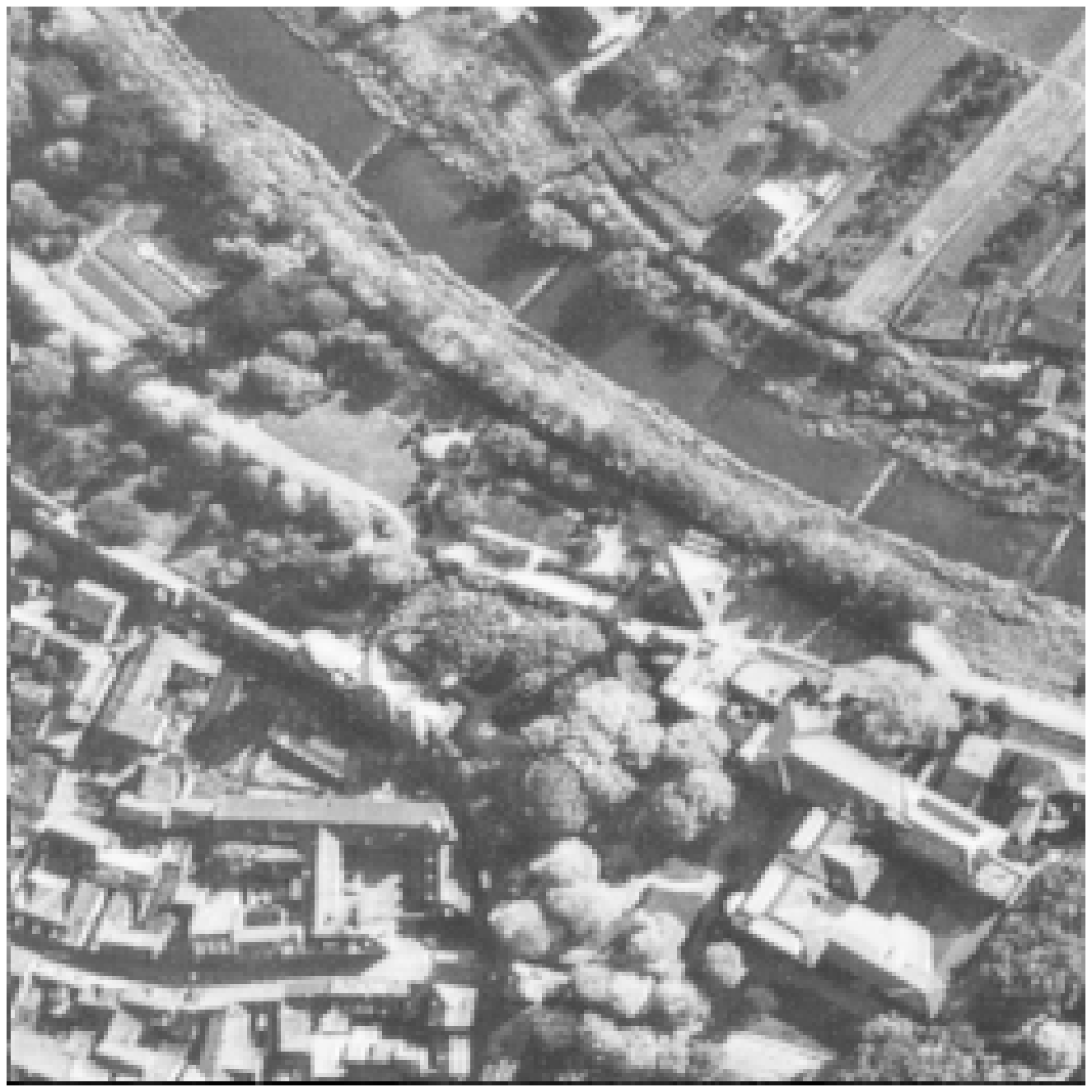}}
  \hspace{0pt}
  \subfigure[]{
    \label{fig5.1:subfig:e} 
    \includegraphics[width=1.3in,clip]{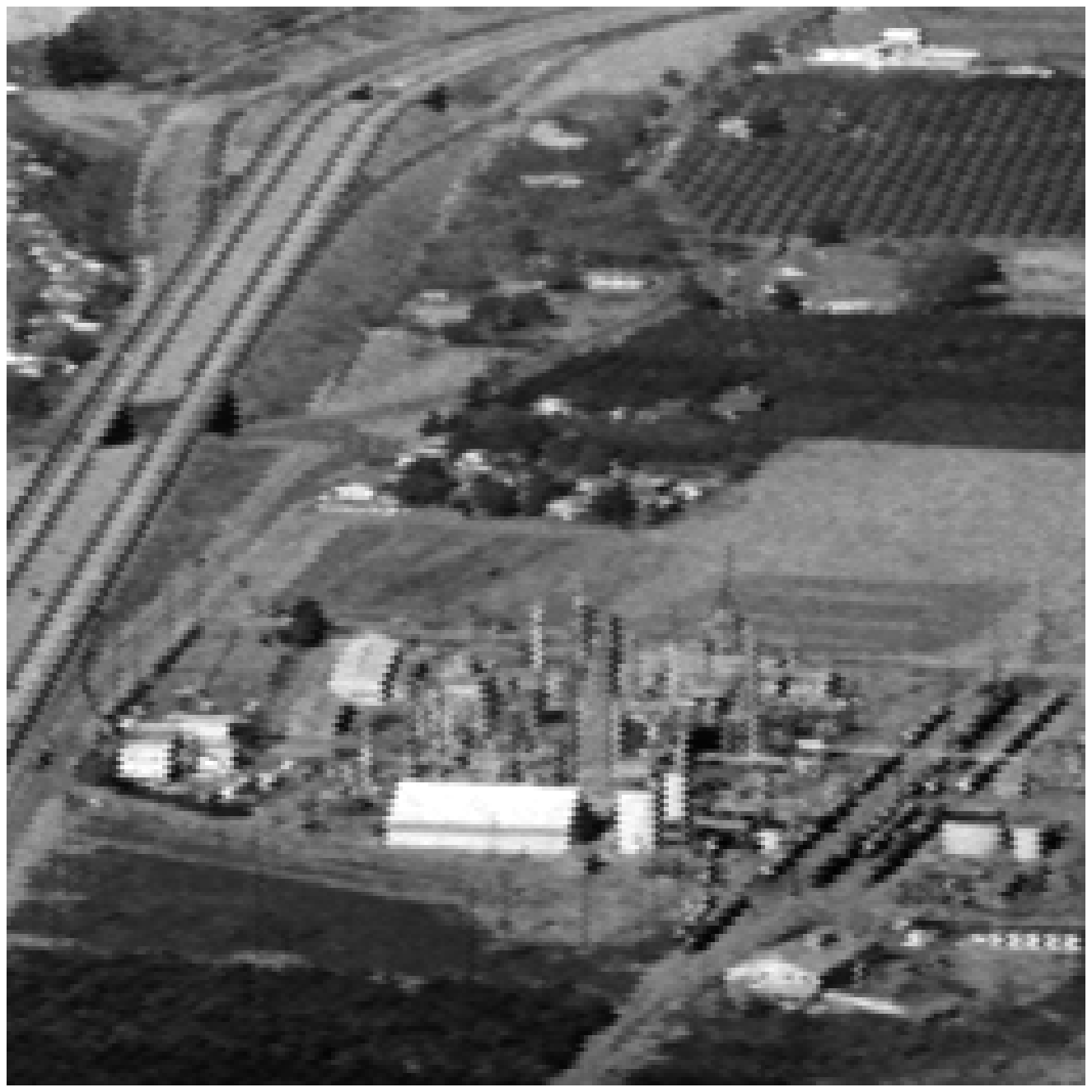}}
  \hspace{0pt}
  \subfigure[]{
    \label{fig5.1:subfig:f} 
    \includegraphics[width=1.3in,clip]{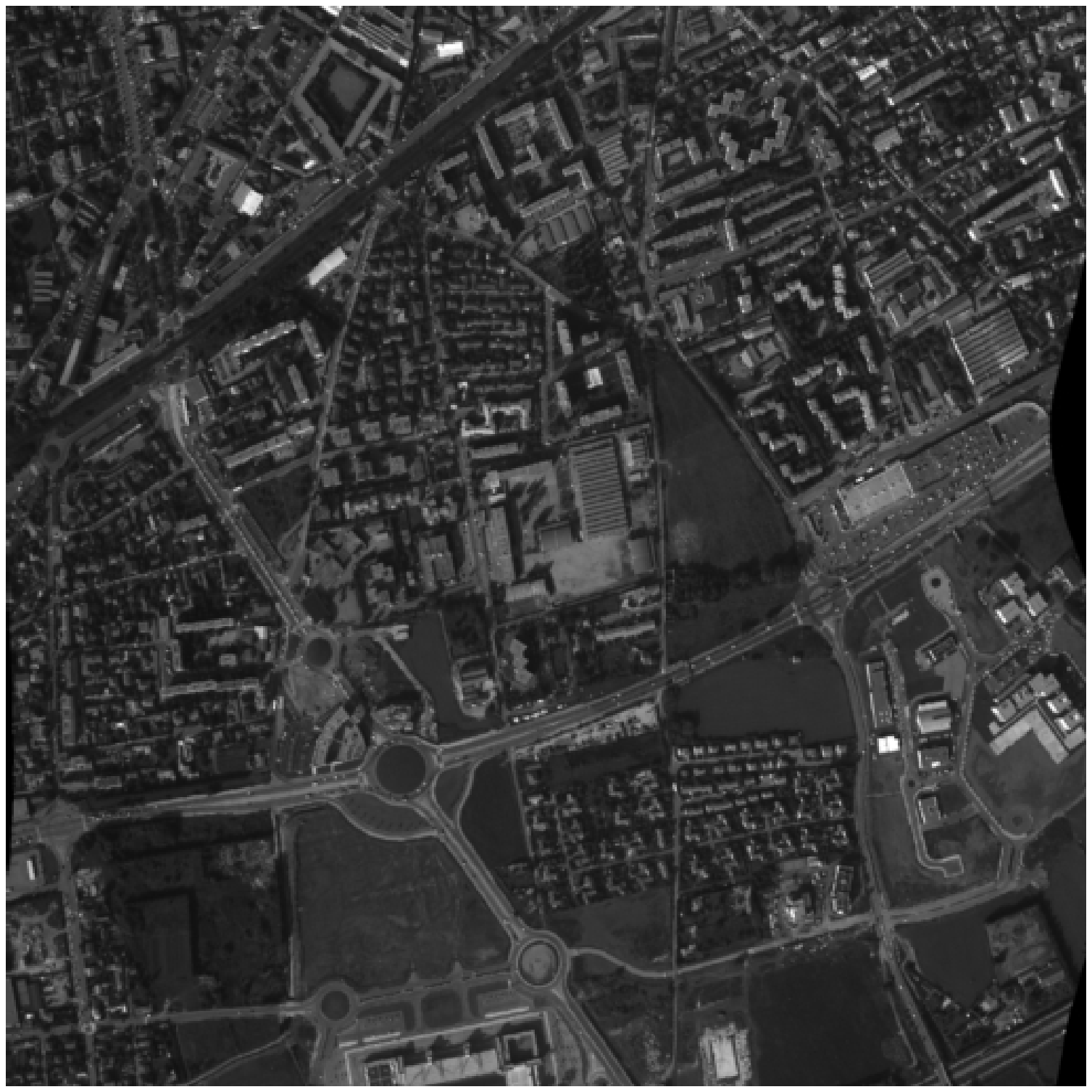}}
\caption{Original images: (a)Cameraman, (b)Barbara, (c)Lena, (d)Remote1, (e)Remote2, (f)Nimes
}
\label{fig5.1}
\end{figure}

The codes of the proposed algorithms and the methods used for comparison are entirely written in Matlab, and all the experiments are performed under Windows XP and MATLAB R2008a running on a laptop with an Intel Pentium Dual CPU (2.0G Hz) and 1 GB Memory. We use six test images, which include three remote sensing images, in Figure 5.1 for the experiments below. The size of Figure \ref{fig5.1:subfig:a}-\ref{fig5.1:subfig:e} is $256\times 256$, and the size of Figure \ref{fig5.1:subfig:f} is $512\times 512$.

\subsection{the performance of the proposed DP-LADM algorithm}

We now evaluate the performance of the DP-LADM algorithm, including the influence of the initial regularization parameter $\tau^{0}$ to the final denoised images by Algorithm 1, and verification of the convergence properties proved in section \ref{sec3}. Three images called ``Cameraman", ``Barbara" and ``Remote2" (see Figure \ref{fig5.1}) are used for our test. The corresponding noisy images are generated by multiplying the original image by a realization of Gamma noise based on the formulas in (\ref{equ1.1})-(\ref{equ1.2}). The performance of denoised images is measured quantitatively by means of the peak signal-to-noise ratio (PSNR), which is defined by
\begin{equation}\label{equ5.1}
\textrm{PSNR}(u,\bar{u})=10\lg
\left\{\frac{255^{2}mn}{\|u-\bar{u}\|_2^{2}}\right\},
\end{equation}
where $u$ and $\bar{u}$ denote the original image and the denoised image respectively. During the implementation of Algorithm 1, the parameters $\rho$ is chosen to be $0.75$, the value of $\tau^{k}$ is updated by the Newton method every three iteration, the iteration number $Q$ of the Newton method is fixed as $3$, and the constant $\bar{C}$ is chosen to be $1+\frac{1}{2M}+\frac{1}{12M^{2}}-\frac{1}{2M^{3}}$ for $M\leq 5$ and $1+\frac{1}{2M}+\frac{1}{12M^{2}}-\frac{5}{2M^{3}}$ for $M>5$ (they are also used for Algorithm 2). Moreover, in order to accelerate the iteration, we use a variable step $\delta^{k}$ in our algorithm. Specifically, we set $\delta^{0} = 0.16$, and update $\delta^{k}$ as
$\delta^{k} = \frac{\delta^{0}}{0.4\tau^{k}}$ after $\tau^{k}$ is recalculated. We use the stopping criterion for the proposed iterative algorithm as follows
\begin{equation}\label{equ5.2}
\frac{\|e^{u^{k+1}}-e^{u^{k}}\|_{2}}{\|e^{u^{k}}\|_{2}}<\textrm{tol},
\end{equation}
where $u^{k+1}$ denotes the iterate of the scheme (correspond to the denoised log transformed image). Note that $e^{u^{k}}$ in (\ref{equ5.2}) should be replaced by $u^{k}$ for the I-divergence model. In the experiments we choose $tol=3\times 10^{-4}$.

\begin{figure}
  \centering
  \subfigure[]{
    \label{fig5.2:subfig:a} 
    \includegraphics[width=1.7in,clip]{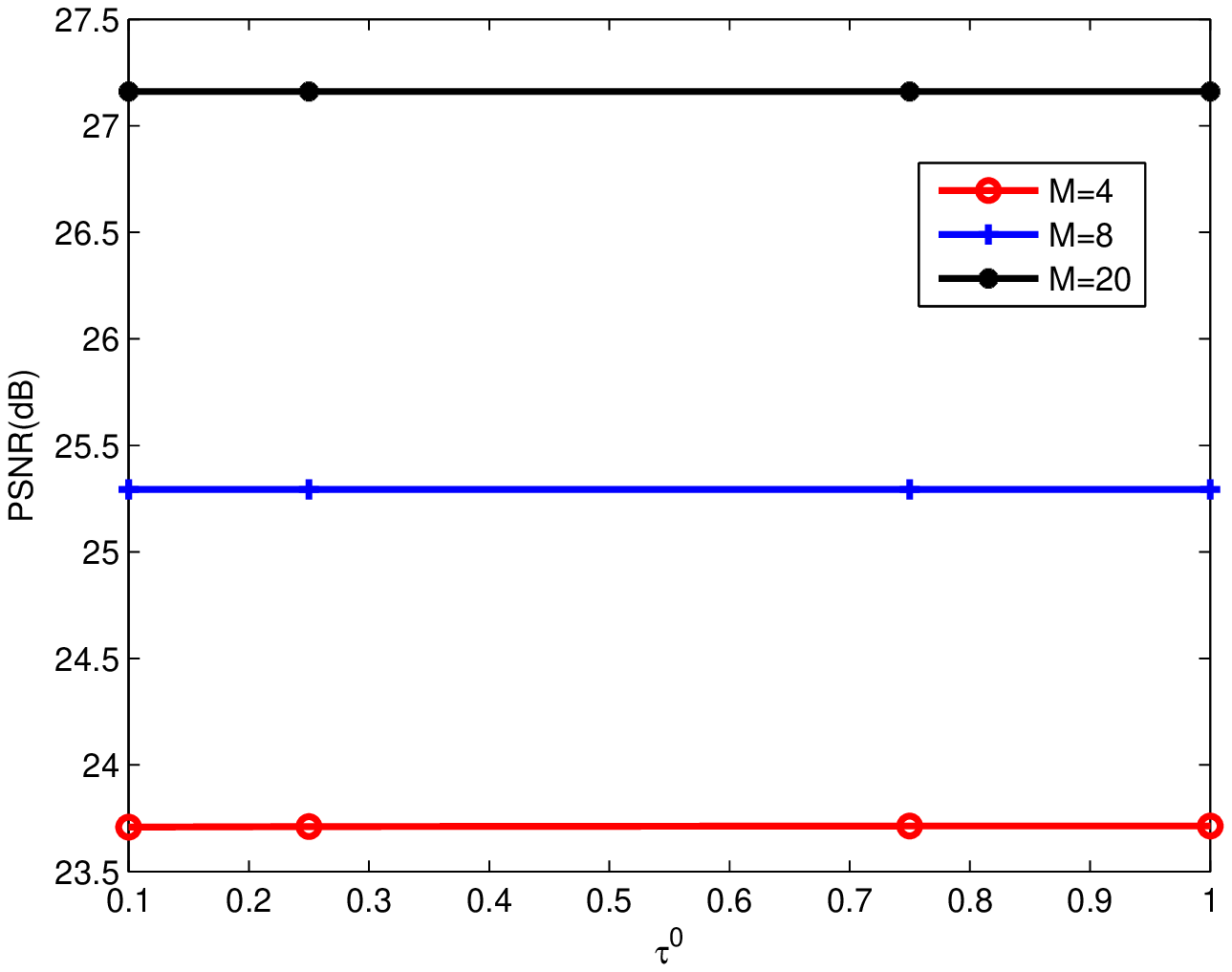}}
  \hspace{0pt}
  \subfigure[]{
    \label{fig5.2:subfig:b} 
    \includegraphics[width=1.7in,clip]{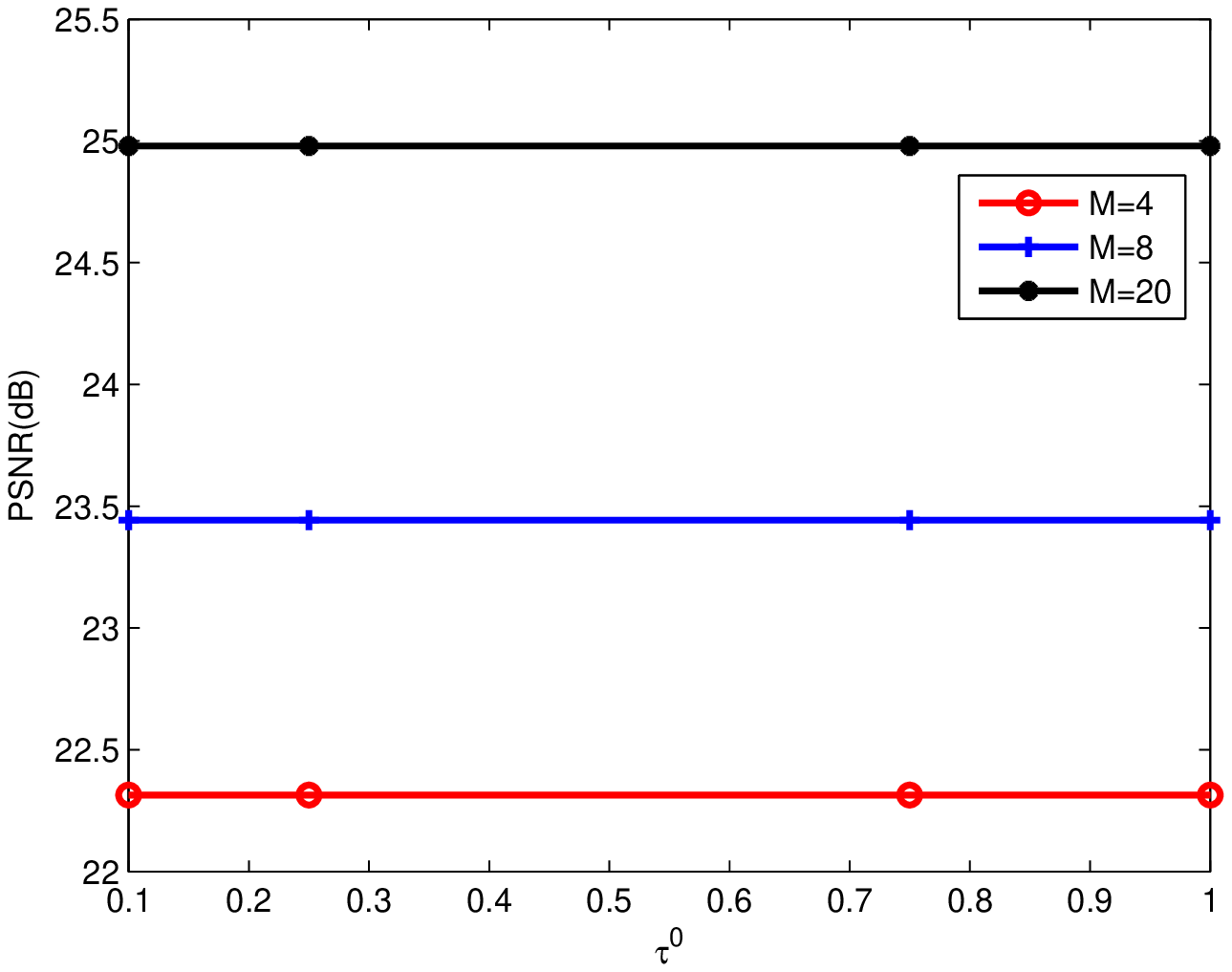}}
  \subfigure[]{
    \label{fig5.2:subfig:c} 
    \includegraphics[width=1.7in,clip]{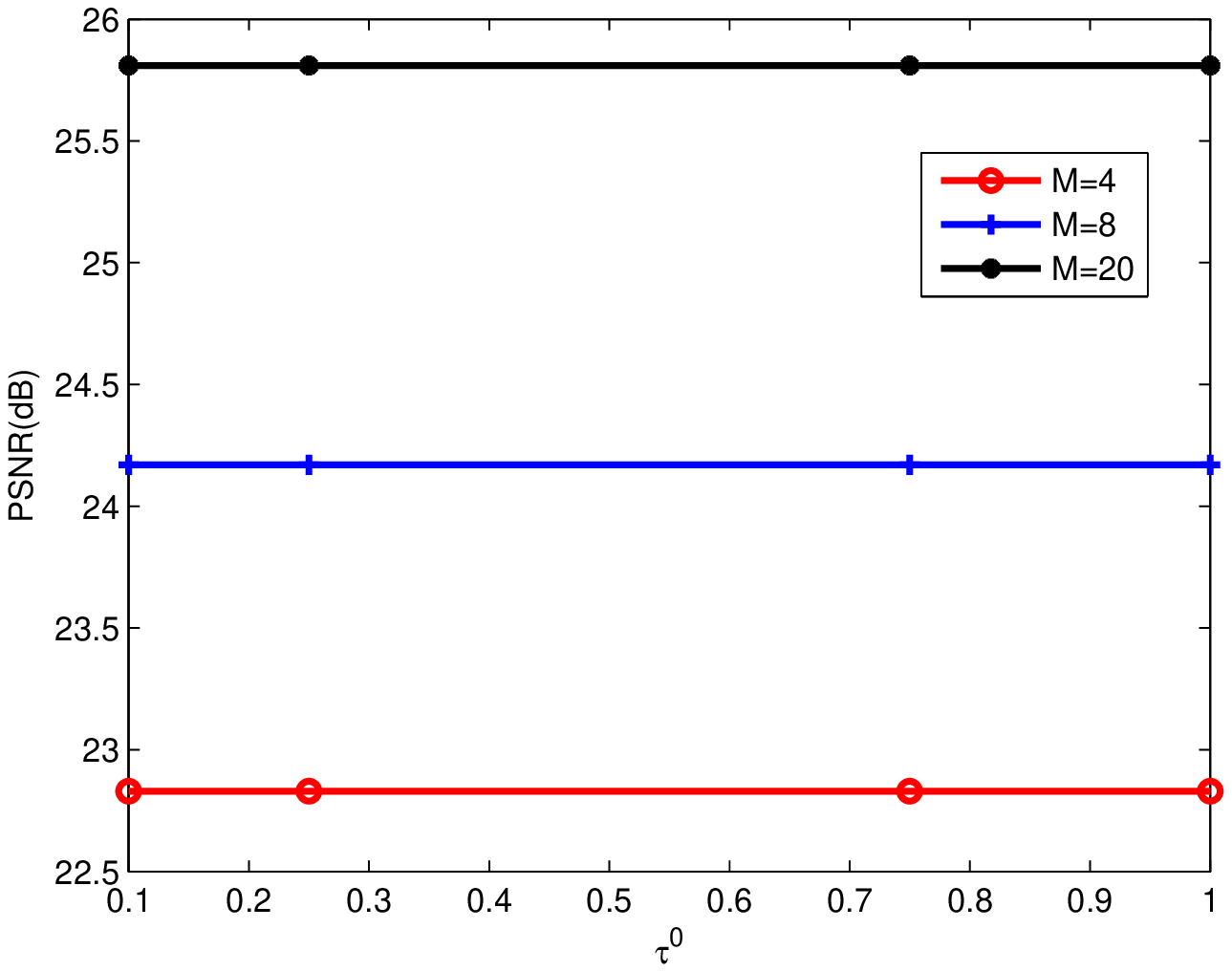}}
\caption{The PSNR values (dB) of denoised images by Algorithm 1 with different $\tau^{0}$. (a) The results of the Cameraman image with $M = 4, 8, 20$, (b) the results of the Barbara image with $M = 4, 8, 20$, (c) the results of the Remote2 image with $M = 4, 8, 20$.
}
\label{fig5.2}
\end{figure}

\begin{figure}
  \centering
  \subfigure[]{
    \label{fig5.3:subfig:a} 
    \includegraphics[width=1.7in,clip]{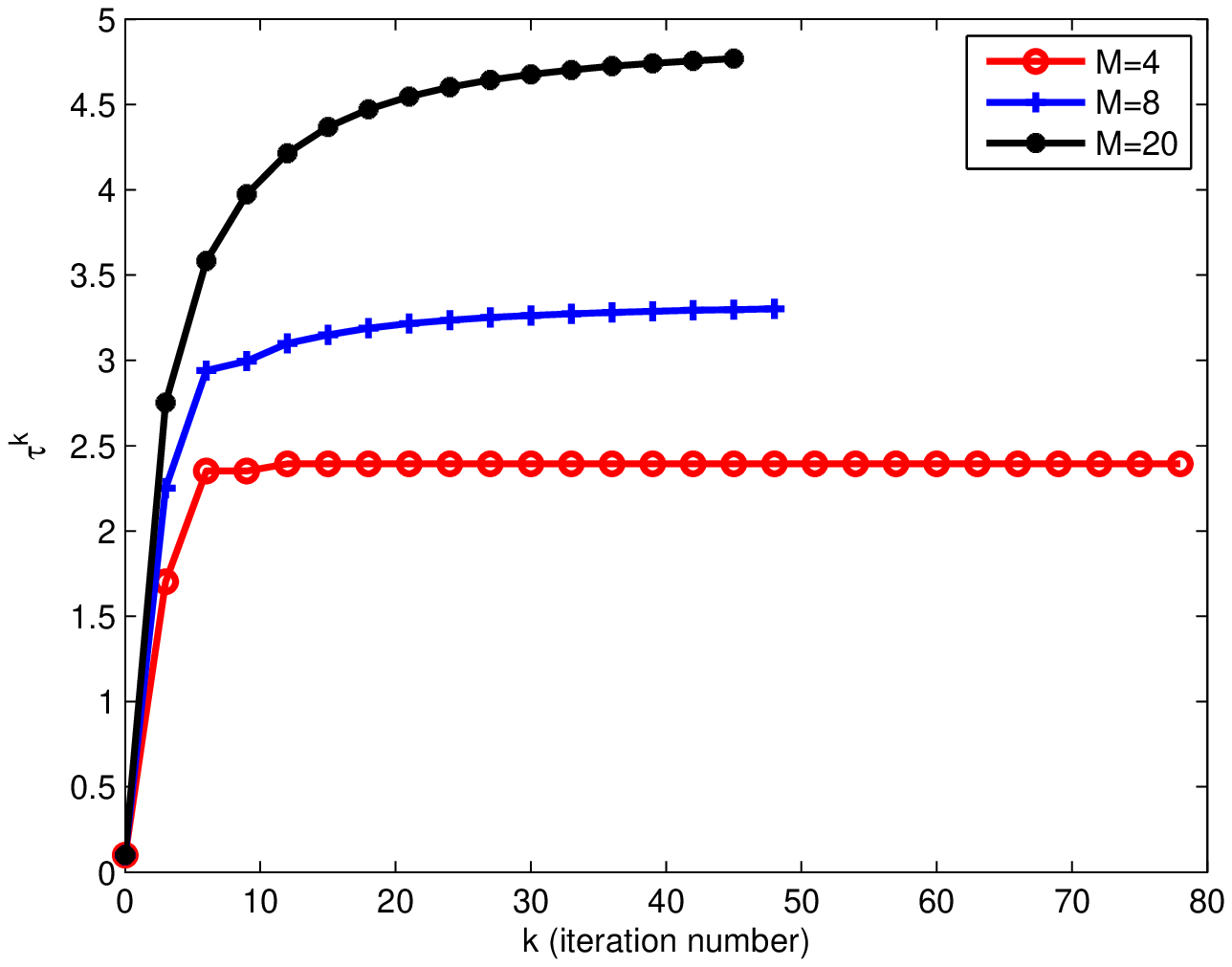}}
  \hspace{0pt}
  \subfigure[]{
    \label{fig5.3:subfig:b} 
    \includegraphics[width=1.7in,clip]{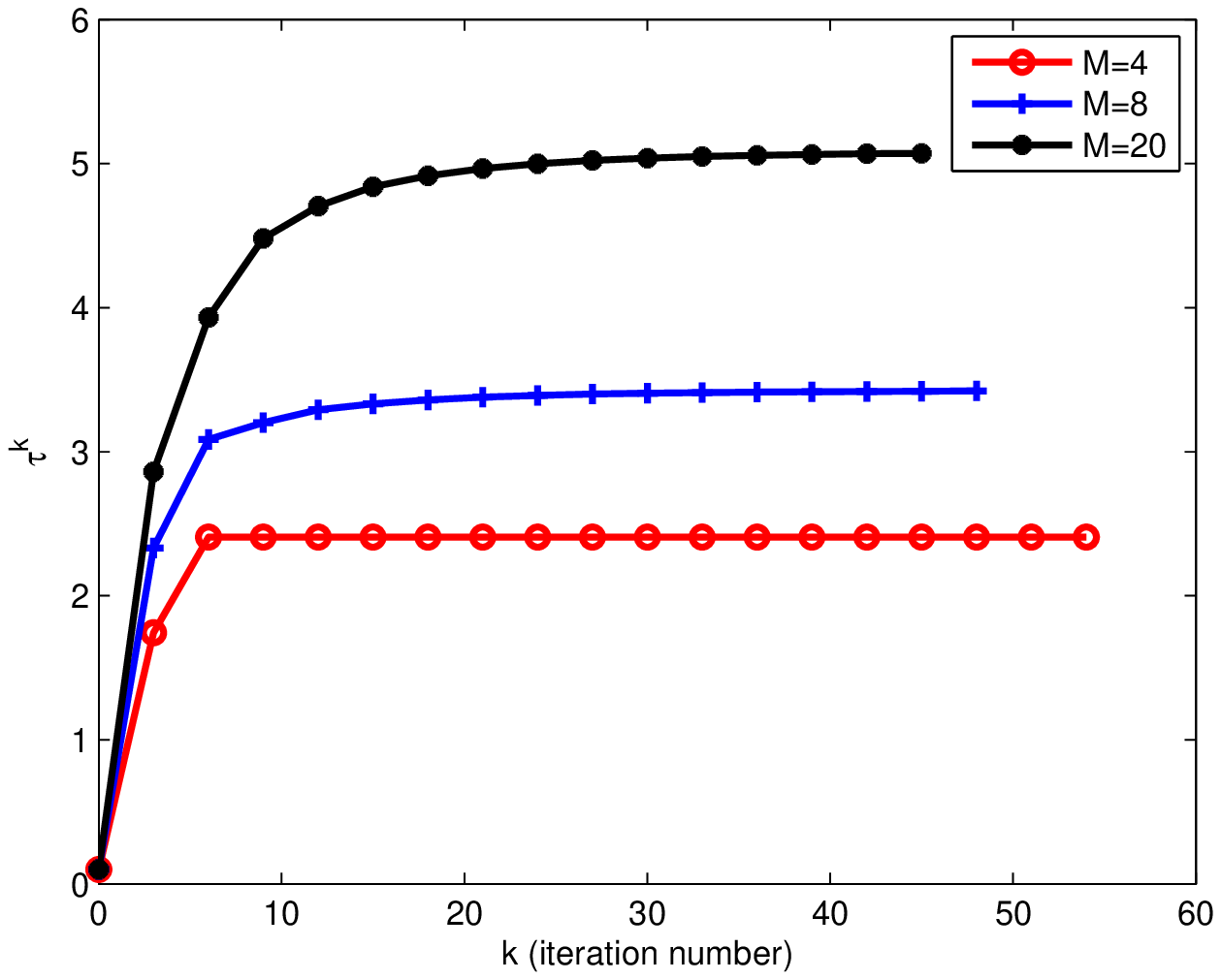}}
  \subfigure[]{
    \label{fig5.3:subfig:c} 
    \includegraphics[width=1.7in,clip]{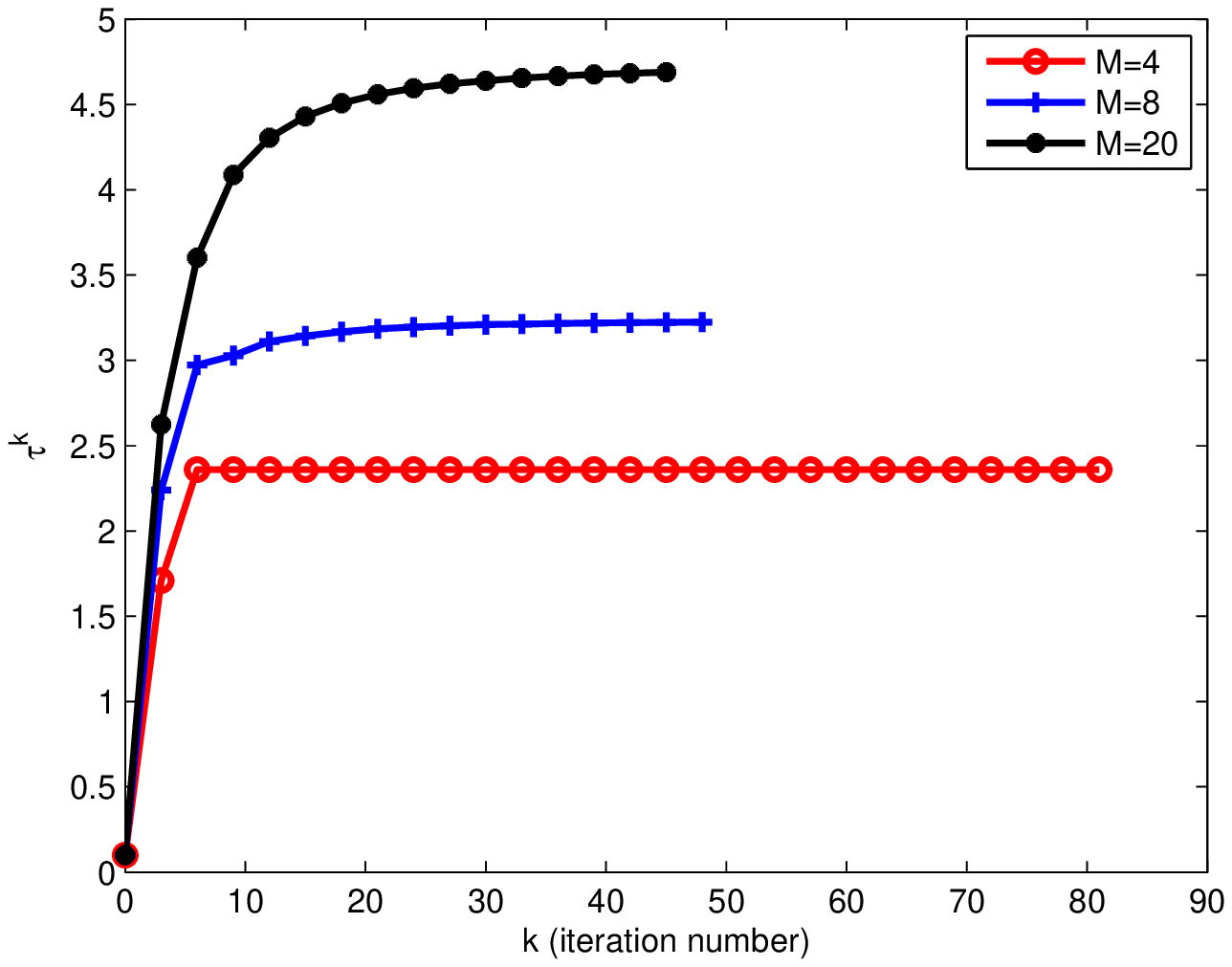}}
  \subfigure[]{
    \label{fig5.3:subfig:d} 
    \includegraphics[width=1.7in,clip]{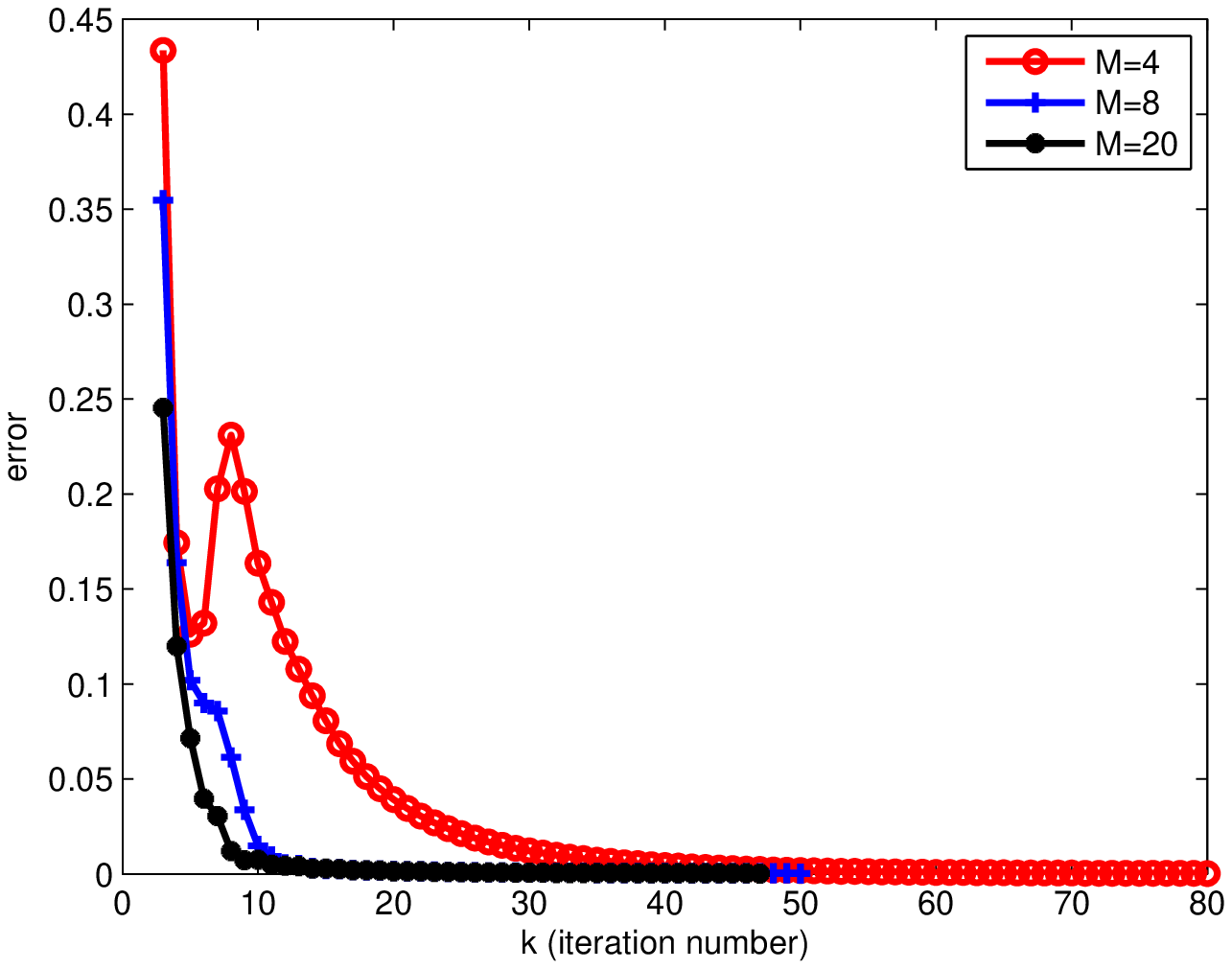}}
  \hspace{0pt}
  \subfigure[]{
    \label{fig5.3:subfig:e} 
    \includegraphics[width=1.7in,clip]{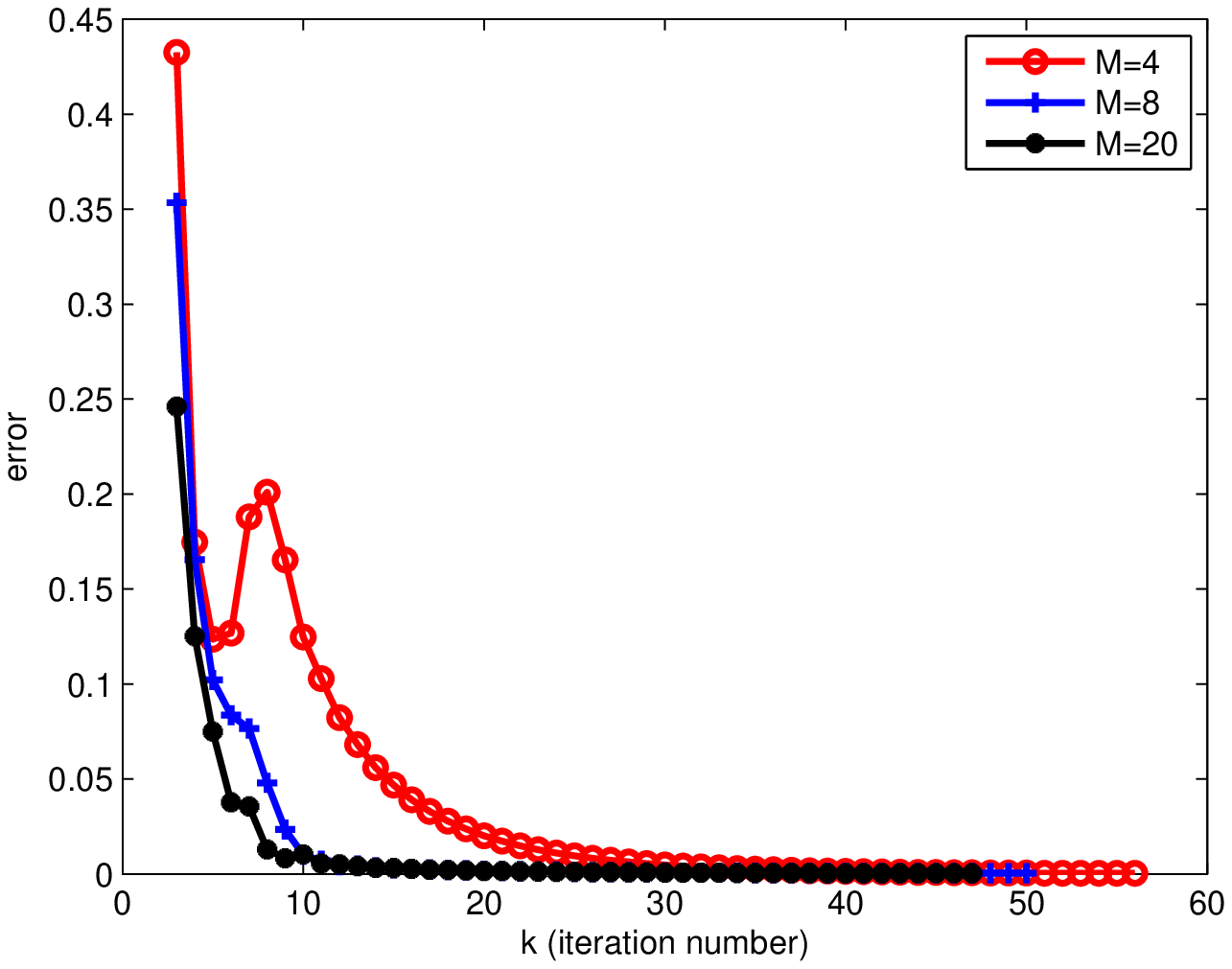}}
  \subfigure[]{
    \label{fig5.3:subfig:f} 
    \includegraphics[width=1.7in,clip]{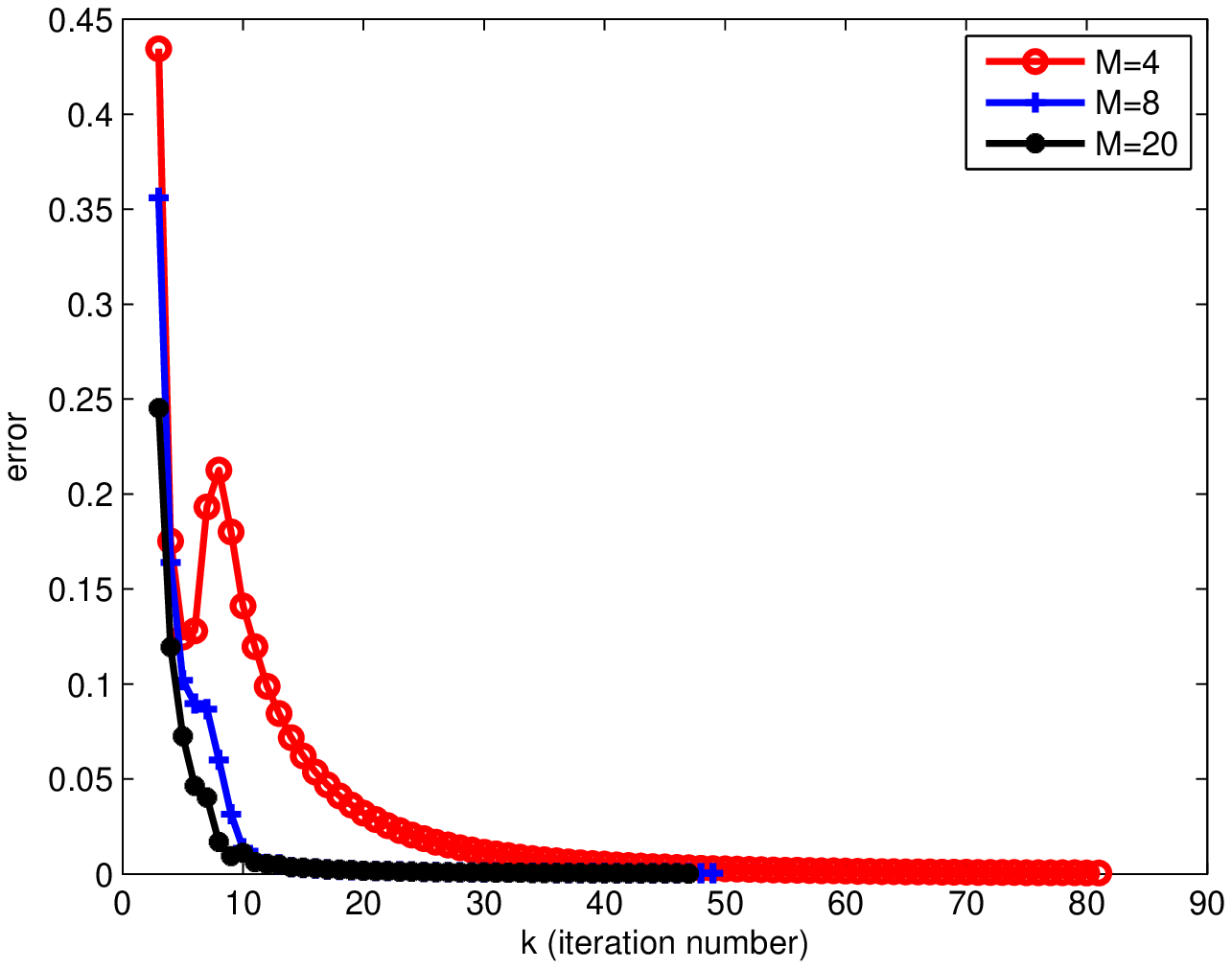}}
  \subfigure[]{
    \label{fig5.3:subfig:g} 
    \includegraphics[width=1.7in,clip]{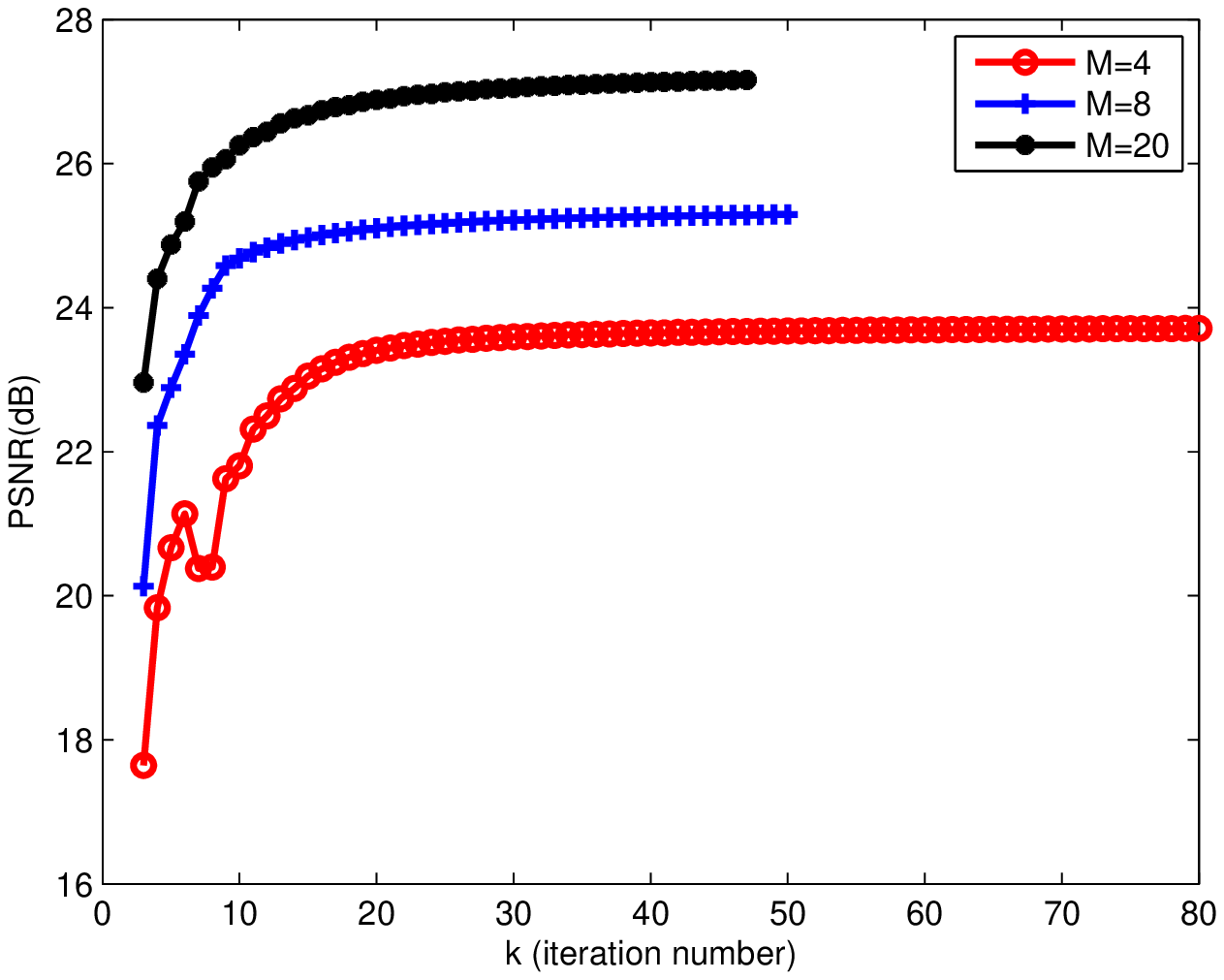}}
  \hspace{0pt}
  \subfigure[]{
    \label{fig5.3:subfig:h} 
    \includegraphics[width=1.7in,clip]{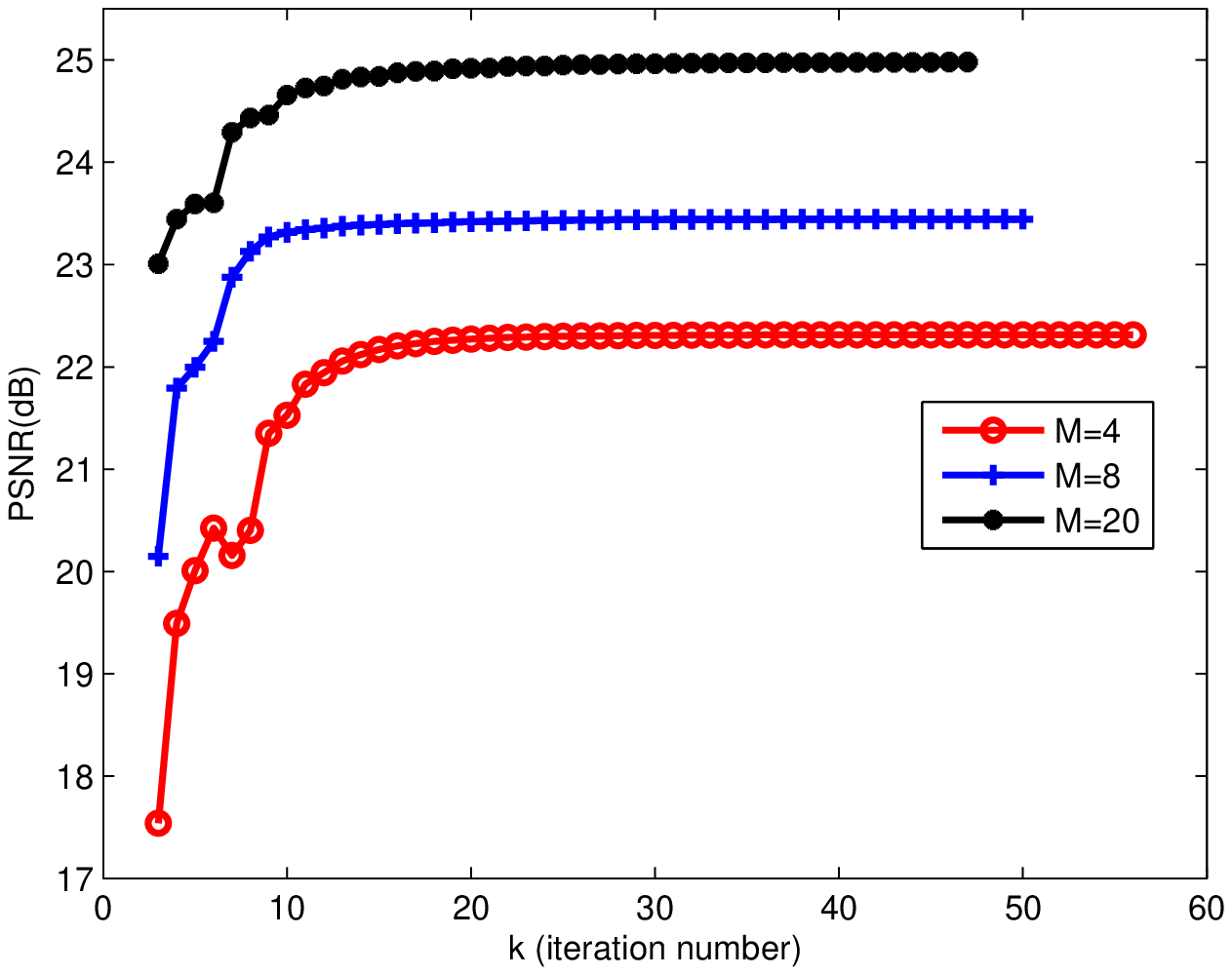}}
  \subfigure[]{
    \label{fig5.3:subfig:i} 
    \includegraphics[width=1.7in,clip]{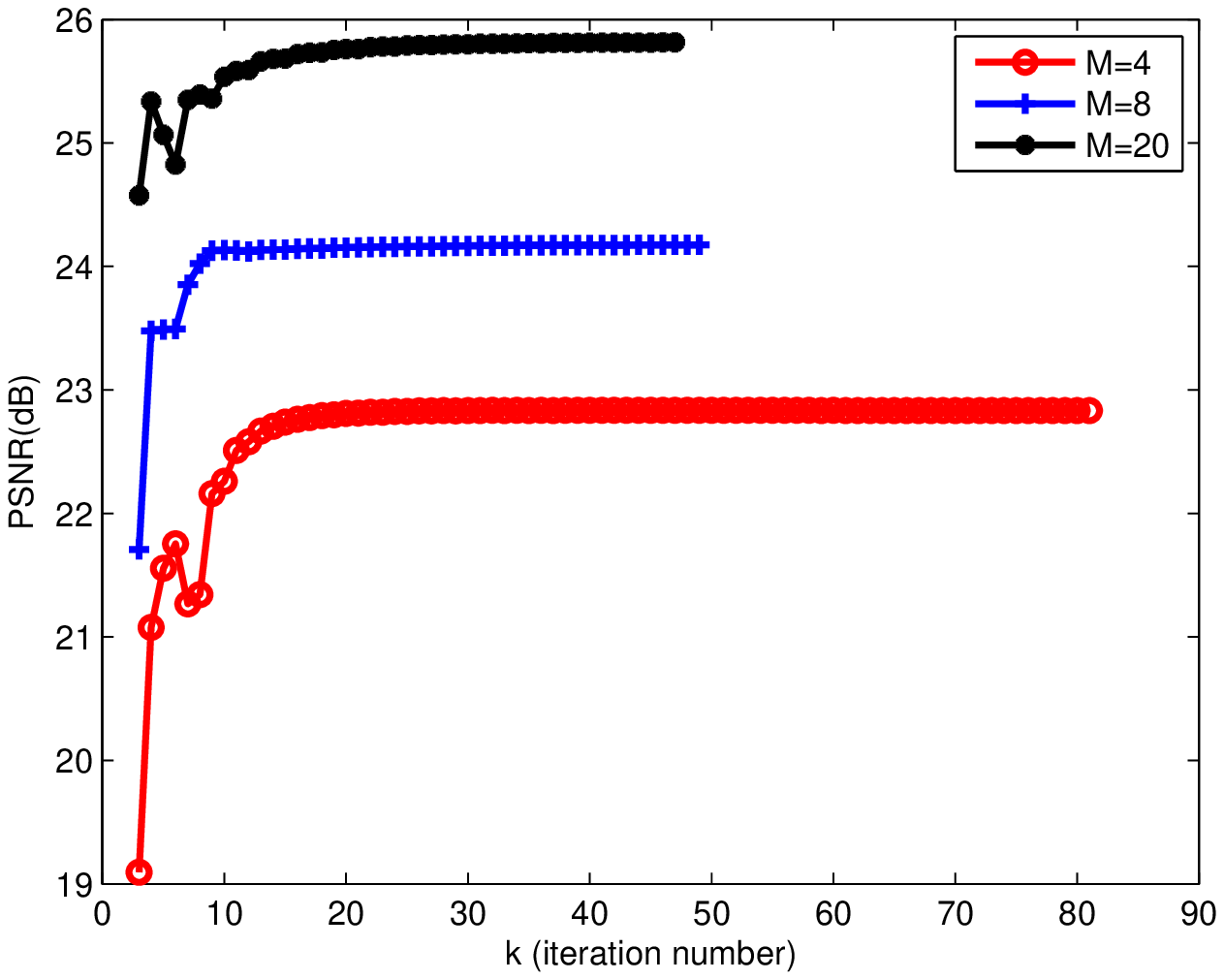}}
\caption{The evolution curves of $\tau^{k}$, the relative error and the PSNR values for different images with different noise level. (a)-(c) plot the evolution curves of $\tau^{k}$ for the image Cameraman, Barbara and Remote2 with $M = 4, 8, 20$ respectively; (d)-(f) plot the evolution curves of the relative error; (g)-(i) plot the evolution curves of the PSNR values.}
\label{fig5.3}
\end{figure}

In order to evaluate the influence of $\tau^{0}$ to the final results of Algorithm 1, we use images with different noise level for our test. The parameter $M$ in (\ref{equ1.2}) reflects the noise level of Gamma noise. Here we choose $M=4,8,20$. In Figure \ref{fig5.2}, we plot the PSNR values of the images denoised by our algorithm with $\tau^{0}$ varying from $0.1$ to $1.0$. From the plots we observe that the PSNR values are rather stable for $0.1\leq \tau^{0}\leq 1.0$ under different noise conditions. In fact, the value of $\tau$ converges to more or less the same value very quickly (refer to Figure \ref{fig5.3} for the convergence case of $\tau$), and therefore the influence of initial $\tau^{0}$ is negligible. We choose $\tau^{0}=0.1$ in the following experiments.

In the next, we verify the convergence properties proved in subsection \ref{subsec3.1}. Figure \ref{fig5.3:subfig:a}-\ref{fig5.3:subfig:c} show the evolution curves of $\tau^{k}$ with the increasing iteration number $k$, which indicate that $\tau^{k}$ converges to some $\bar{\tau}$ very quickly. Moreover, we plot the evolution curves of the relative error $\frac{\|e^{u^{k+1}}-e^{u^{k}}\|_{2}}{\|e^{u^{k}}\|_{2}}$ and the PSNR values versus the iteration number in Figure \ref{fig5.3:subfig:d}-\ref{fig5.3:subfig:i}. From the plots we observe that the convergence properties of the proposed DP-LADM algorithm are really guaranteed by our experiments.

Finally, we show the advantage of the DP-LADM algorithm compared with the PLAD methods. Three images contaminated by Gamma noise with $M=8$ are used for comparison. For the PLAD method of solving the exponential model, we choose the regularization parameter $\lambda$ to be varying from $\frac{1.0}{M}$ to $\frac{5.0}{M}$. Moreover, $\rho = 0.3$, $\delta =0.4$ are chosen for $\lambda<\frac{5.0}{M}$, and $\delta =0.3$ is used for $\lambda=\frac{5.0}{M}$. Table \ref{tab5.1} shows the denoised results of the PLAD algorithm with different regularization parameters. From the PSNR values we observe that the PLAD algorithm is rather sensitive to the parameter $\lambda$ and needs many trials to obtain the optimal value. On the contrary, the proposed DP-LADM algorithm is able to estimate the value of the regularization parameter during the iteration, and obtains satisfactory results with the estimator (see the last column in Table \ref{tab5.1}).

\begin{table} [htbp]
\centering \caption{The values of PSNR (dB) for different regularization parameter $\lambda$ in the PLAD algorithm \cite{SIAMJSC:Linearized}}
\scalebox{0.9}{
\begin{tabular}{|c|c|c|c|c|c|c|}
  \hline
 \backslashbox{image}{$\lambda$}& $\frac{1.0}{M}$ & $\frac{2.0}{M}$ & $\frac{3.0}{M}$ & $\frac{4.0}{M}$ & $\frac{5.0}{M}$ & Algorithm 1\\
  \hline
  Cameraman & 20.54 &  24.99 & 25.08 & 24.32 & 23.65 & \textbf{25.29}\\
  \hline
  Barbara & 20.15 &  23.40 & 23.13 & 22.62 & 22.19 & \textbf{23.44} \\
  \hline
  Lena & 20.97 &  25.58 & 25.85 & 25.08 & 24.29 & \textbf{26.00} \\
  \hline
\end{tabular}}
\label{tab5.1}
\end{table}

\subsection{Comparison with other noise removal methods}

In this subsection, we further report the experiment results comparing our algorithms with those of the current state-of-the-art algorithms \cite{SIAMJSC:Linearized, IEEETIP:STV} for multiplicative noise removal. The stopping criterion in (\ref{equ5.2}) is used for our test here.

\begin{figure}
  \centering
  \subfigure[]{
    \label{fig5.4:subfig:a} 
    \includegraphics[width=2.5in,clip]{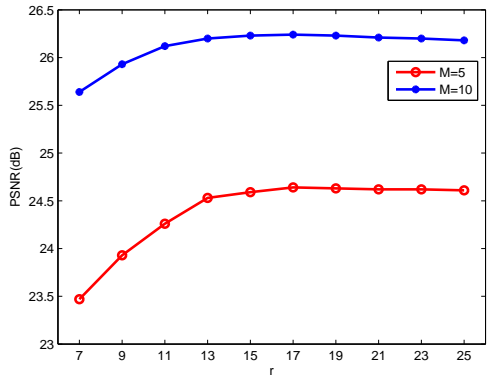}}
  \hspace{0pt}
  \subfigure[]{
    \label{fig5.4:subfig:b} 
    \includegraphics[width=2.5in,clip]{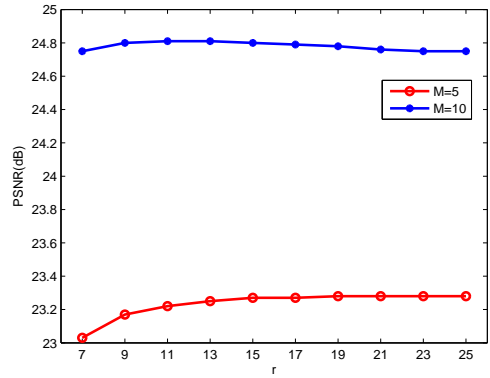}}
\caption{The PSNR values (dB) of denoised images by Algorithm 2 with different window size $r$. (a)Cameraman, (b)Remote2.}
\label{fig5.4}
\end{figure}

Table \ref{tab5.2} lists the PSNR values, the number of iterations and the CPU time of different algorithms for the natural images in Figure \ref{fig5.1}. In this table, ``PLAD\_EXP" and ``PLAD\_DIV" represent the PLAD algorithms for the exponential model and I-divergence model respectively (They were verified to be more efficient than the PLAD method with the correction step (PLADC), see Table 4.4 in \cite{SIAMJSC:Linearized}, thus we don't consider the PLADC algorithms here). ``STV\_AL" represents the spatially-adapted total variation model proposed in \cite{IEEETIP:STV}. Here $(\cdot, \cdot, \cdot )$ denotes the PSNR values, iteration numbers and CPU time in sequence. Specifically, the outer iteration number of ``STV\_AL" is fixed to be $3$.
Note that these methods are all manual parameter models. For the PLAD algorithms, we use the same setting as that adopted in \cite{SIAMJSC:Linearized}, i.e., we fix $\alpha =0.3$, $\delta =0.4$ for the PLAD\_EXP and $\alpha =0.01$, $\delta =8.0$ for the PLAD\_DIV. Besides, $\lambda$ is chosen to be $\frac{2.0}{M}$ for $M=5$ and $\frac{3.0}{M}$ for $M >5$. For the spatially-adapted total variation model, the initial regularization parameter $\tau^{0}$,  the local window size $r$ and the step size $\delta$ are chosen to be $0.1M$, $17$ and $M$ respectively. Moreover, $\tau^{0}$ is set to be $0.1$ in our algorithms. In Algorithm 2, the values of $\tau^{k}$ are updated by the Newton iteration every twenty iteration for the Cameraman and Lena images, and computed every three iteration for the Barbara image with $M=10,15$. The local window size $r$ is chosen to be $17$. In fact, Algorithm 2 is unsensitive to the different values of $r$ while $r$ ranges from $13$ to $25$, see Figure \ref{fig5.4} for instance.

\begin{table} [htbp]
\centering \caption{The comparison of different methods: the given numbers are PSNR (dB)/Iteration number/CPU time(second) }
\scalebox{0.9}{
\begin{tabular}{|c|c|c|c|c|c|c|}
  \hline
  Image & M & PLAD\_EXP \cite{SIAMJSC:Linearized} & PLAD\_DIV \cite{SIAMJSC:Linearized} & STV\_AL \cite{IEEETIP:STV} & Algorithm 1 & Algorithm 2 \\
  \hline
    & 5 & 24.26/50/4.47 & 24.35/103/7.09 & 24.64/3/26.67 & 24.26/51/5.32 & \textbf{24.64}/60/5.89 \\
  \cline{2-7}
    Cameraman & 10 & 25.72/42/3.11 & 25.80/110/6.27 & 26.11/3/23.09 & 25.77/50/5.20 & \textbf{26.24}/77/7.75 \\
  \cline{2-7}
    & 15 & 26.74/32/2.88 & 26.71/67/3.84 & 26.98/3/19.42 & 26.64/46/4.86 & \textbf{27.16}/76/7.45 \\
  \hline
    & 5 & 22.73/54/3.81 & 22.70/76/4.72 & 22.83/3/27.95 & 22.72/53/5.53 & \textbf{22.90}/80/7.99 \\
  \cline{2-7}
    Barbara & 10 & 23.63/46/3.34 & 23.62/69/3.94 & \textbf{24.30}/3/23.63 & 23.78/50/5.31 & 24.17/32/4.76 \\
  \cline{2-7}
    & 15 & 24.65/32/2.82 & 24.62/68/3.91 & \textbf{25.20}/3/19.34 & 24.44/48/4.92 & 24.92/31/4.55 \\
  \hline
    & 5 & 24.80/54/4.71 & 24.77/73/4.17 & \textbf{24.81}/3/26.00 & 24.80/55/5.76 & 24.79/76/7.41 \\
  \cline{2-7}
    Lena & 10 & 26.47/46/4.03 & 26.46/66/3.84 & \textbf{26.86}/3/21.39 & 26.46/52/5.47 & 26.65/75/7.33 \\
  \cline{2-7}
    & 15 & 27.54/34/3.03 & 27.47/64/3.67 & \textbf{27.97}/3/18.17 & 27.38/52/5.61 & 27.76/74/7.31 \\
  \hline
\end{tabular}}
\label{tab5.2}
\end{table}

From the table we observe that Algorithm 2 overall outperforms other algorithms while considering both the PSNR values and CPU time. On the one hand, the LDP-LADM algorithm obtains better PSNR values than the PLAD methods, and on the other hand, it takes less CPU time than the STV\_AL method.

\begin{figure}
  \centering
  \subfigure[]{
    \label{fig5.5:subfig:a} 
    \includegraphics[width=1.5in,clip]{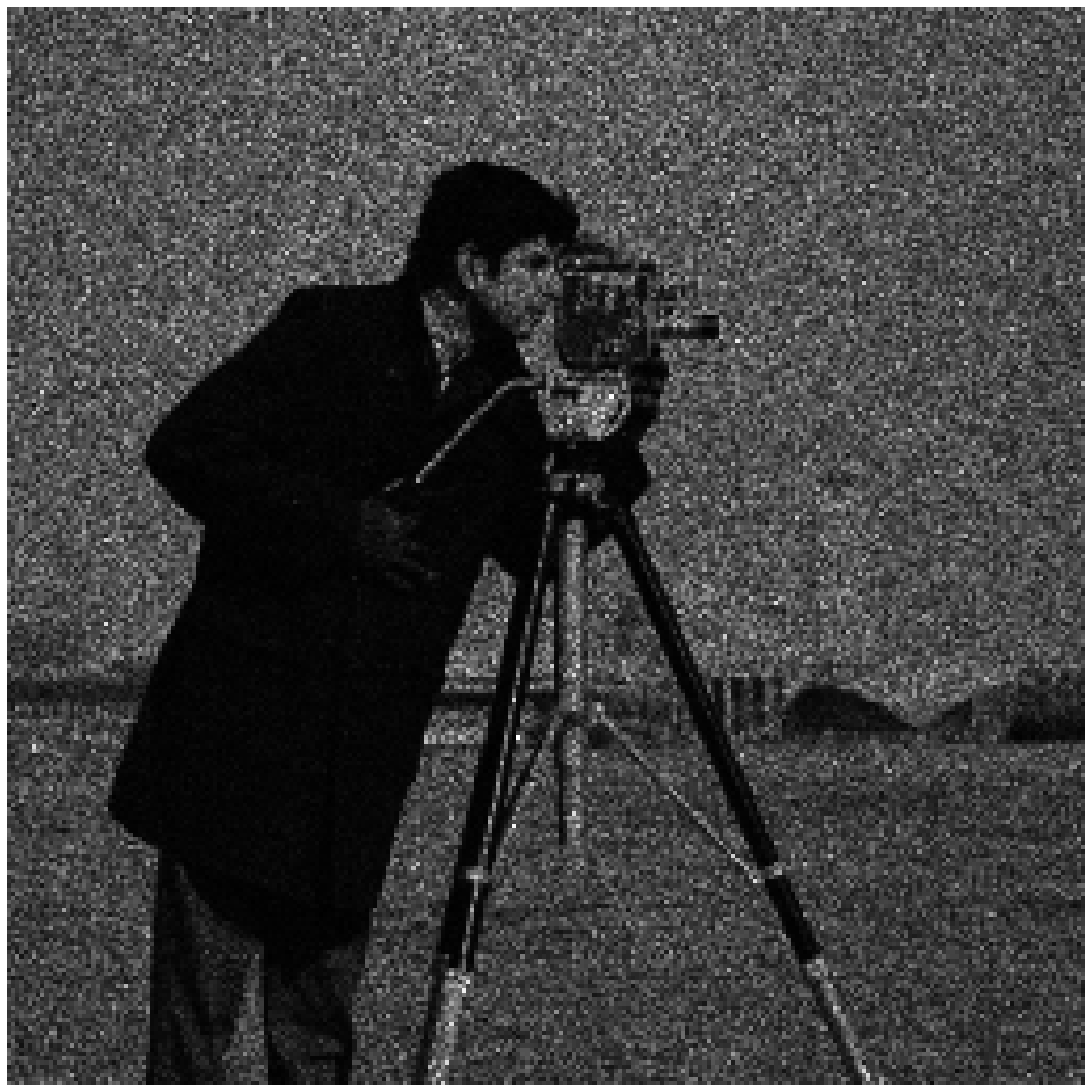}}
  \subfigure[]{
    \label{fig5.5:subfig:b} 
    \includegraphics[width=1.5in,clip]{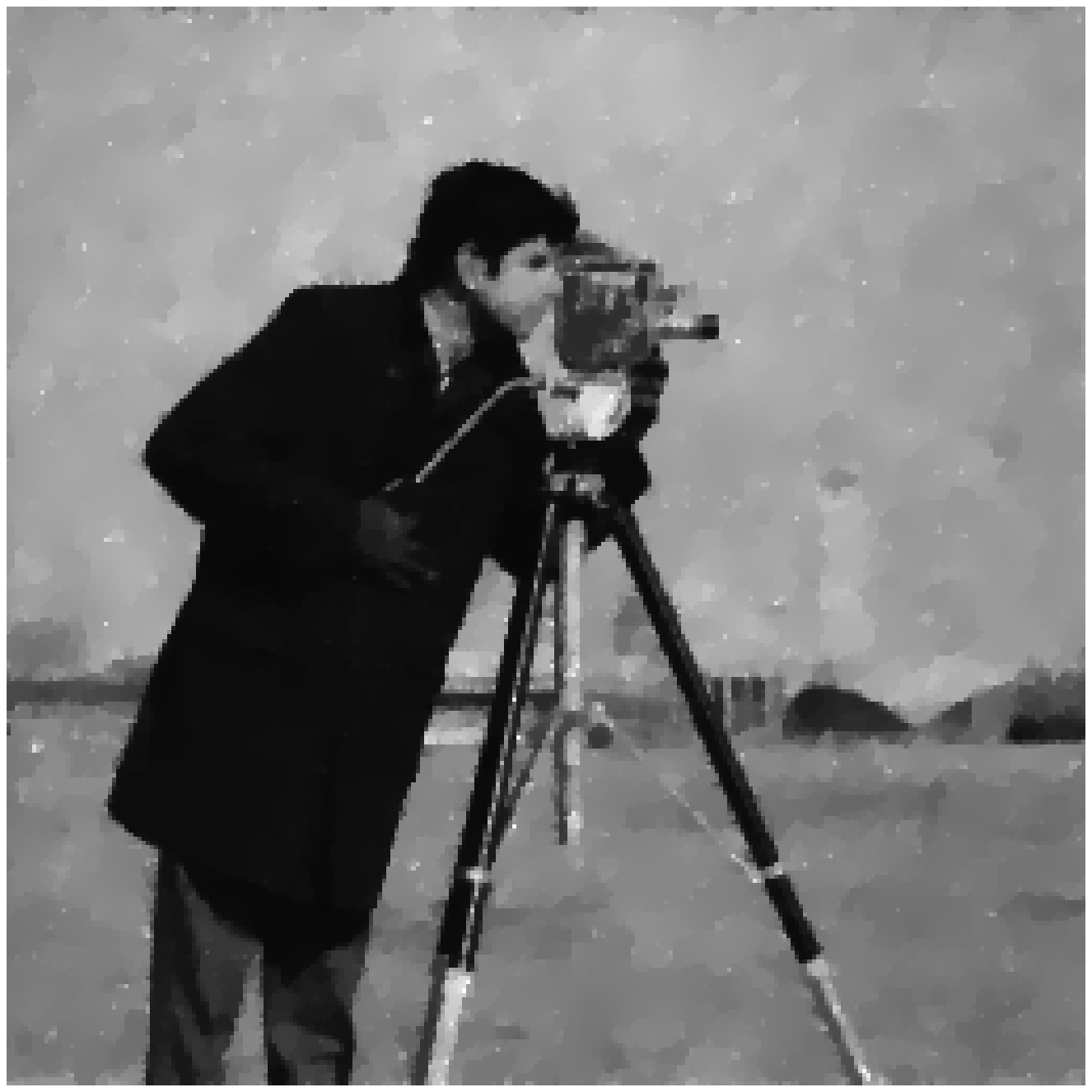}}
  \subfigure[]{
    \label{fig5.5:subfig:c} 
    \includegraphics[width=1.5in,clip]{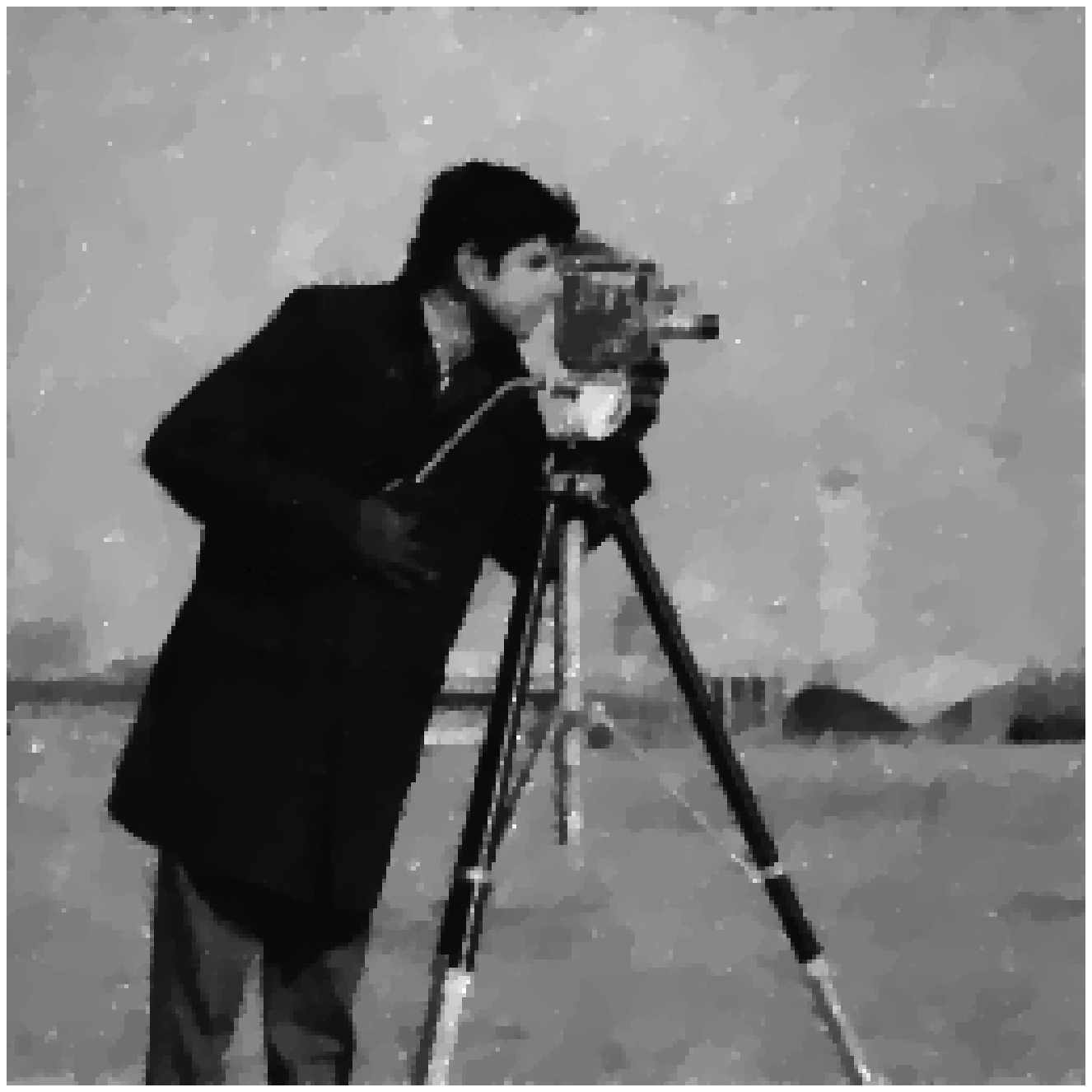}}
  \subfigure[]{
    \label{fig5.5:subfig:d} 
    \includegraphics[width=1.5in,clip]{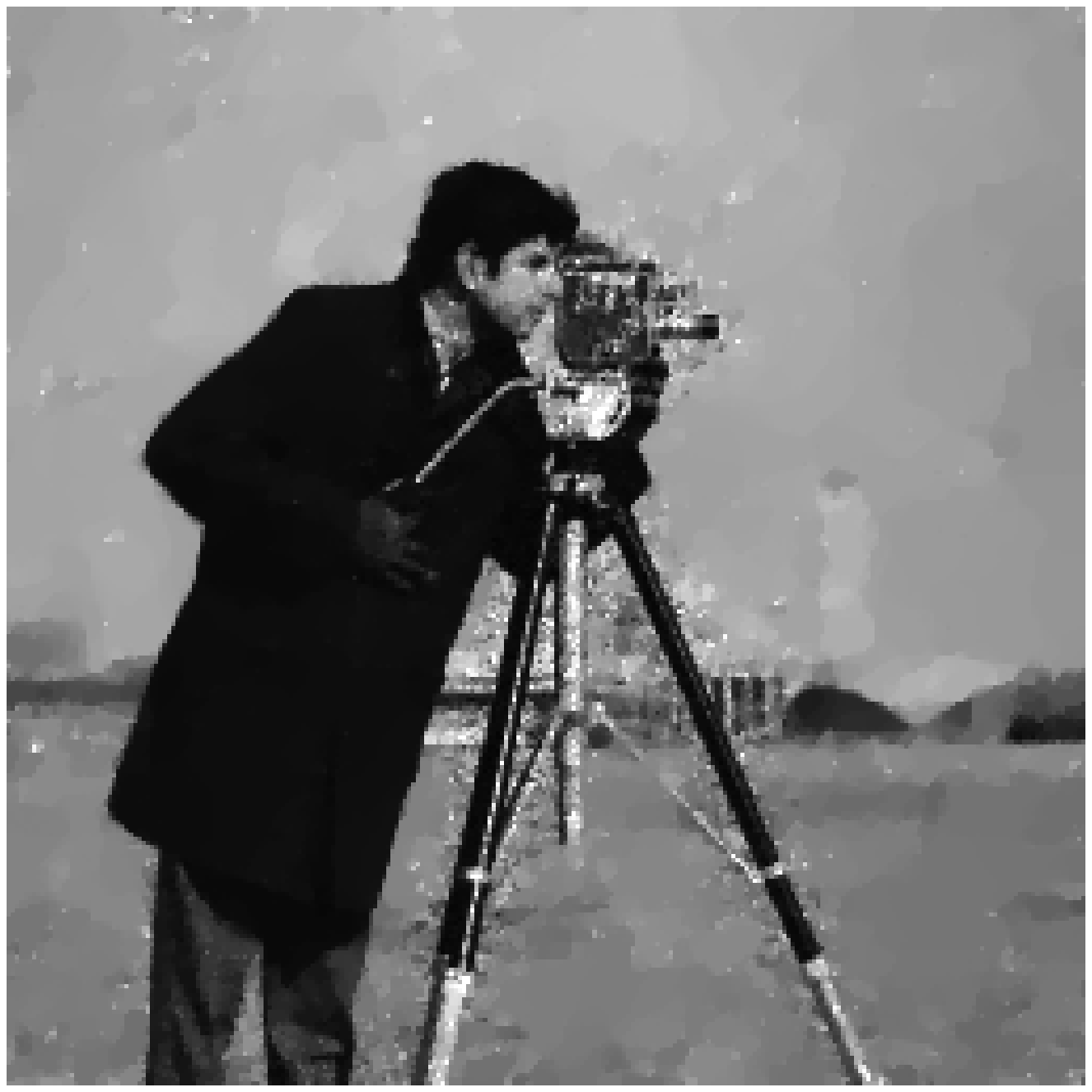}}
  \subfigure[]{
    \label{fig5.5:subfig:e} 
    \includegraphics[width=1.5in,clip]{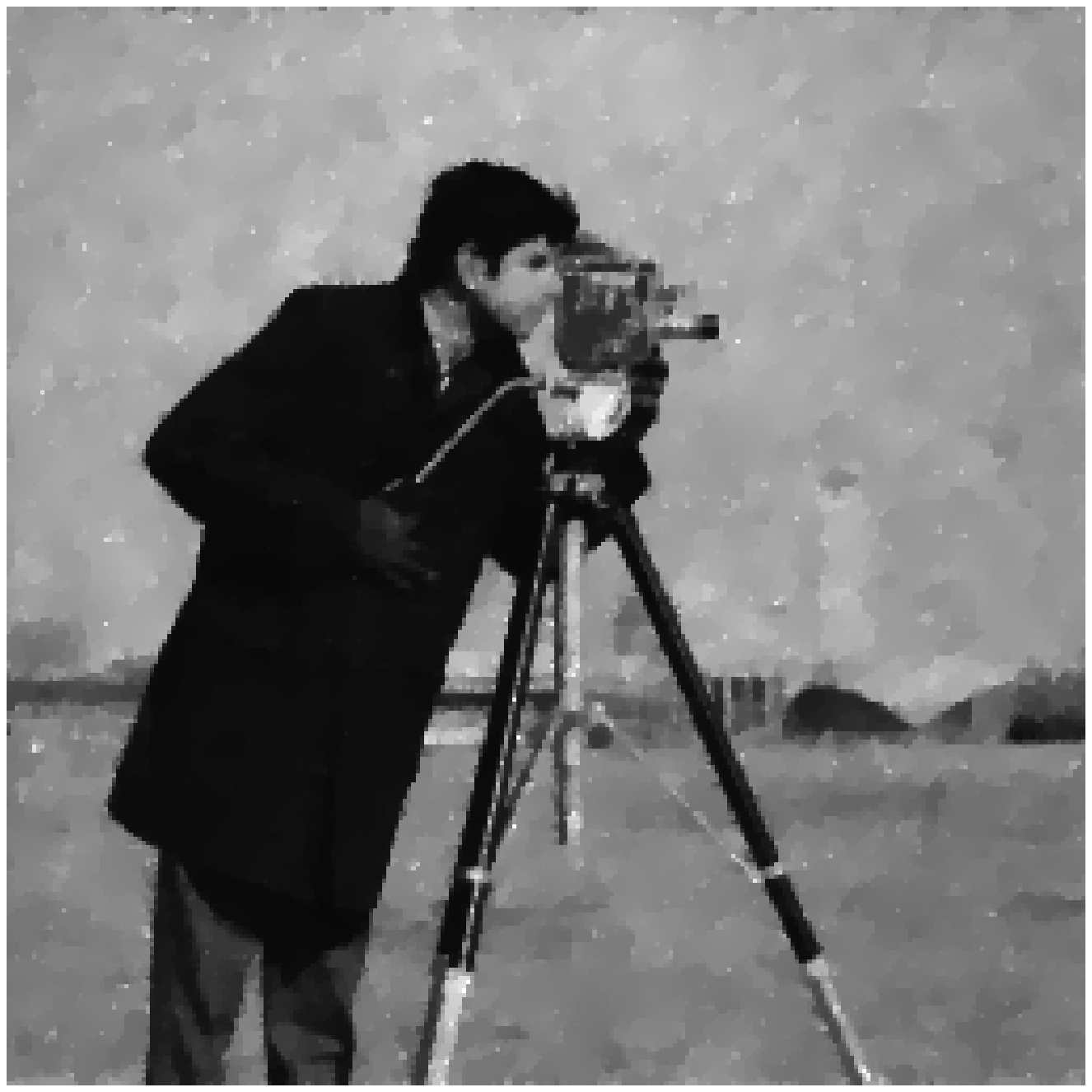}}
  \subfigure[]{
    \label{fig5.5:subfig:f} 
    \includegraphics[width=1.5in,clip]{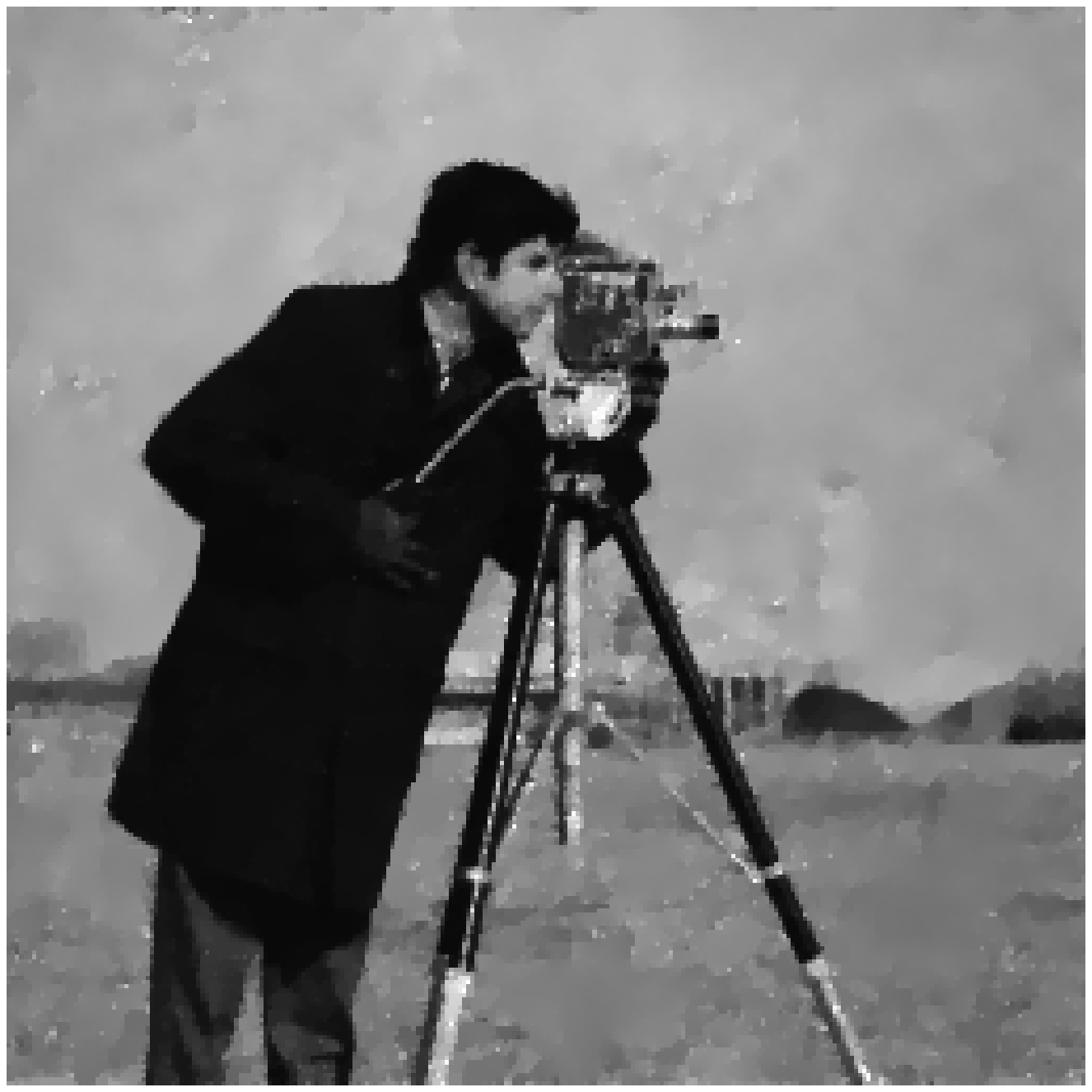}}
\caption{(a)The noisy image with M=10, (b)the image denoised by the
PLAD\_EXP, (c)the image denoised by PLAD\_DIV, (d)the image denoised by
STV\_AL, (e)the image denoised by Algorithm 1, (f)the image denoised by Algorithm 2.
}
\label{fig5.5}
\end{figure}

\begin{figure}
  \centering
  \subfigure[]{
    \label{fig5.6:subfig:a} 
    \includegraphics[width=1.0in,clip]{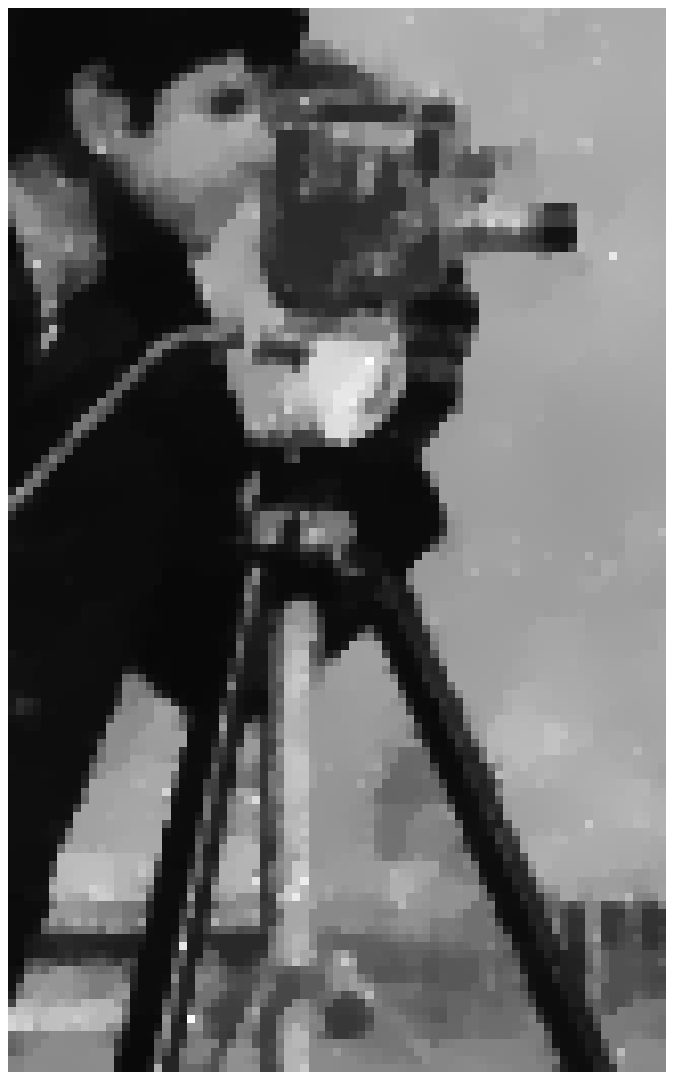}}
  \subfigure[]{
    \label{fig5.6:subfig:b} 
    \includegraphics[width=1.0in,clip]{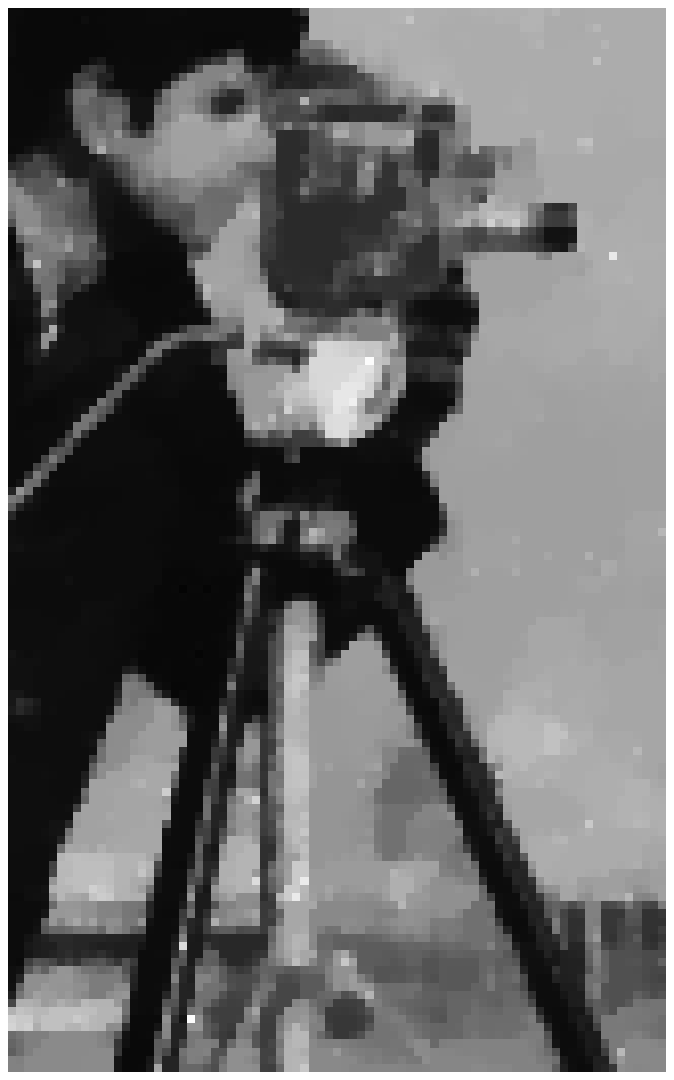}}
  \subfigure[]{
    \label{fig5.6:subfig:c} 
    \includegraphics[width=1.0in,clip]{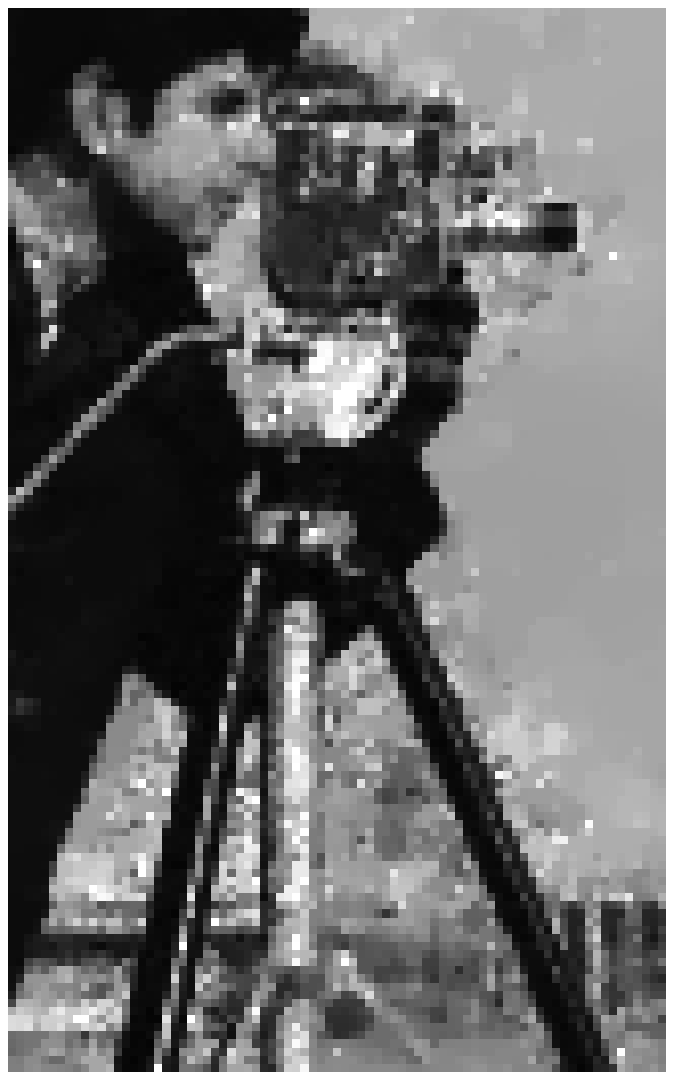}}
  \subfigure[]{
    \label{fig5.6:subfig:d} 
    \includegraphics[width=1.0in,clip]{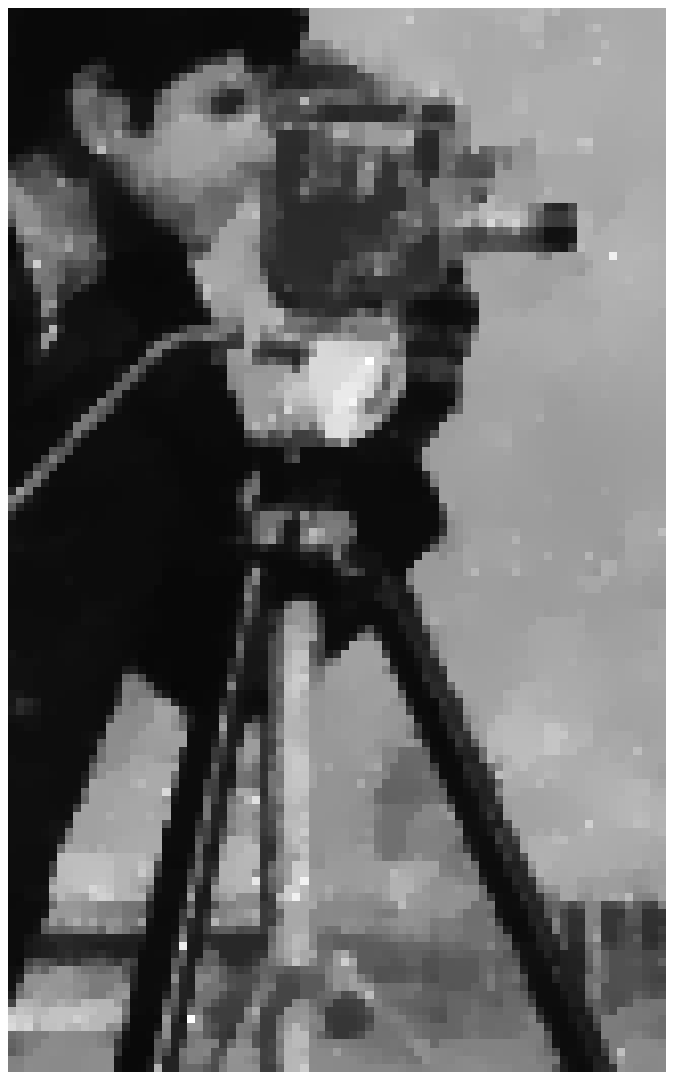}}
  \subfigure[]{
    \label{fig5.6:subfig:e} 
    \includegraphics[width=1.0in,clip]{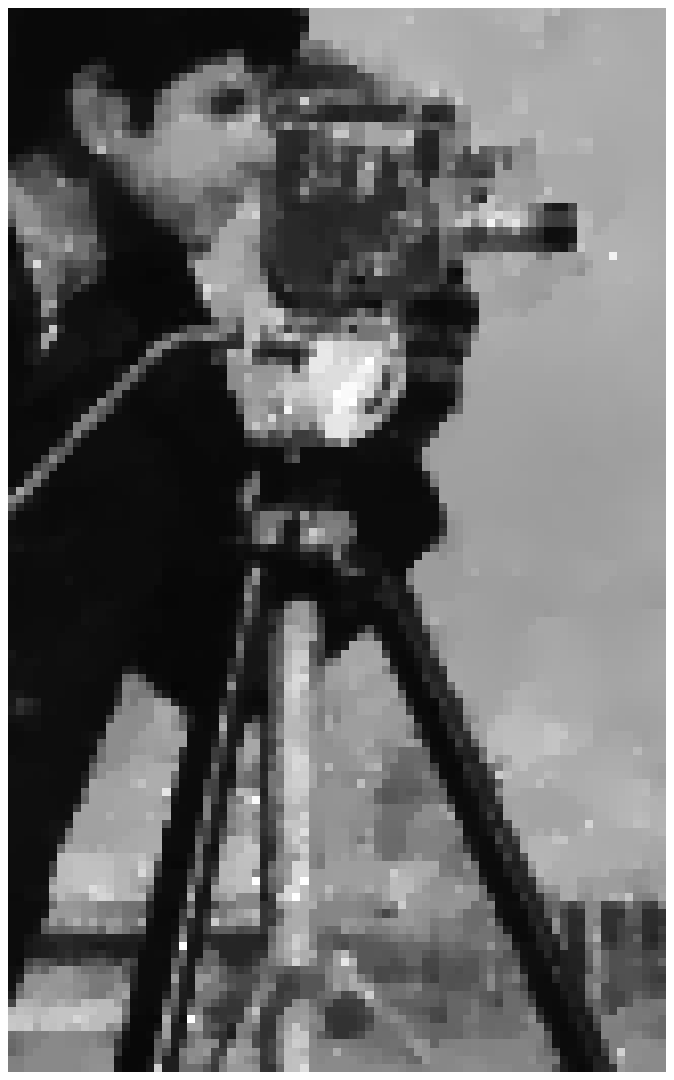}}
\caption{The zoomed version of the denoised images in Fig. \ref{fig5.5}. (a)PLAD\_EXP, (b)PLAD\_DIV, (c)STV\_AL, (d)Algorithm 1. (e)Algorithm 2.
}
\label{fig5.6}
\end{figure}

In Figure \ref{fig5.5}, the noisy and restored images corresponding to the Cameraman image with $M=10$ are shown. The results in Figure \ref{fig5.5}(b)(c)(e) are generated by TV models with scale regularization parameter, and those in Figure \ref{fig5.5}(d)(f) are produced by TV models with spatially-adaptive regularization parameter. Note that in Figure \ref{fig5.5:subfig:f} fine details such as the camera and the tripod are sharper as compared to \ref{fig5.5}(b)(c)(e) due to the use of the spatially varying parameter. Meanwhile, we also observe that some artificial effects appear in the edges of the camera and the tripod of Figure \ref{fig5.5:subfig:d} owing to the over-fitting. However, it does not exist in Figure \ref{fig5.5:subfig:f}. In order to make the comparison clearer, we zoom into certain regions of these results in Figure \ref{fig5.6}.

\begin{figure}
  \centering
  \subfigure[]{
    \label{fig5.7:subfig:a} 
    \includegraphics[width=1.5in,clip]{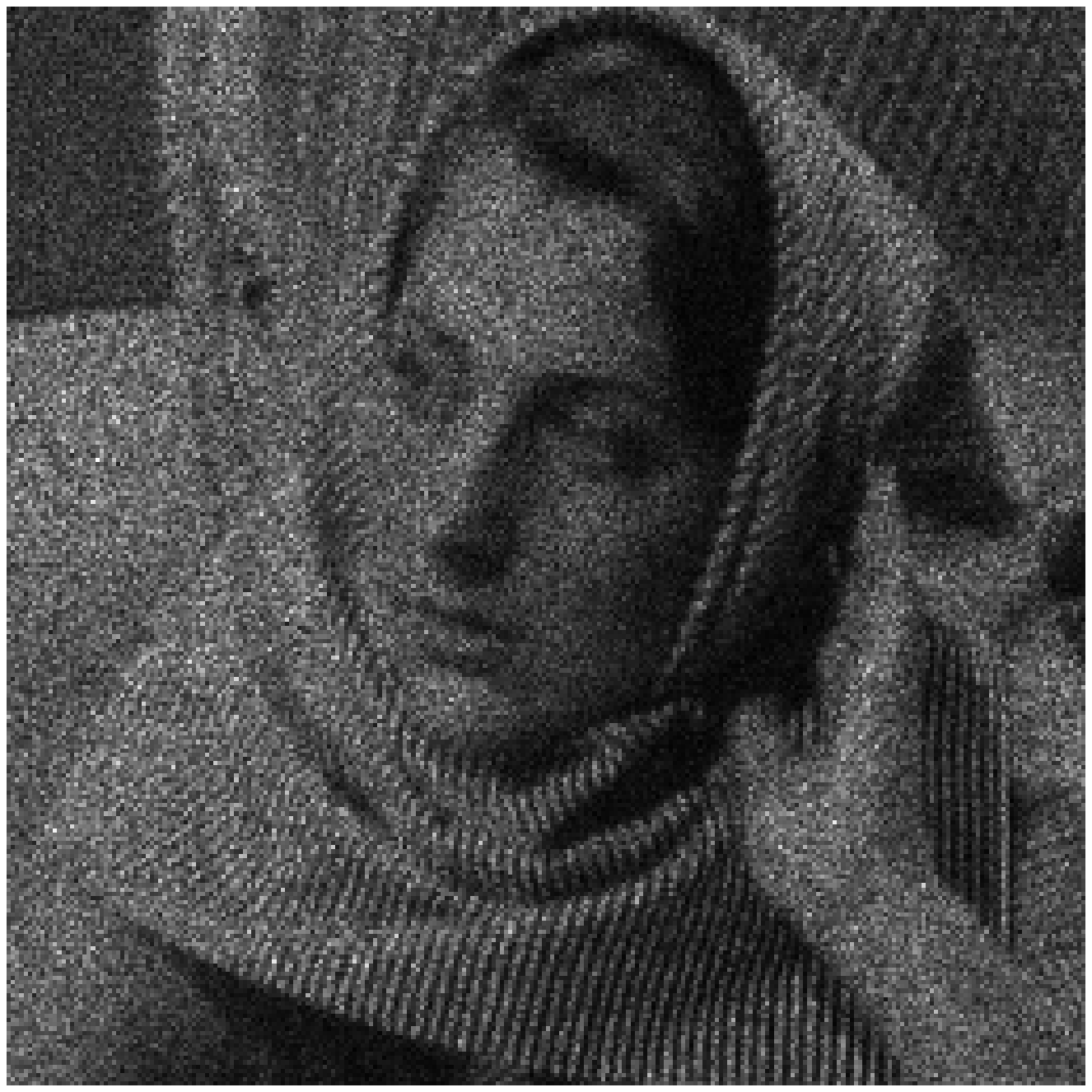}}
  \subfigure[]{
    \label{fig5.7:subfig:b} 
    \includegraphics[width=1.5in,clip]{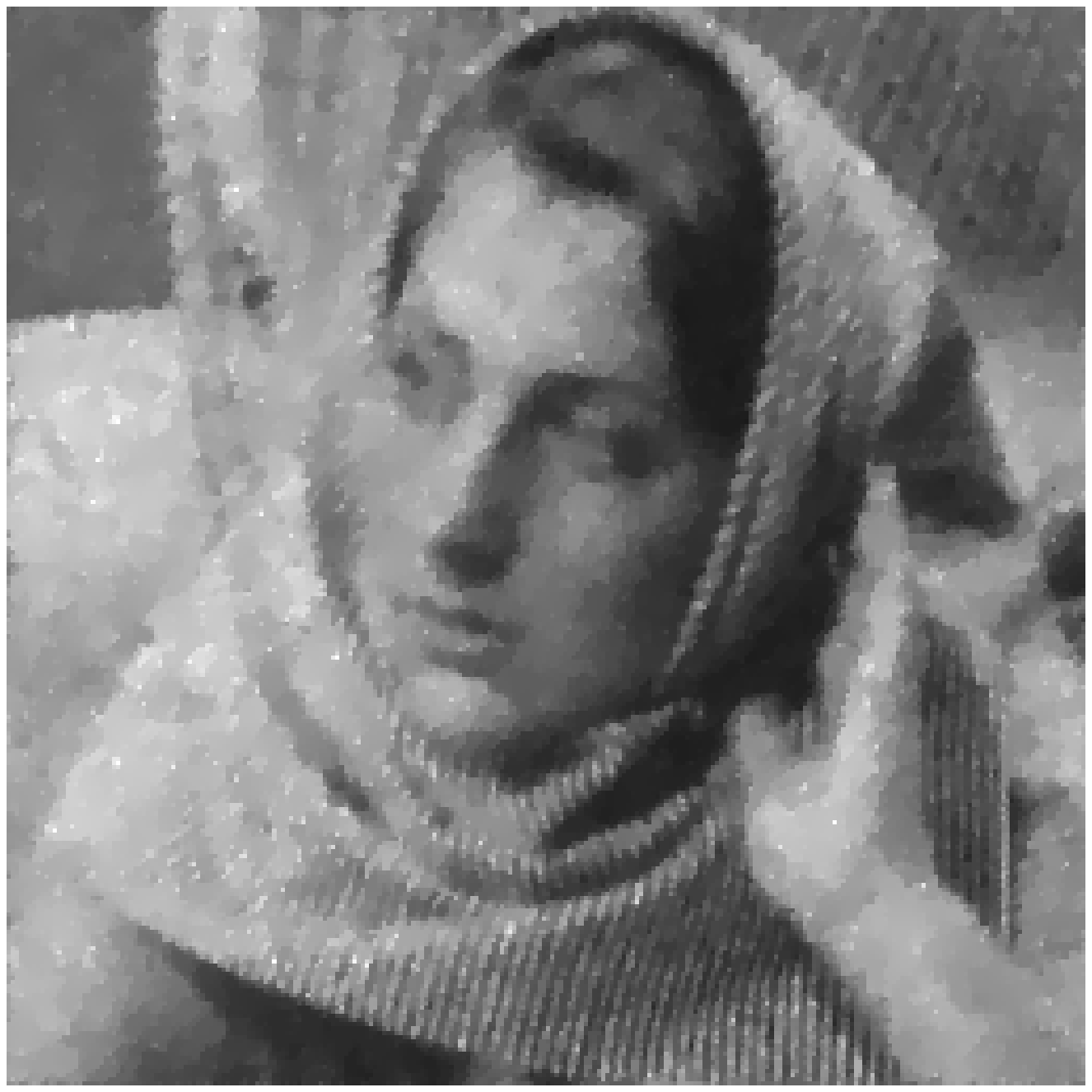}}
  \subfigure[]{
    \label{fig5.7:subfig:c} 
    \includegraphics[width=1.5in,clip]{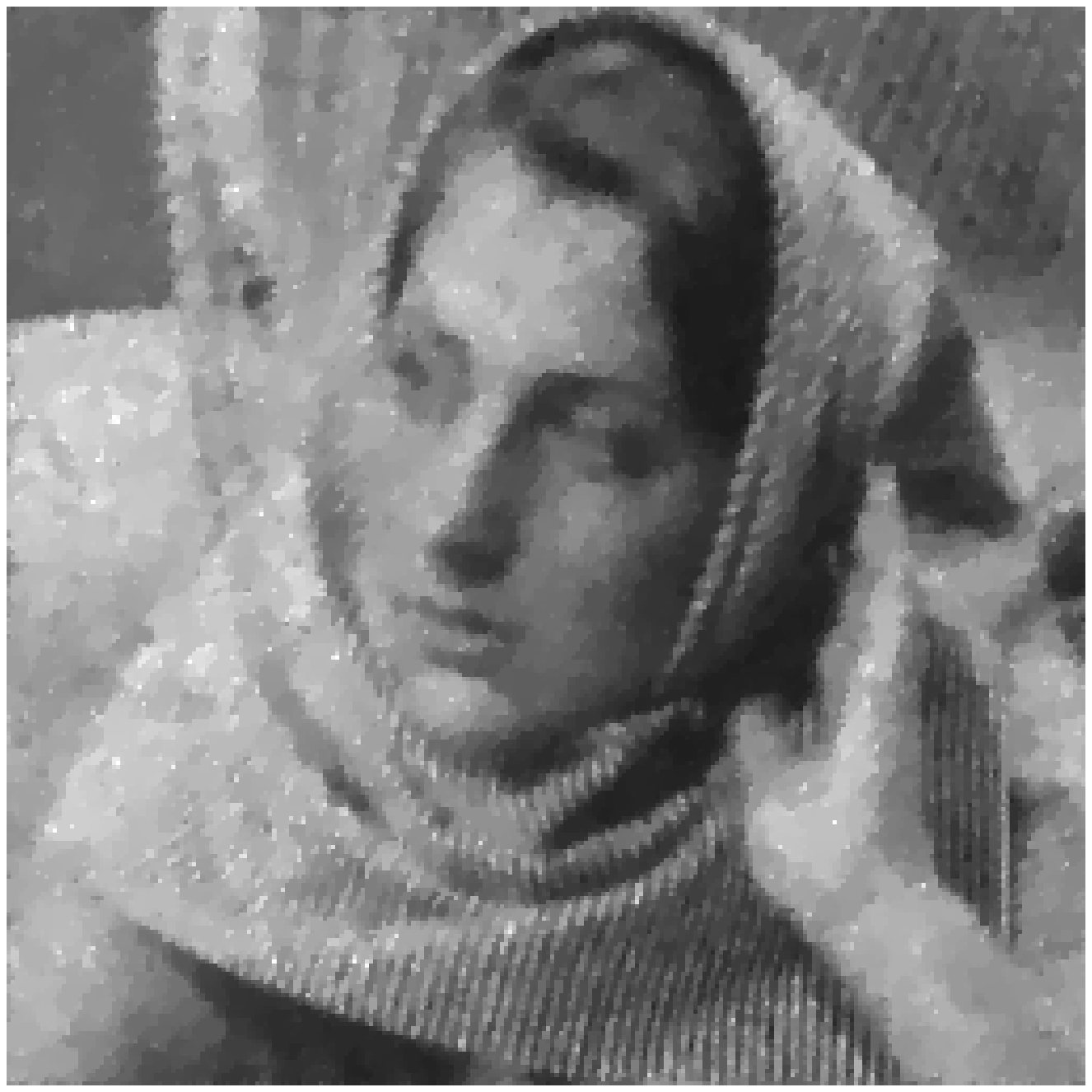}}
  \subfigure[]{
    \label{fig5.7:subfig:d} 
    \includegraphics[width=1.5in,clip]{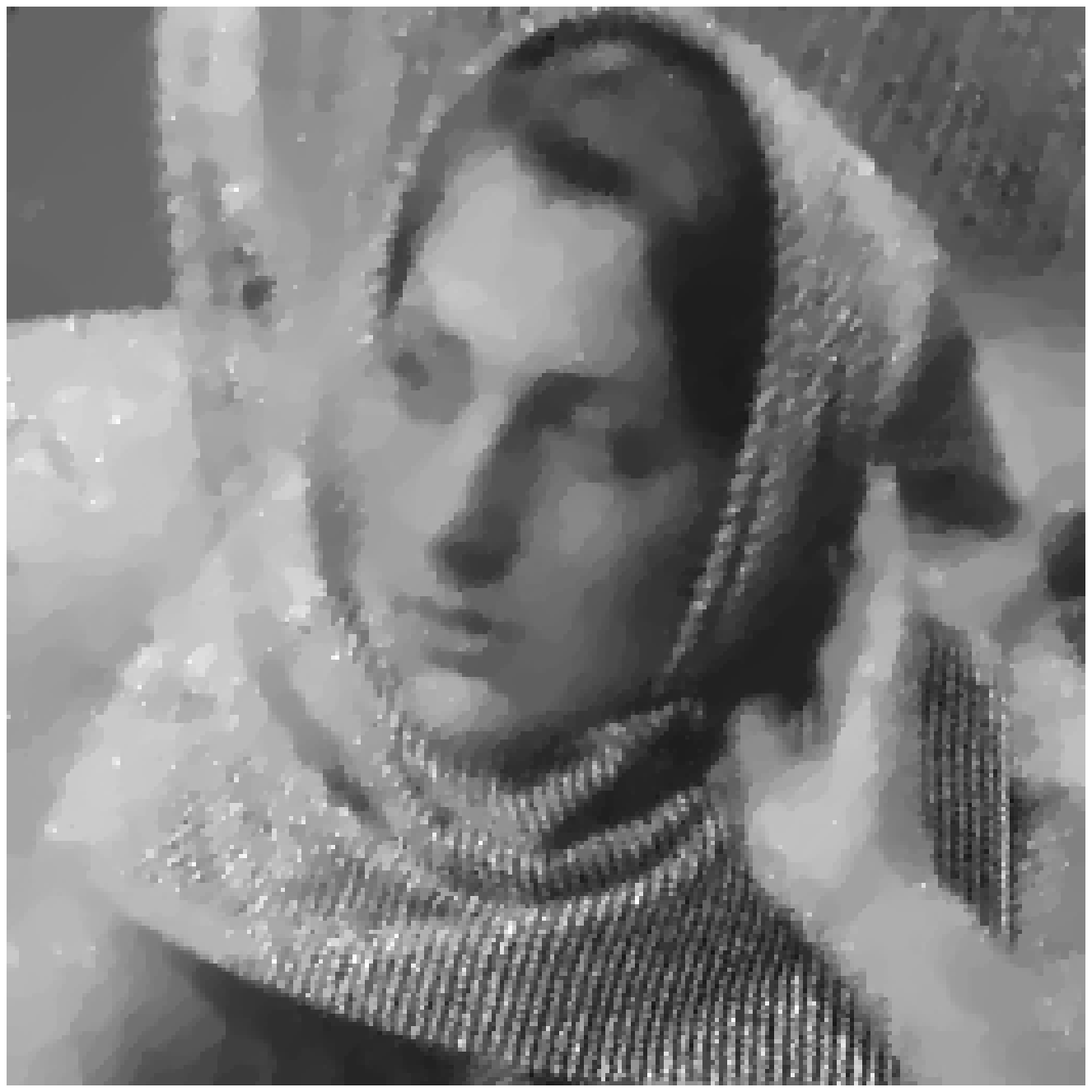}}
  \subfigure[]{
    \label{fig5.7:subfig:e} 
    \includegraphics[width=1.5in,clip]{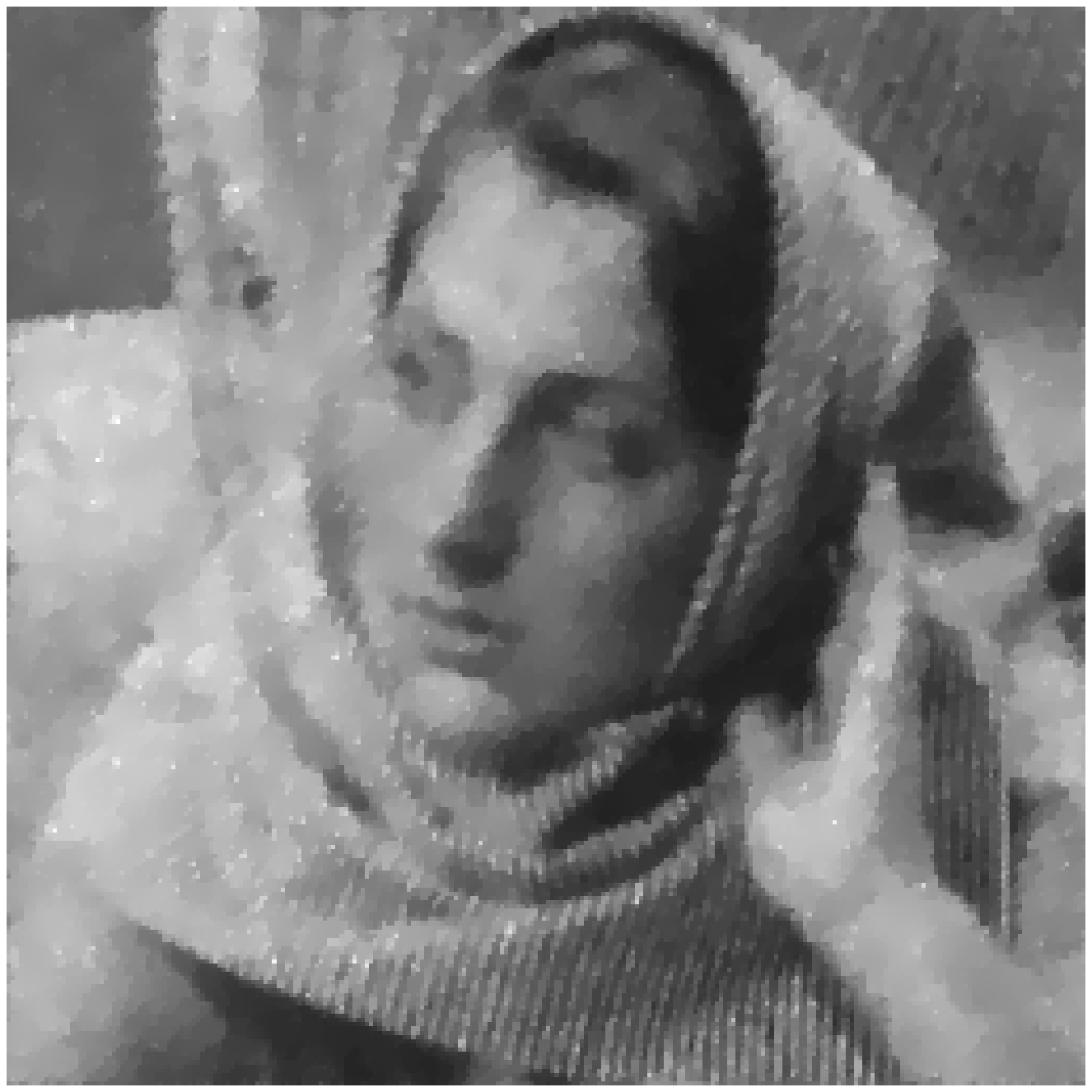}}
  \subfigure[]{
    \label{fig5.7:subfig:f} 
    \includegraphics[width=1.5in,clip]{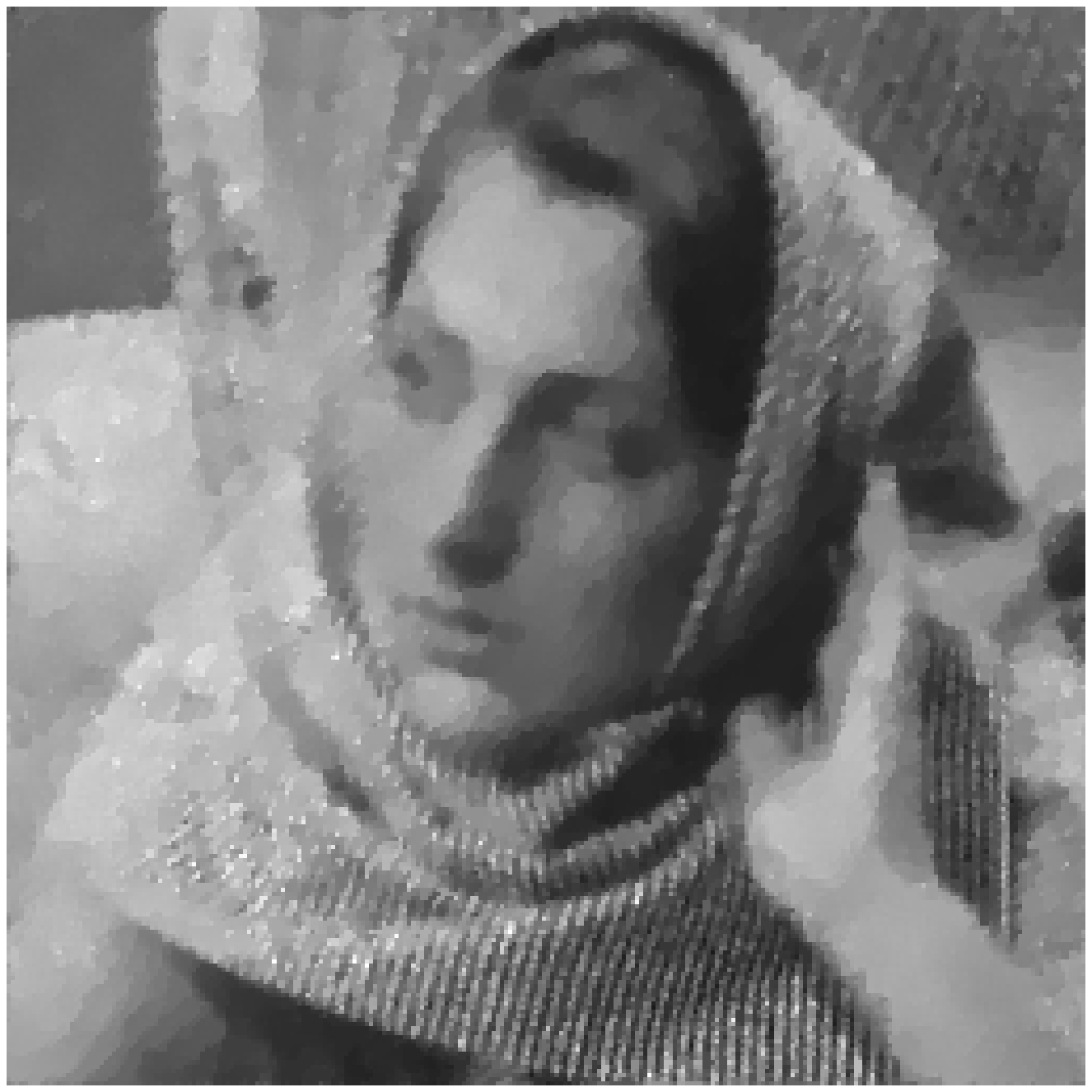}}
\caption{(a)The noisy image with M=15, (b)the image denoised by the
PLAD\_EXP, (c)the image denoised by PLAD\_DIV, (d)the image denoised by
STV\_AL, (e)the image denoised by Algorithm 1, (f)the image denoised by Algorithm 2.
}
\label{fig5.7}
\end{figure}

\begin{figure}
  \centering
  \subfigure[]{
    \label{fig5.8:subfig:b} 
    \includegraphics[width=1.0in,clip]{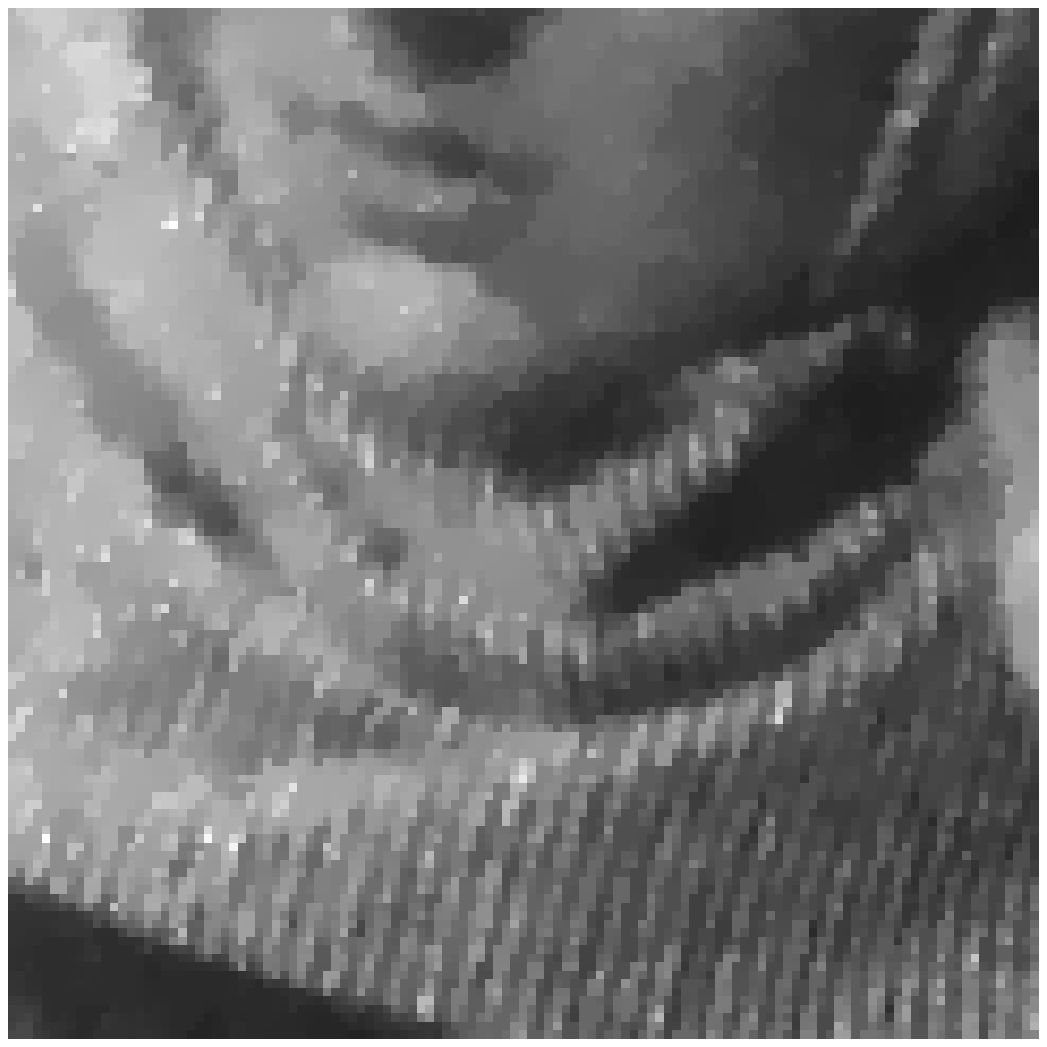}}
  \subfigure[]{
    \label{fig5.8:subfig:c} 
    \includegraphics[width=1.0in,clip]{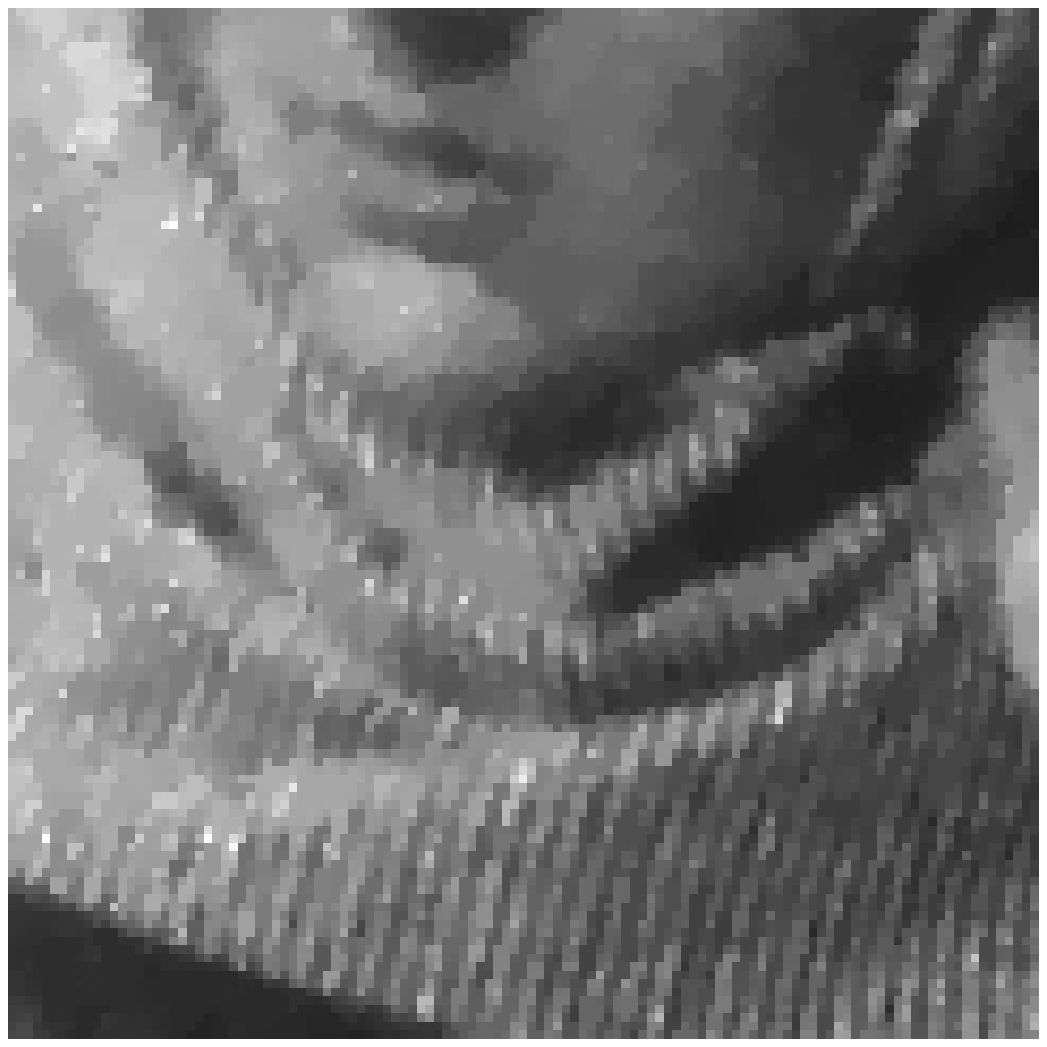}}
  \subfigure[]{
    \label{fig5.8:subfig:d} 
    \includegraphics[width=1.0in,clip]{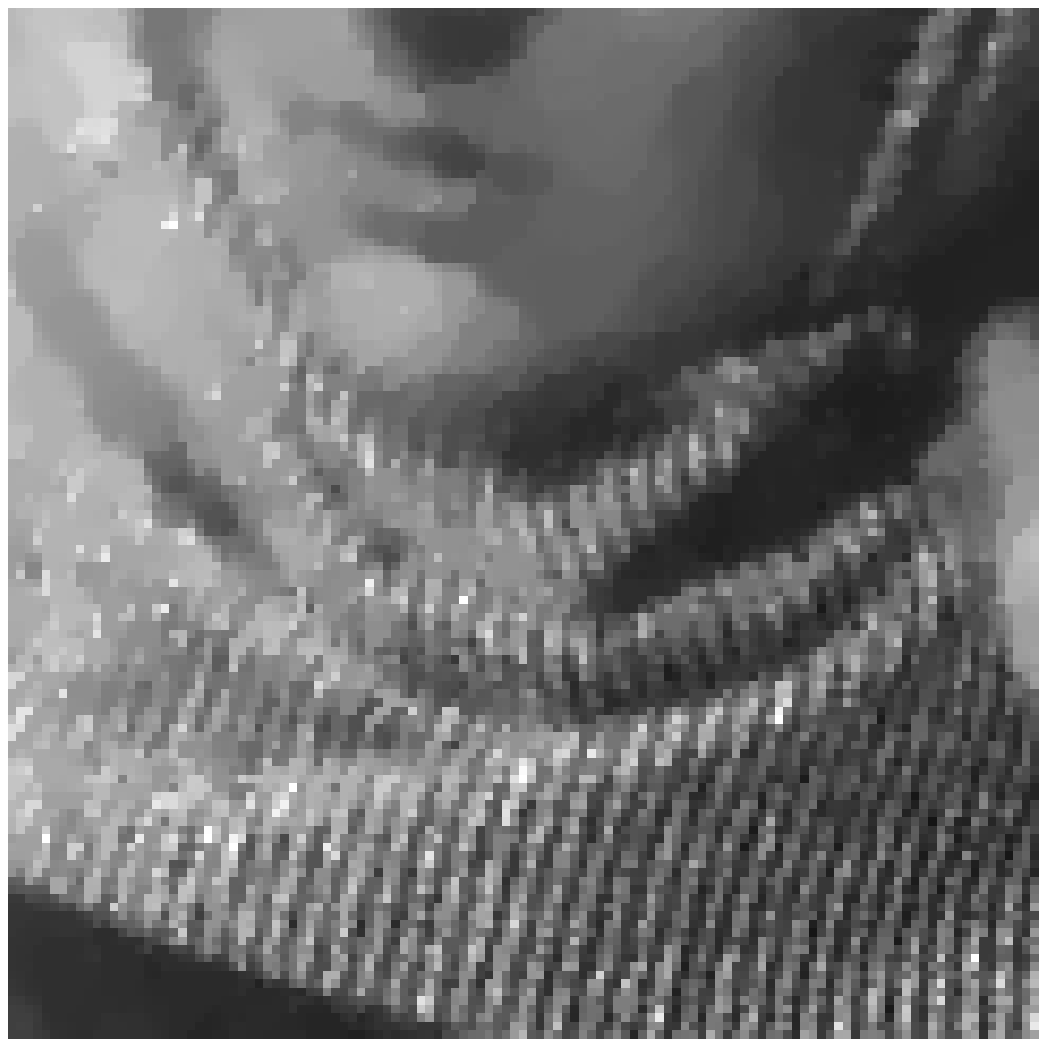}}
  \subfigure[]{
    \label{fig5.8:subfig:e} 
    \includegraphics[width=1.0in,clip]{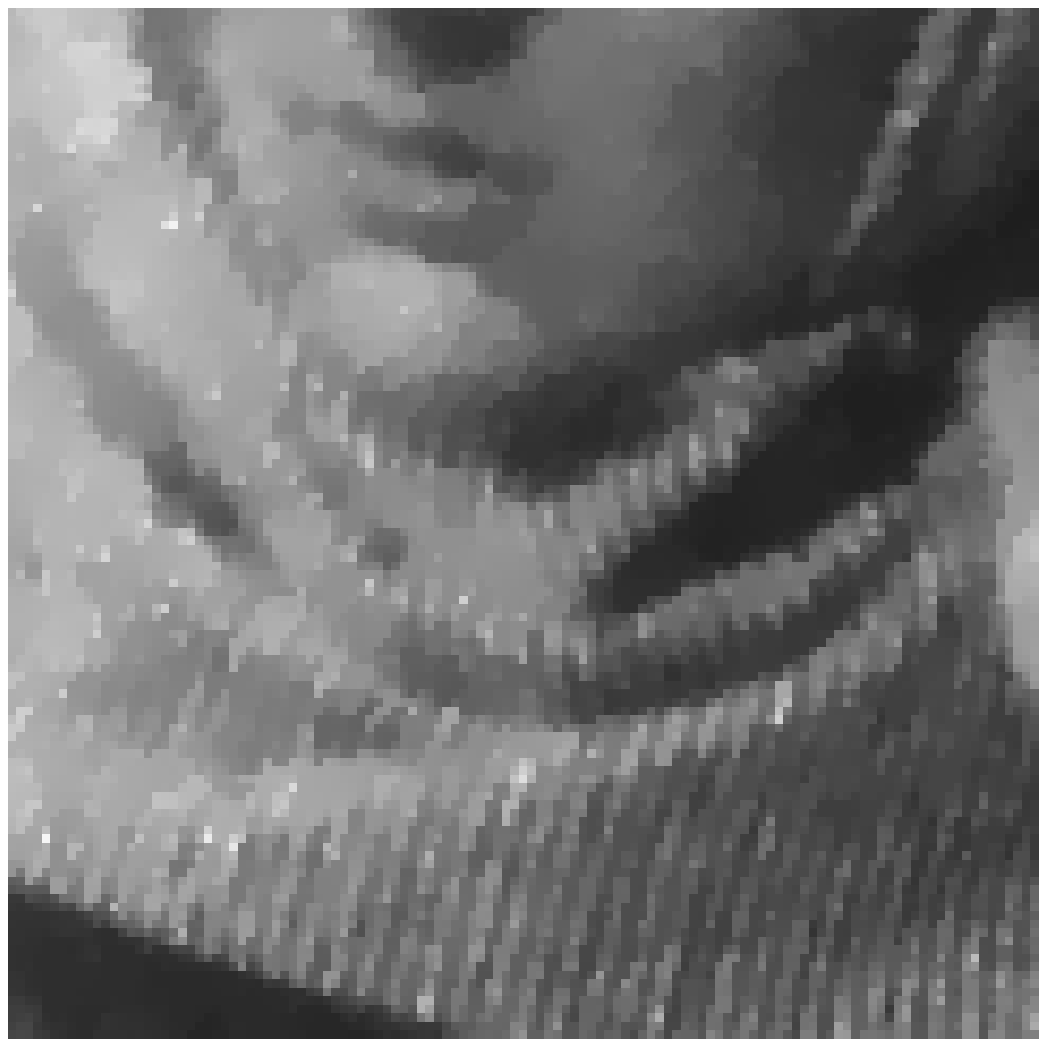}}
  \subfigure[]{
    \label{fig5.8:subfig:e} 
    \includegraphics[width=1.0in,clip]{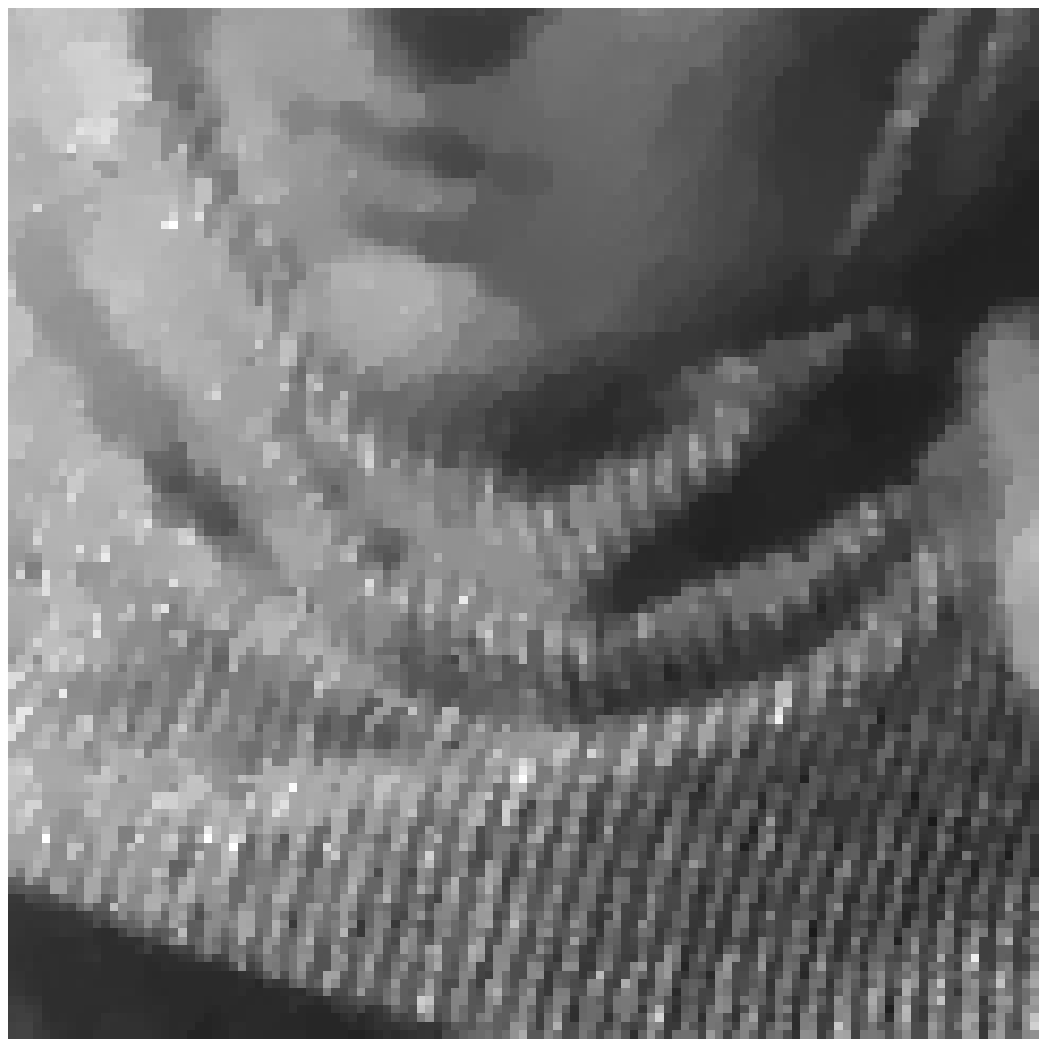}}
\caption{The zoomed version of the denoised images in Fig. \ref{fig5.7}. (a)PLAD\_EXP, (b)PLAD\_DIV, (c)STV\_AL, (d)Algorithm 1. (e)Algorithm 2.
}
\label{fig5.8}
\end{figure}

\begin{figure}
  \centering
  \subfigure[]{
    \label{fig5.9:subfig:a} 
    \includegraphics[width=1.5in,clip]{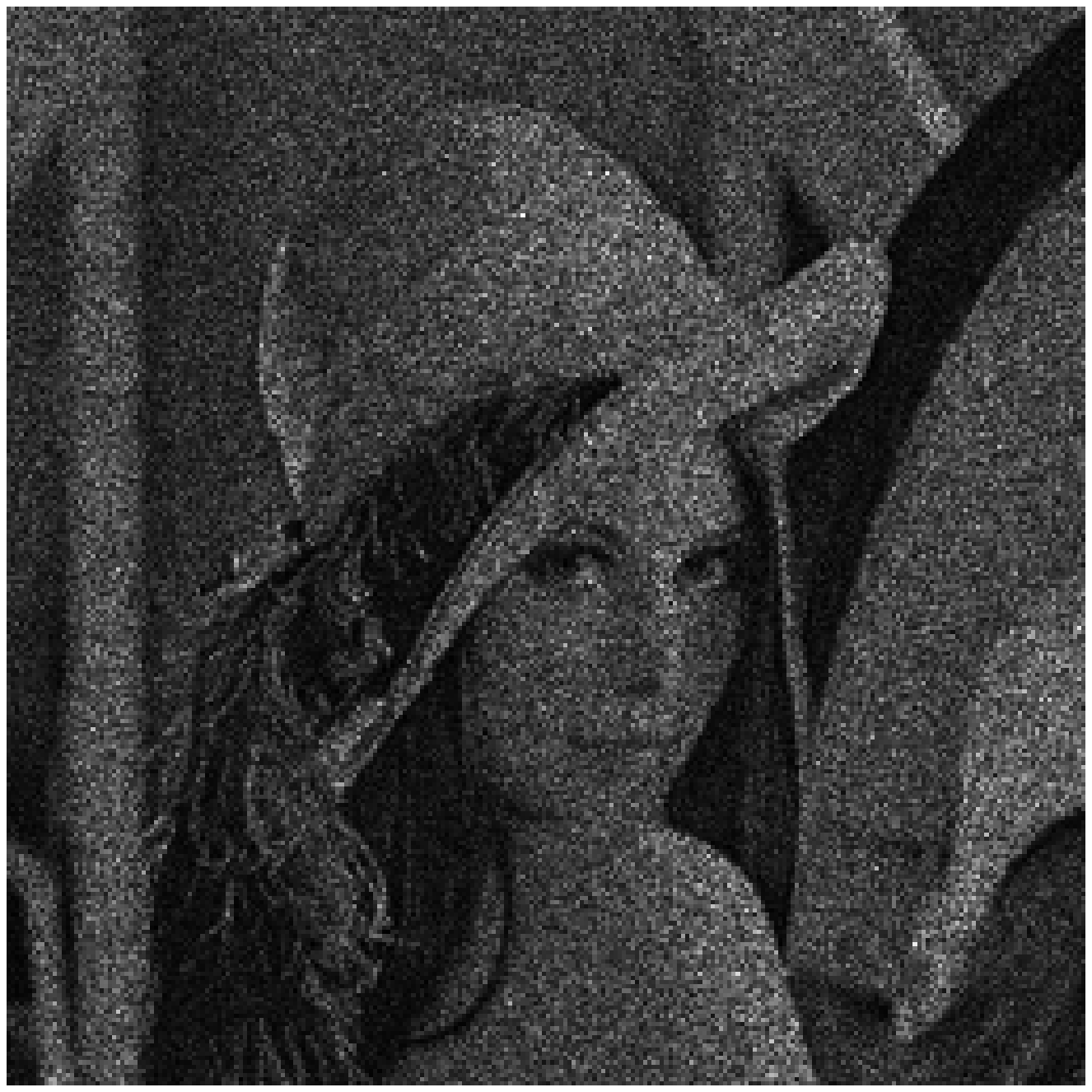}}
  \subfigure[]{
    \label{fig5.9:subfig:b} 
    \includegraphics[width=1.5in,clip]{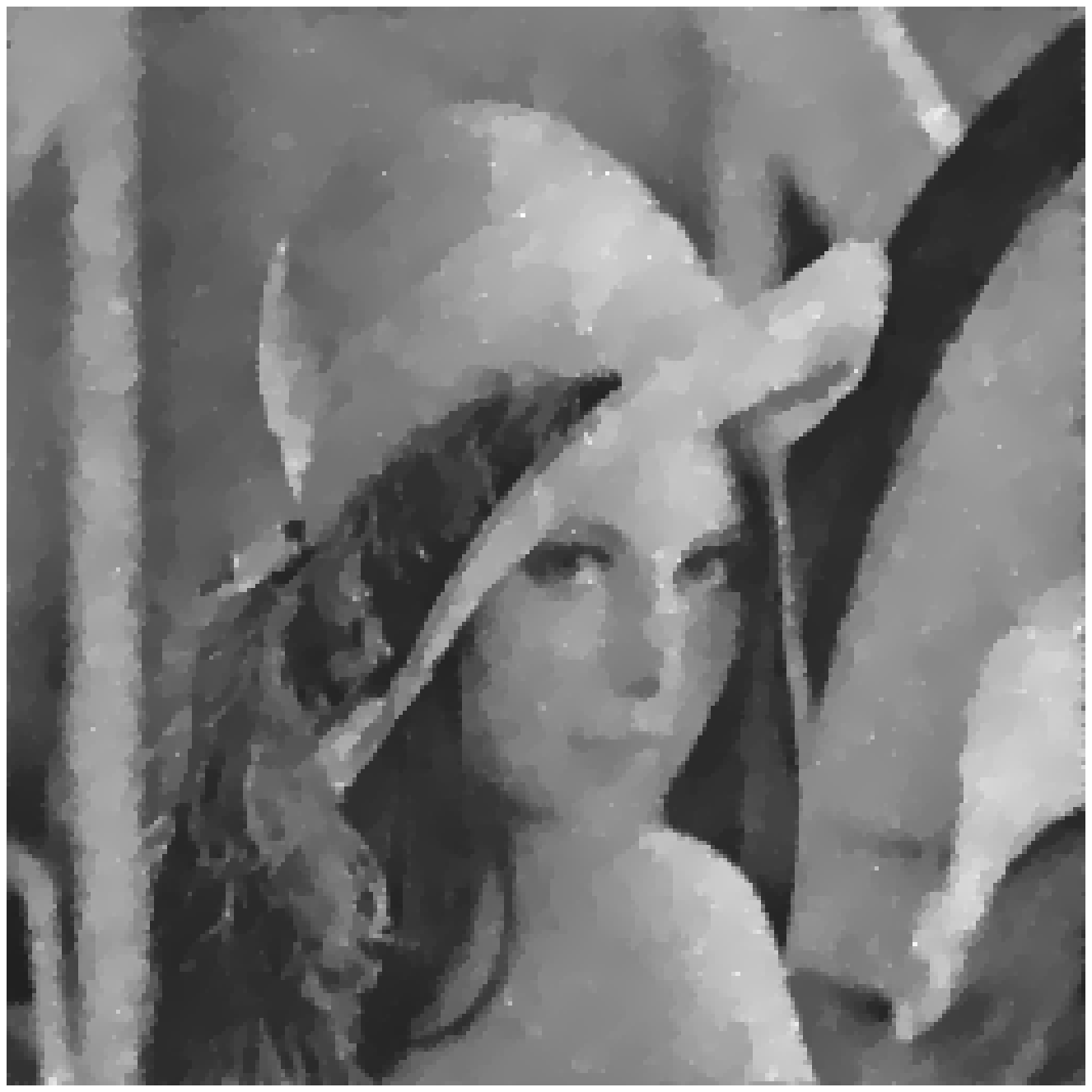}}
  \subfigure[]{
    \label{fig5.9:subfig:c} 
    \includegraphics[width=1.5in,clip]{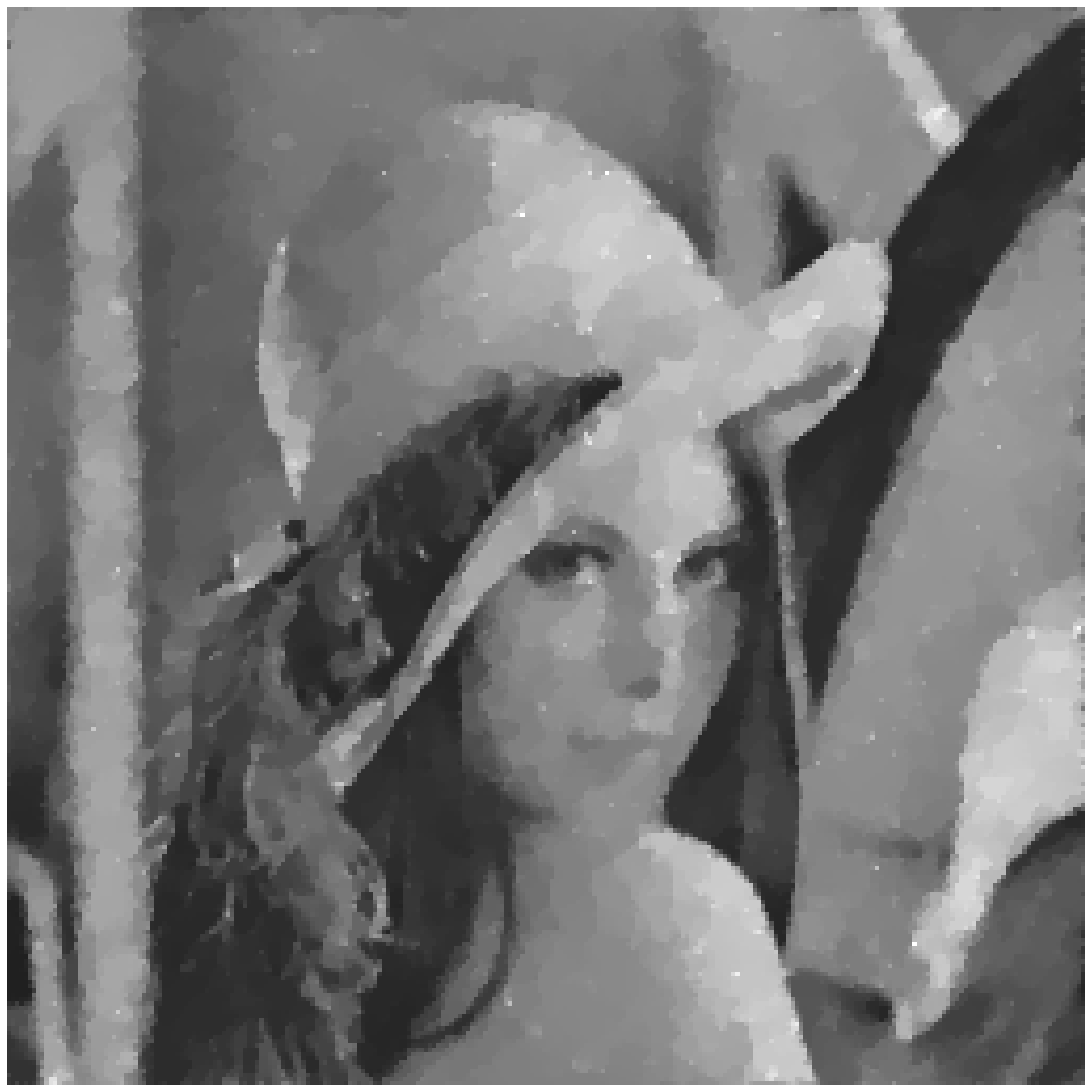}}
  \subfigure[]{
    \label{fig5.9:subfig:d} 
    \includegraphics[width=1.5in,clip]{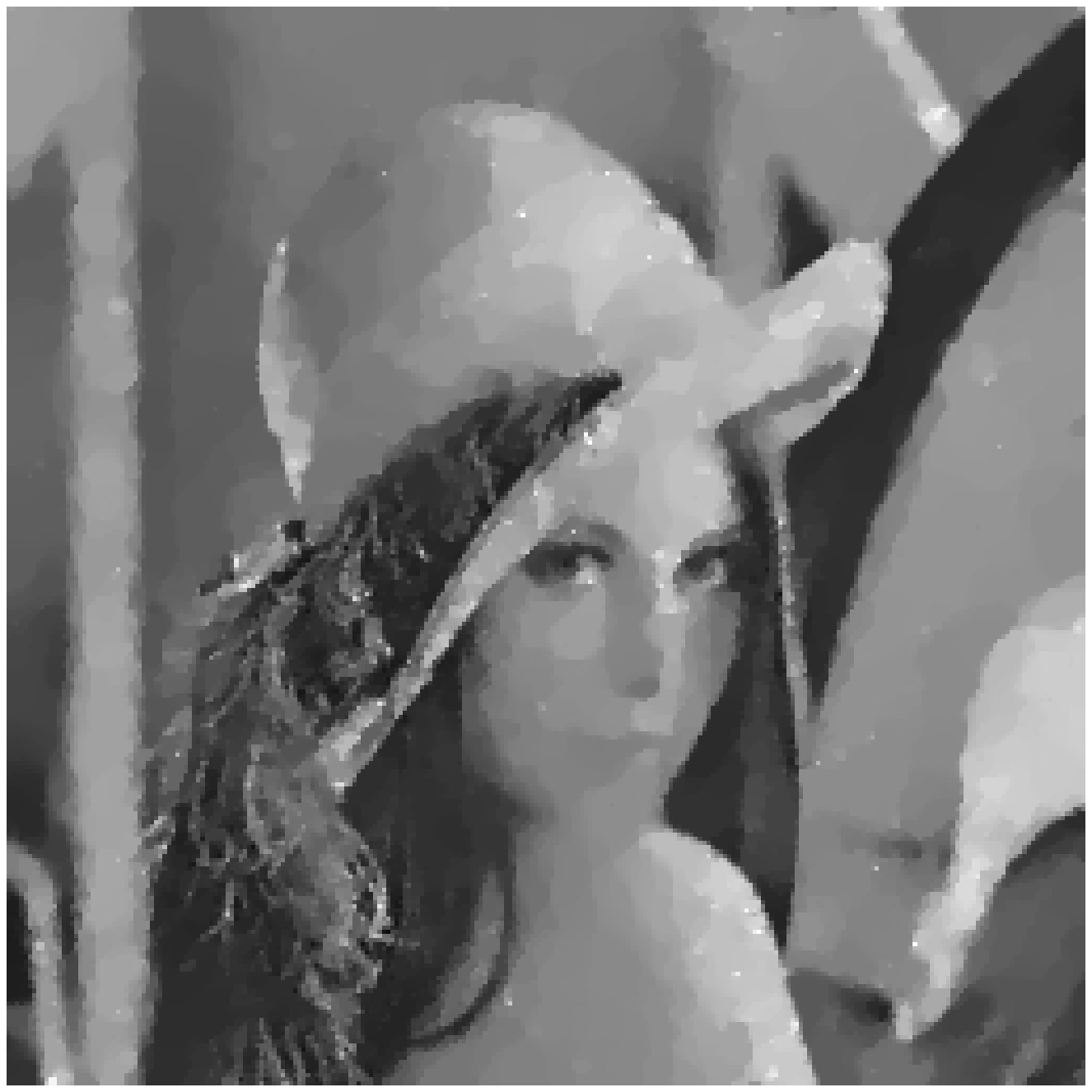}}
  \subfigure[]{
    \label{fig5.9:subfig:e} 
    \includegraphics[width=1.5in,clip]{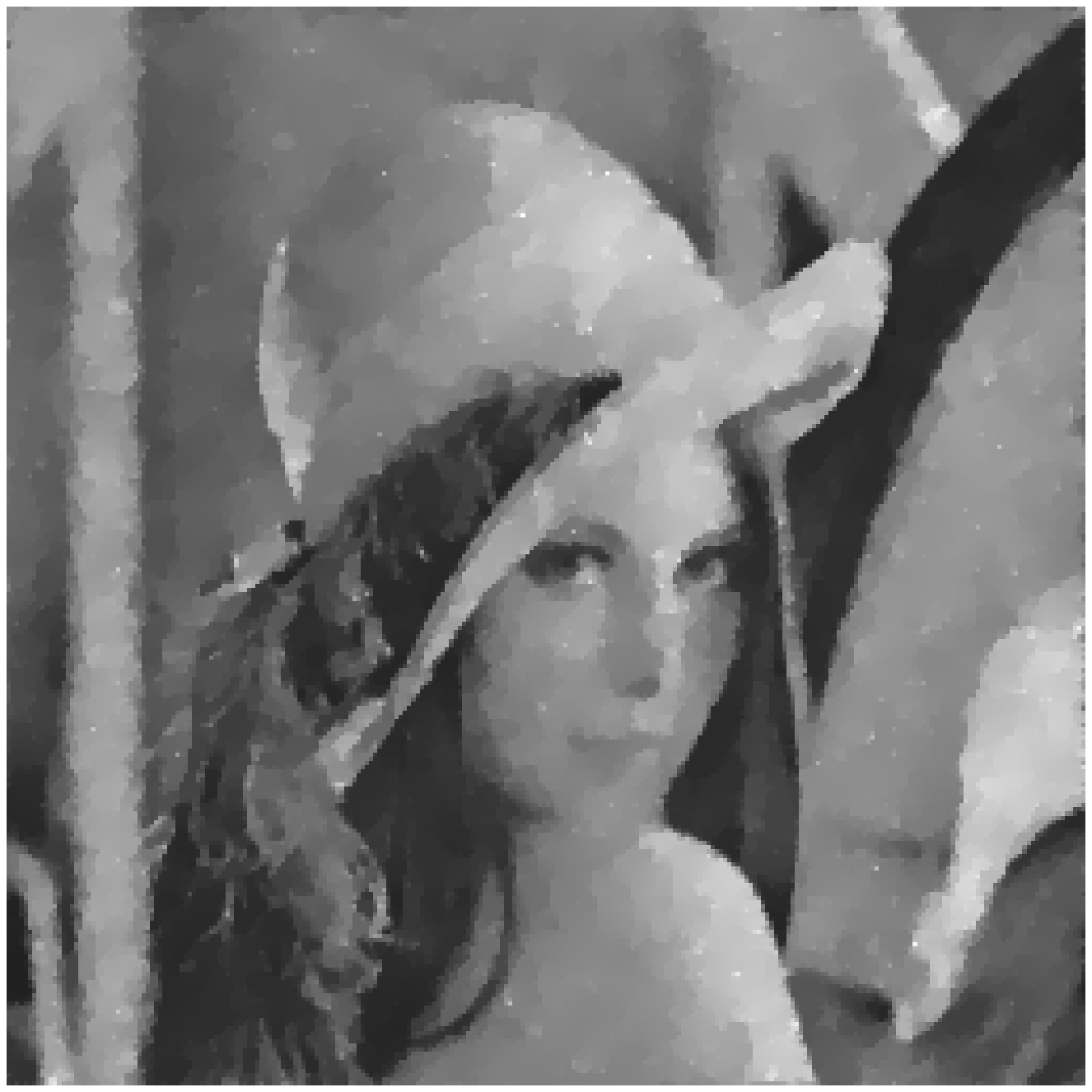}}
  \subfigure[]{
    \label{fig5.9:subfig:f} 
    \includegraphics[width=1.5in,clip]{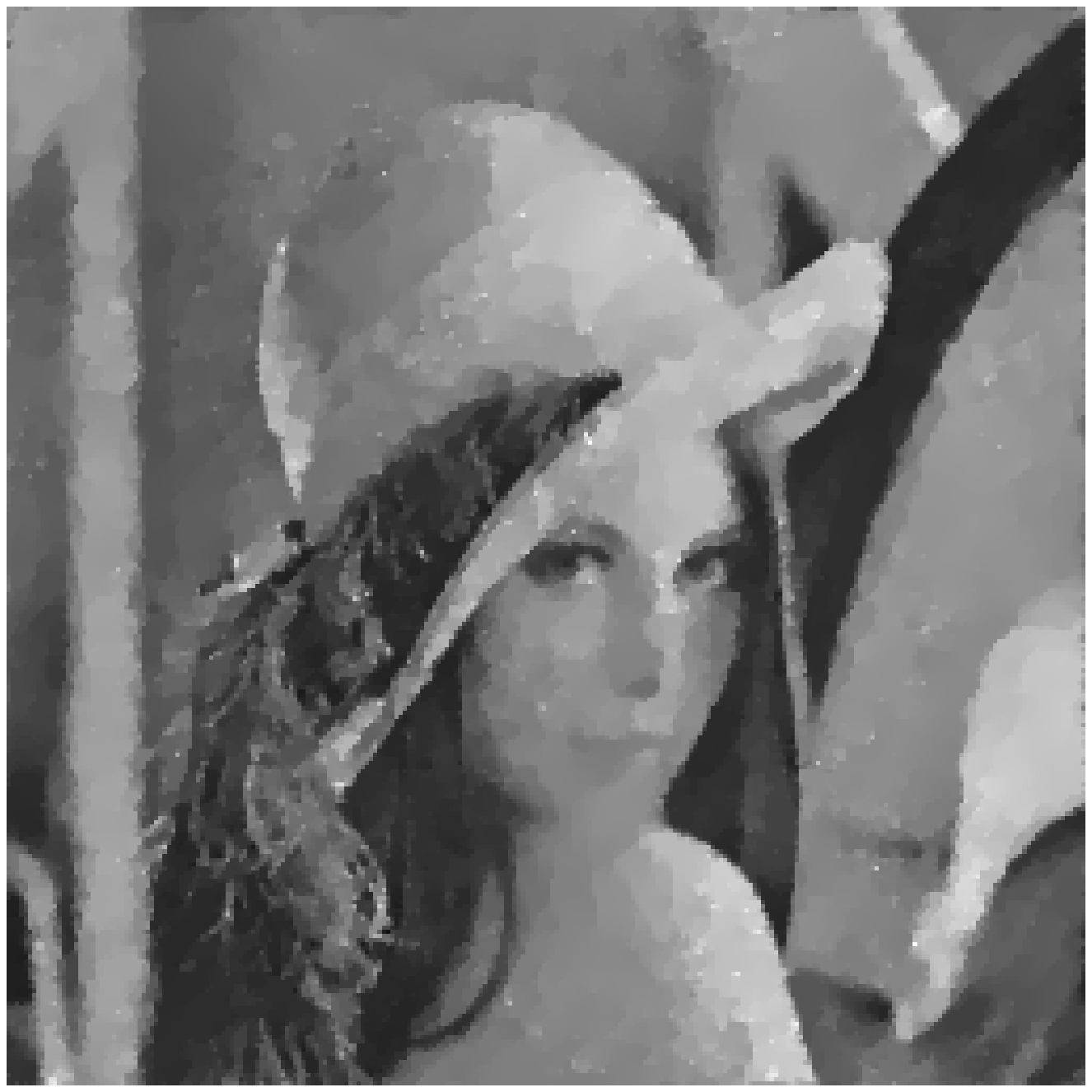}}
\caption{(a)The noisy image with M=10, (b)the image denoised by the
PLAD\_EXP, (c)the image denoised by PLAD\_DIV, (d)the image denoised by
STV\_AL, (e)the image denoised by Algorithm 1, (f)the image denoised by Algorithm 2.
}
\label{fig5.9}
\end{figure}

Figures \ref{fig5.7} and \ref{fig5.9} show the denoised results of the Barbara image with $M=15$ and the Lena image with $M=10$ respectively. In Figure \ref{fig5.7}, we observe that the details in the scarf are sharper in \ref{fig5.7:subfig:d} and \ref{fig5.7:subfig:f} than in \ref{fig5.7:subfig:b}, \ref{fig5.7:subfig:c} and \ref{fig5.7:subfig:e}. Meanwhile, the noise spots in the background are removed sufficiently in \ref{fig5.7:subfig:d} and \ref{fig5.7:subfig:f}, and not in the other results. Part of the texture regions of these results are zoomed in
Figure \ref{fig5.8}. Similarly, we observe that the fairs of the Lena image are better recovered in \ref{fig5.9:subfig:d} and \ref{fig5.9:subfig:f} than in \ref{fig5.9:subfig:b}, \ref{fig5.9:subfig:c} and \ref{fig5.9:subfig:e}.

Finally, we quantify the denoising performance of different algorithms for the remote images in Figure \ref{fig5.1}. Table \ref{tab5.3} lists the PSNR values, the number of iterations and the CPU time. The parameter setting of the PLAD algorithms is the same as that adopted for the experiments in Table \ref{tab5.2}. For the spatially-adapted total variation model, we set $\delta = 10, 15, 20$ for Remote1 and Remote2 with $M=5, 8, 10$ respectively, and fix $\delta = 10$ for the Nimes image. The other parameter setting refers to those used for the experiments in Table \ref{tab5.2}. For Algorithm 2, $\tau^{k}$ is updated by the Newton method every three iteration.

\begin{figure}
  \centering
  \subfigure[]{
    \label{fig5.10:subfig:a} 
    \includegraphics[width=1.5in,clip]{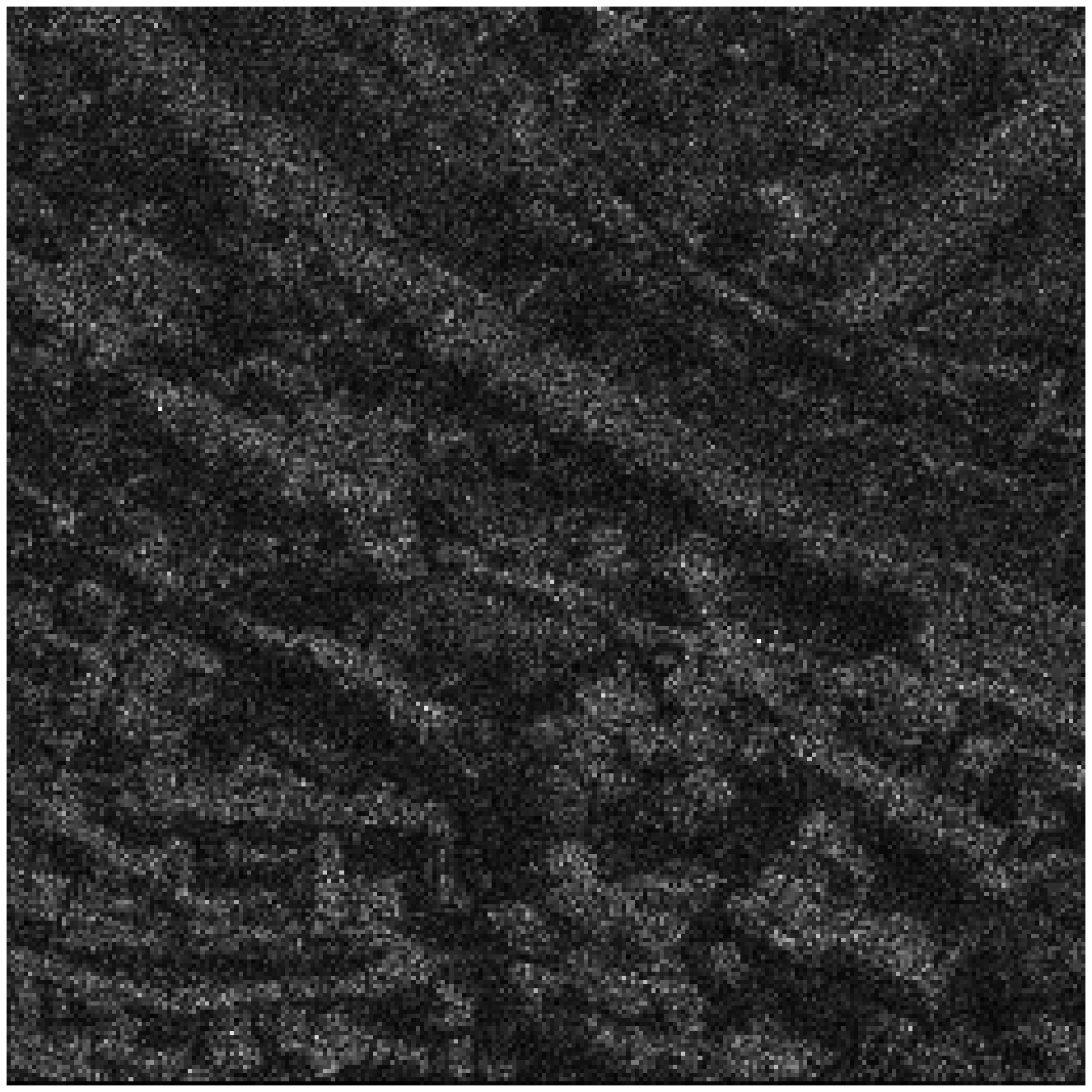}}
  \subfigure[]{
    \label{fig5.10:subfig:b} 
    \includegraphics[width=1.5in,clip]{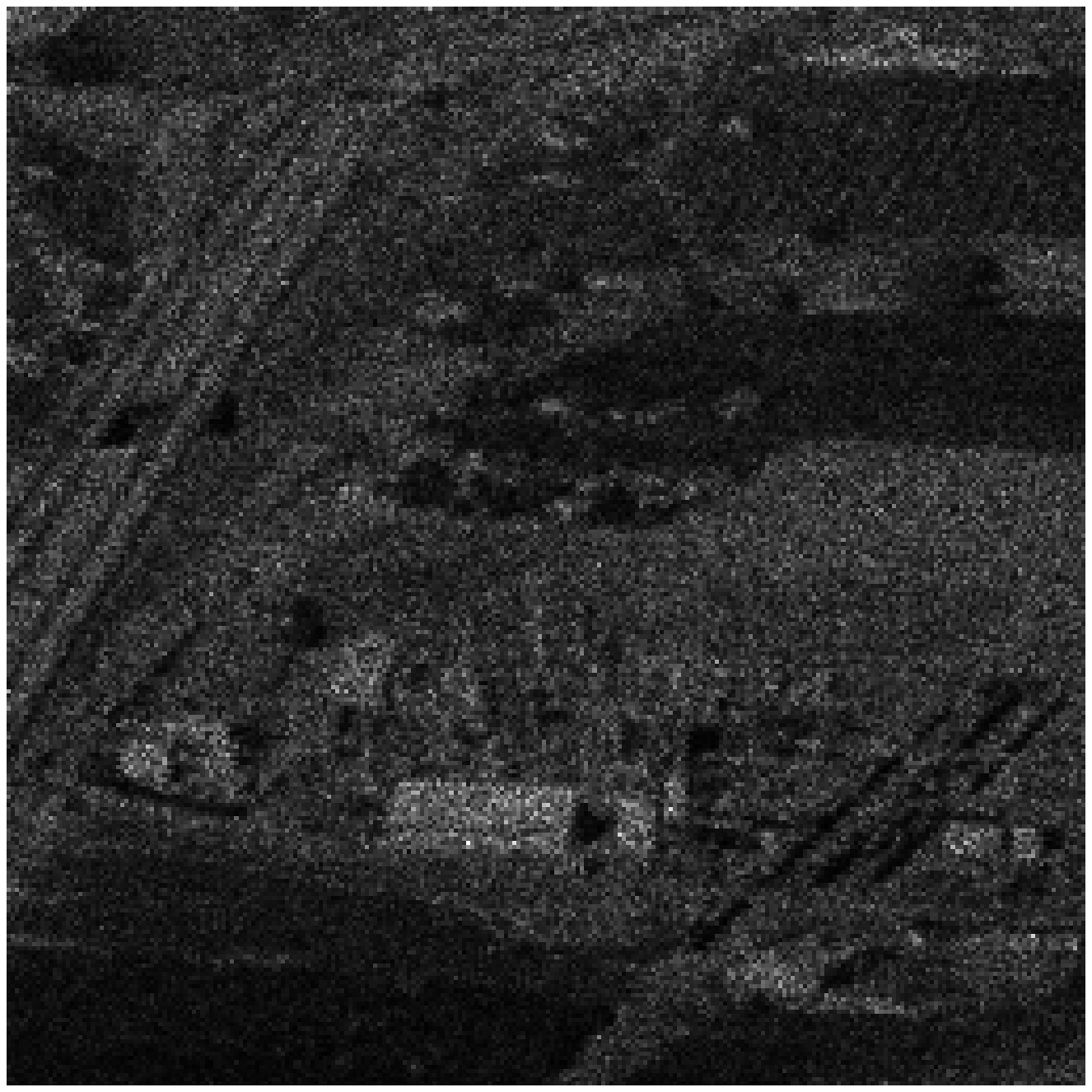}}
  \subfigure[]{
    \label{fig5.10:subfig:c} 
    \includegraphics[width=1.5in,clip]{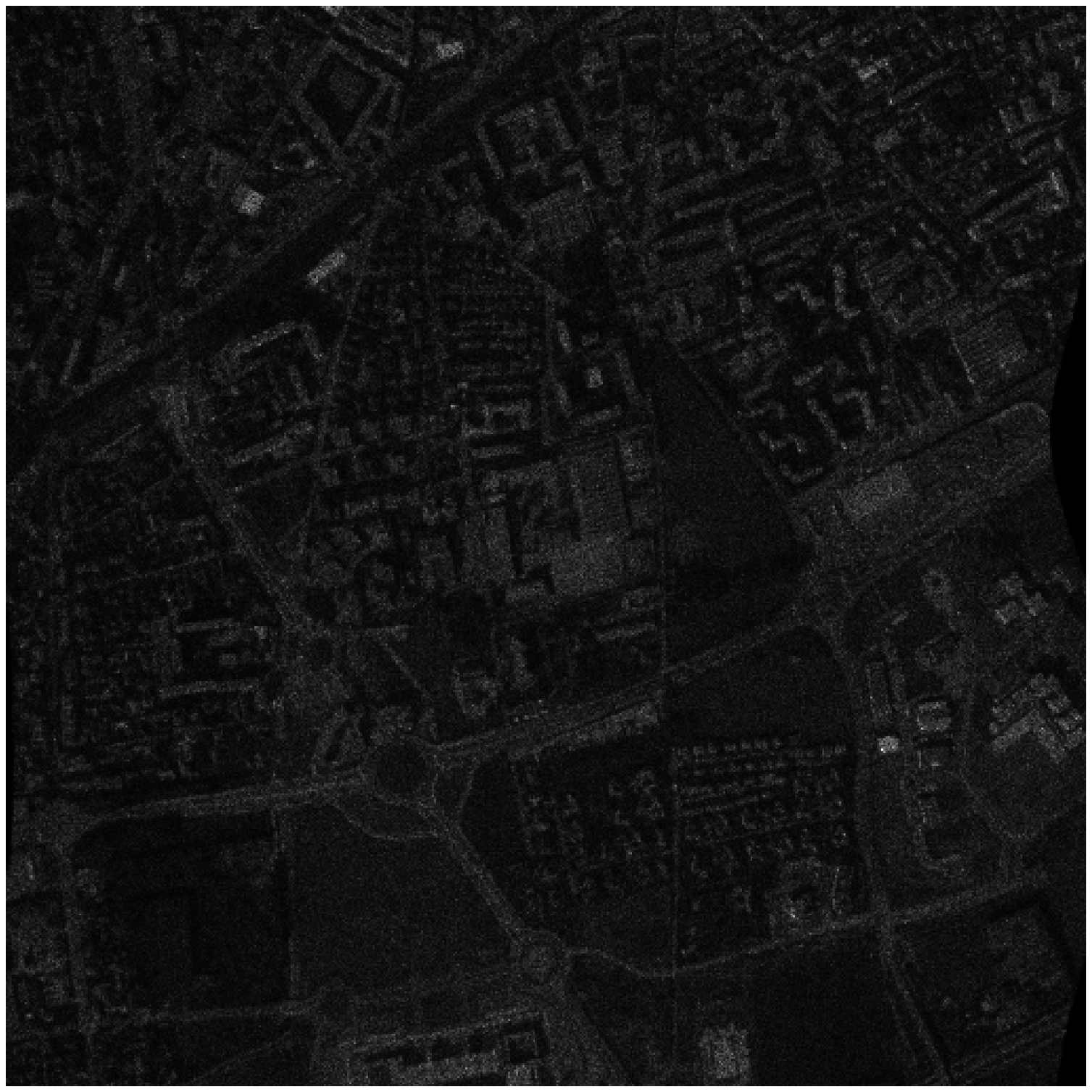}}
\caption{(a)The Remote1 image with M=5, (b)the Remote2 image with M=8, (c)the Nimes image with M=10.
}
\label{fig5.10}
\end{figure}

\begin{figure}
  \centering
  \subfigure[]{
    \label{fig5.11:subfig:a} 
    \includegraphics[width=1.5in,clip]{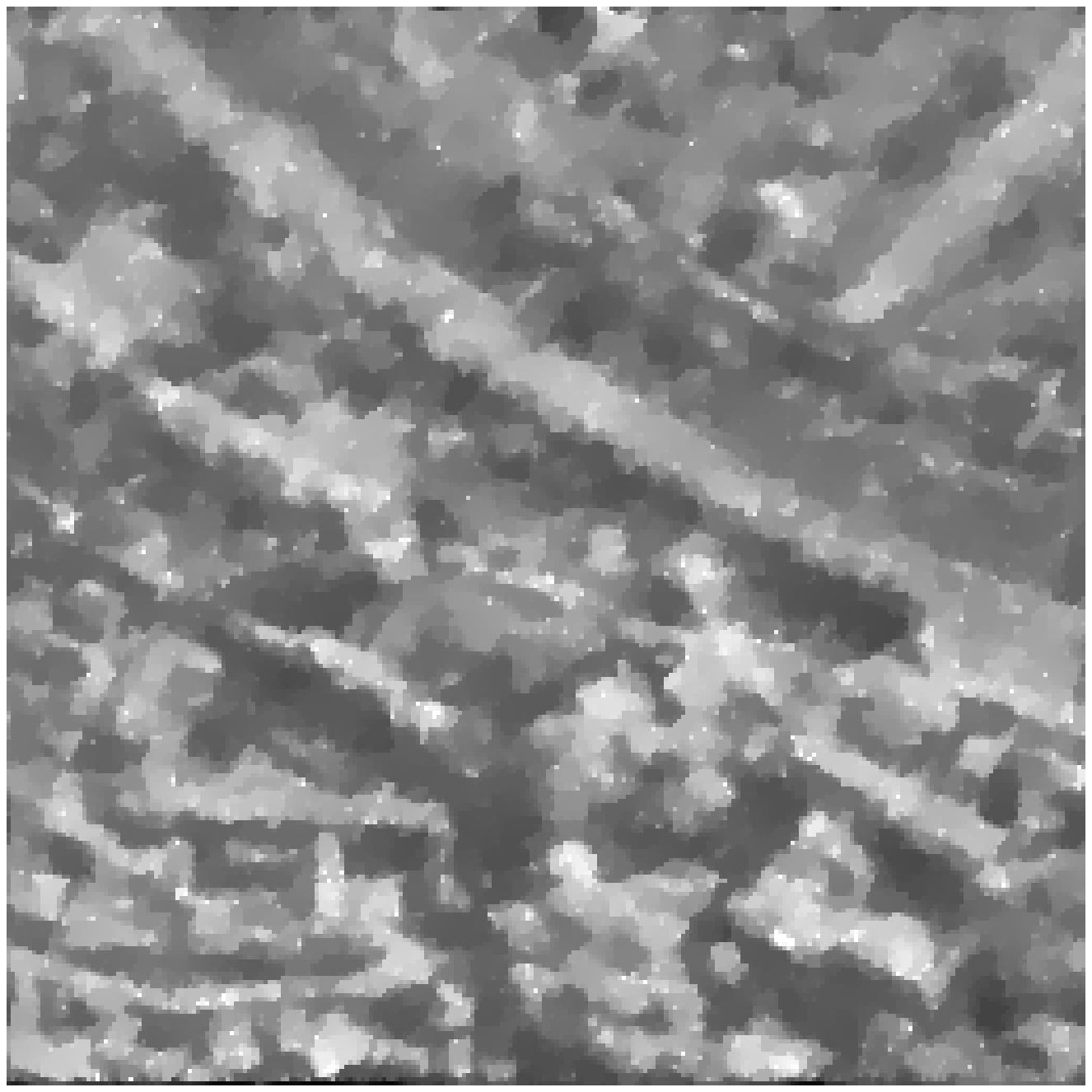}}
  \subfigure[]{
    \label{fig5.11:subfig:b} 
    \includegraphics[width=1.5in,clip]{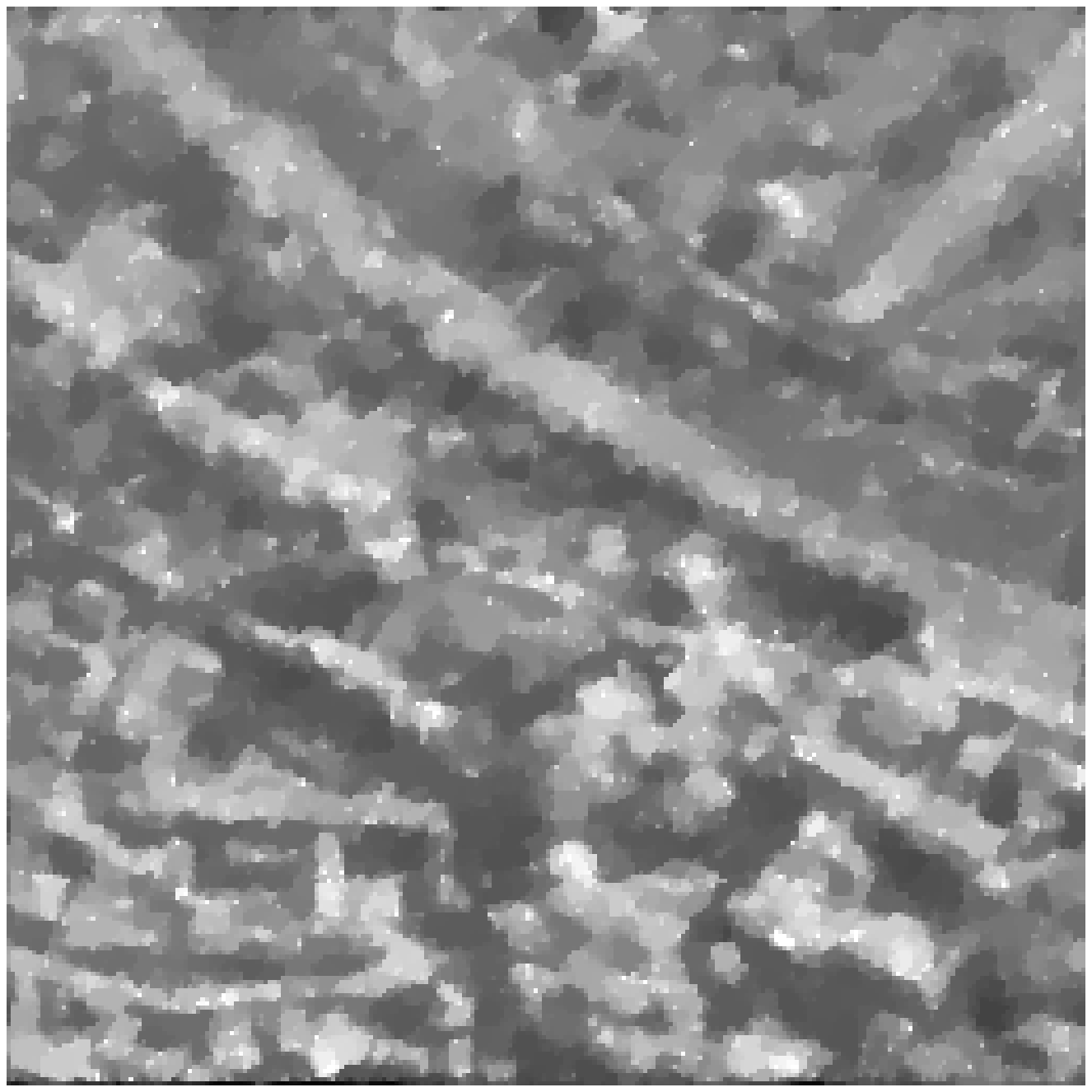}}
  \subfigure[]{
    \label{fig5.11:subfig:c} 
    \includegraphics[width=1.5in,clip]{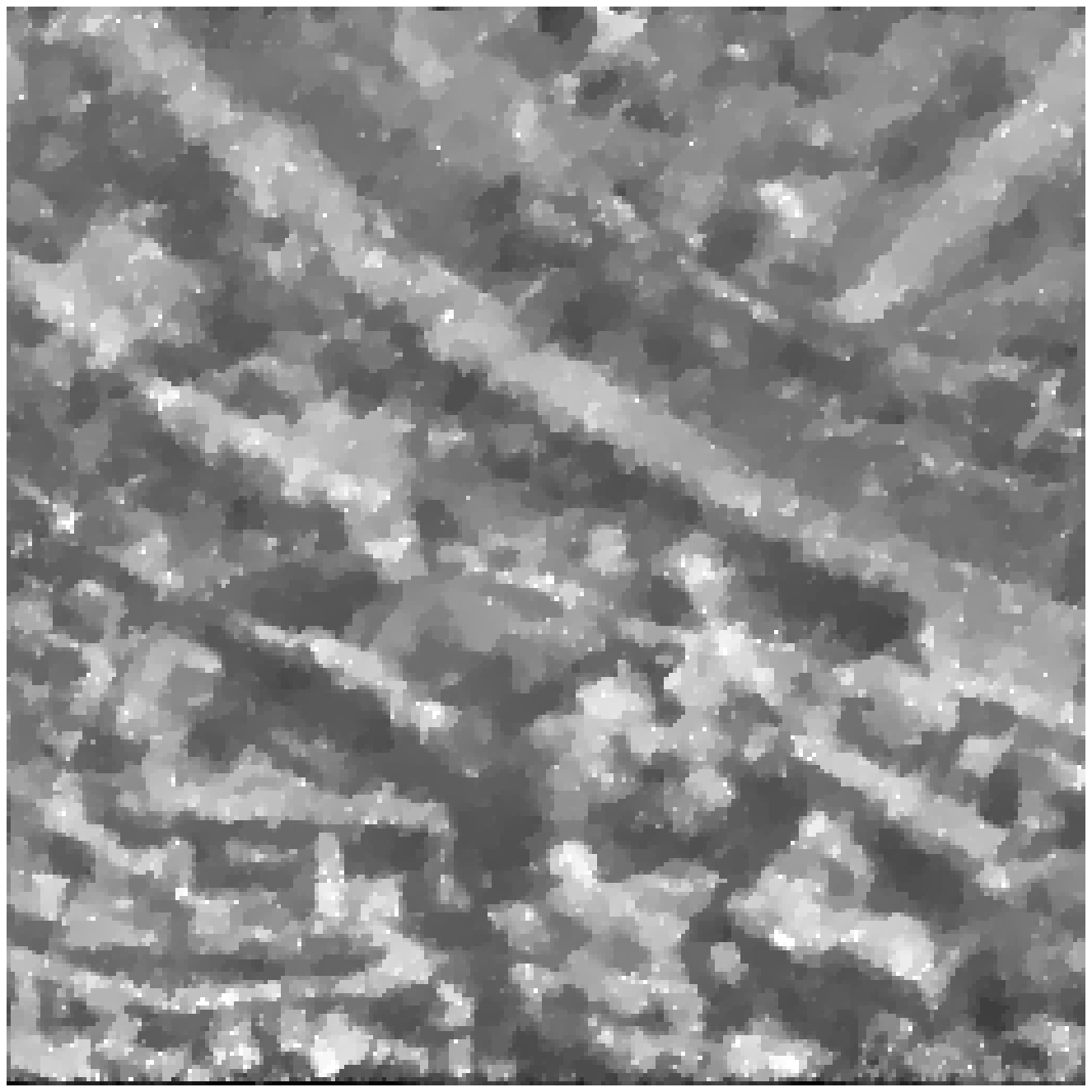}}
  \subfigure[]{
    \label{fig5.11:subfig:d} 
    \includegraphics[width=1.5in,clip]{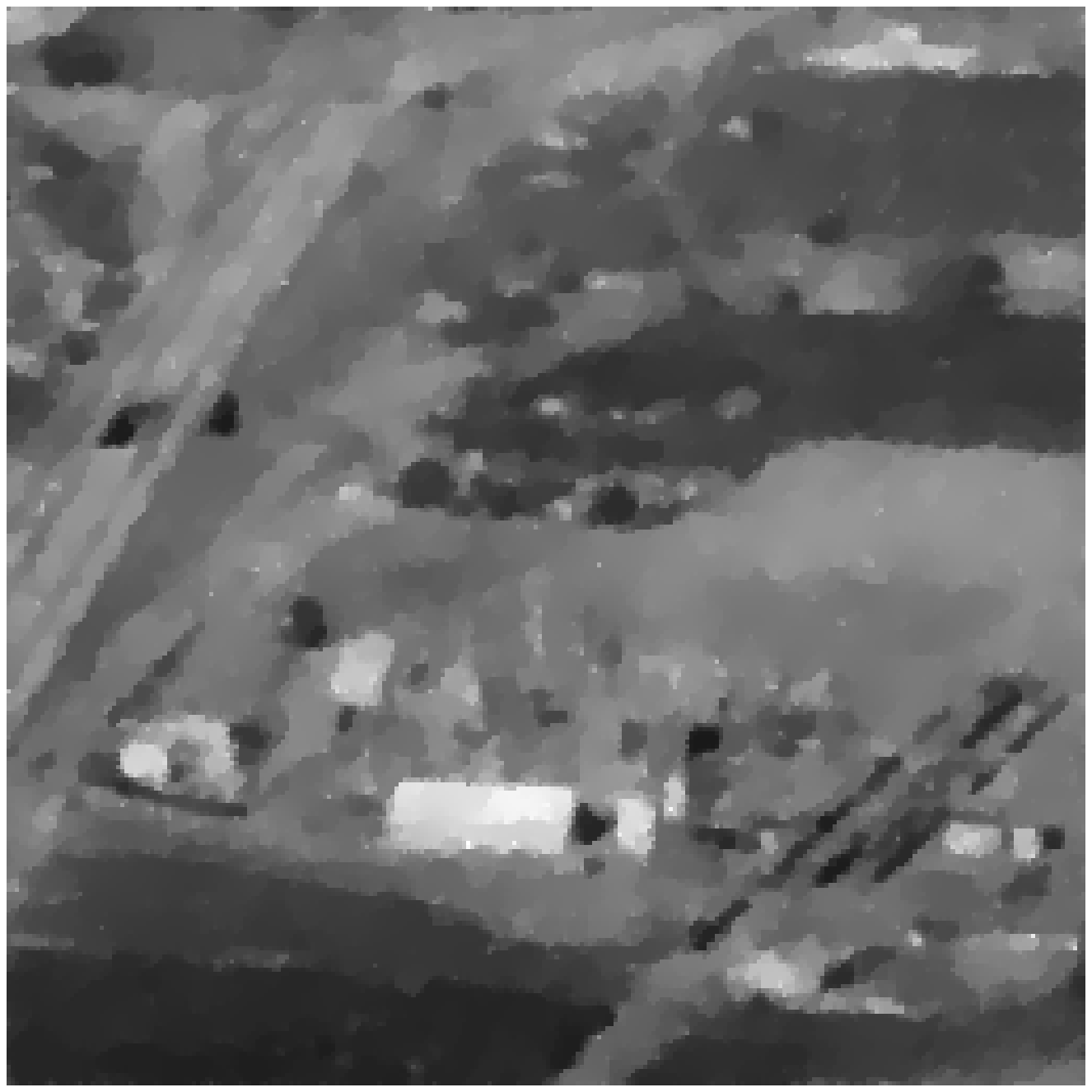}}
  \subfigure[]{
    \label{fig5.11:subfig:e} 
    \includegraphics[width=1.5in,clip]{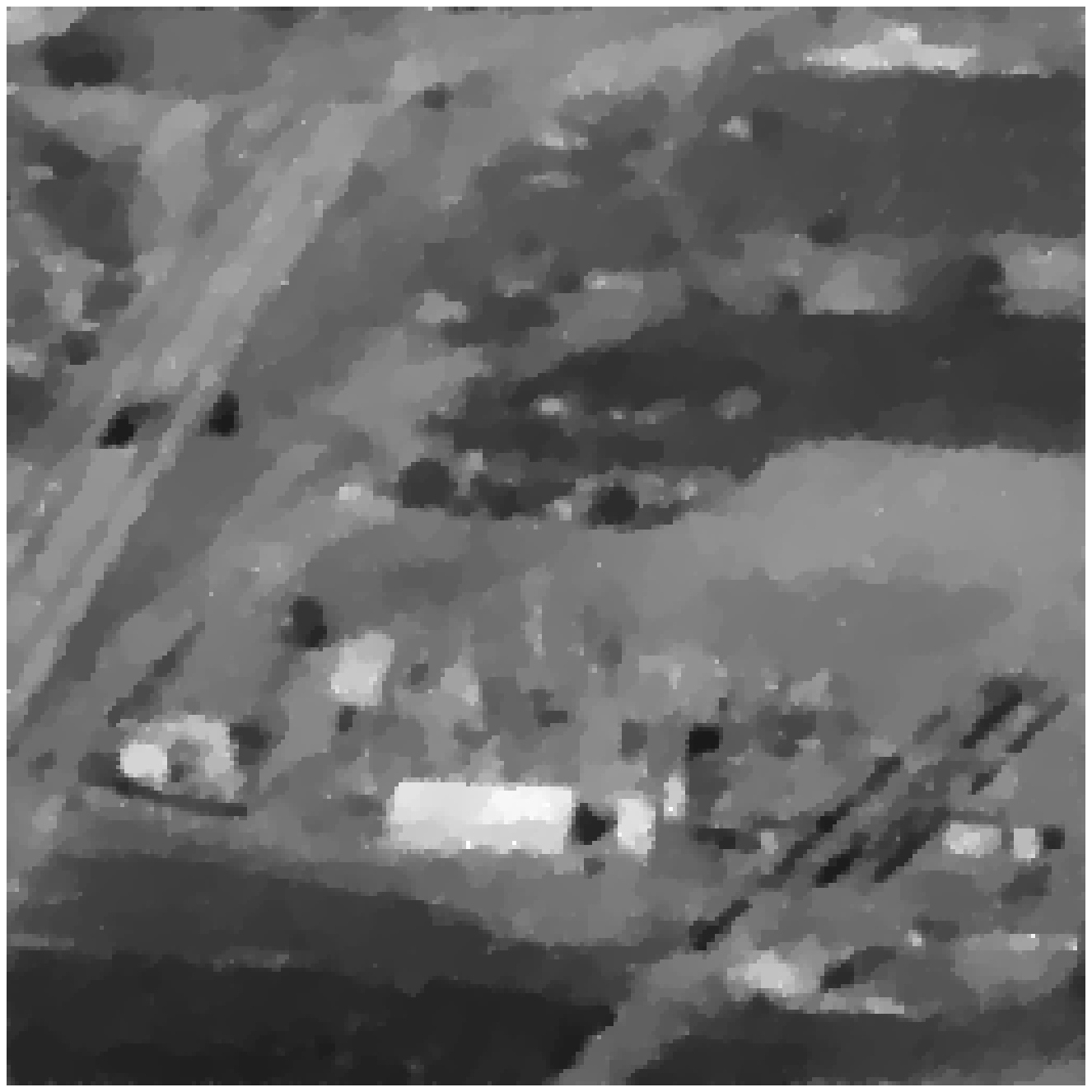}}
  \subfigure[]{
    \label{fig5.11:subfig:f} 
    \includegraphics[width=1.5in,clip]{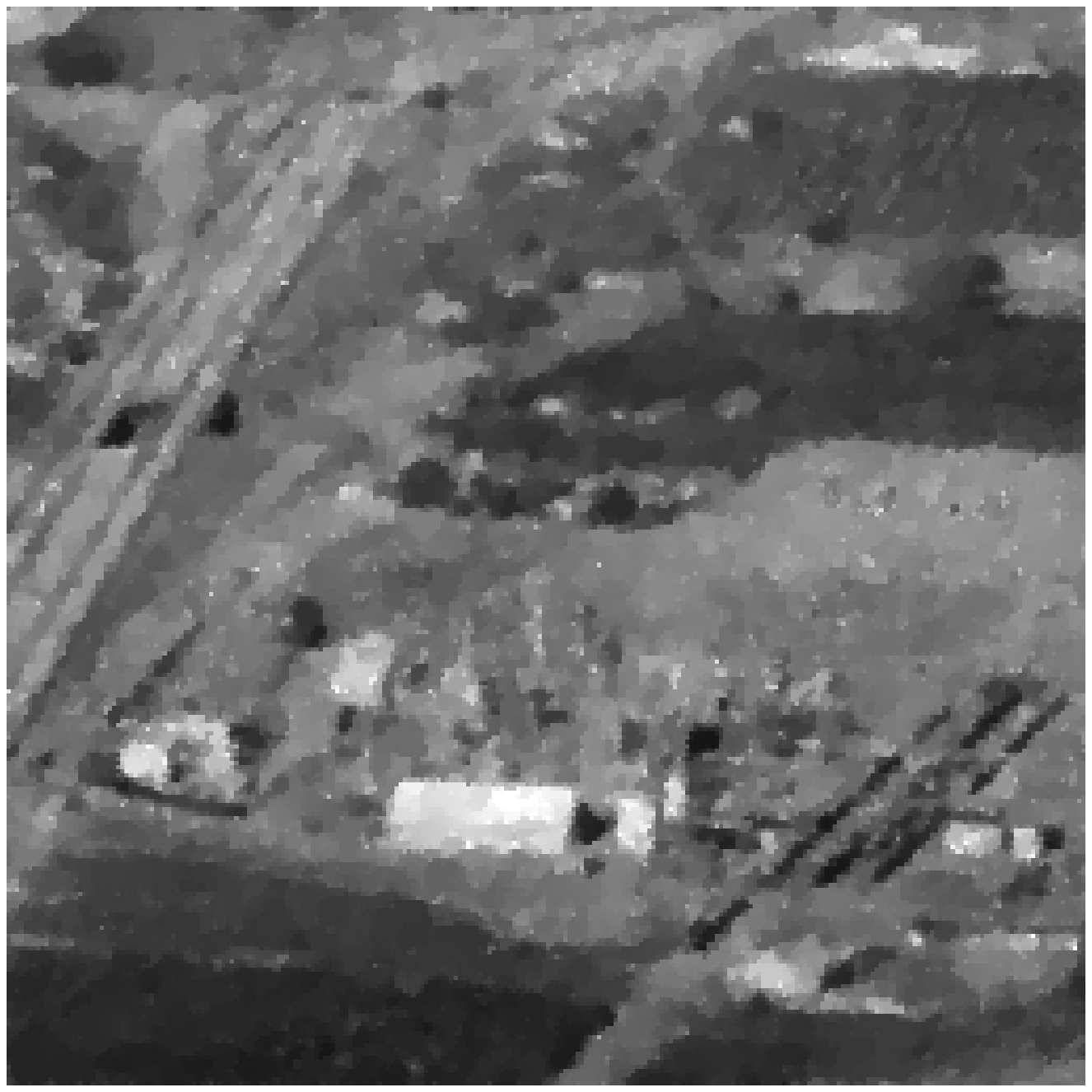}}
  \subfigure[]{
    \label{fig5.11:subfig:g} 
    \includegraphics[width=1.5in,clip]{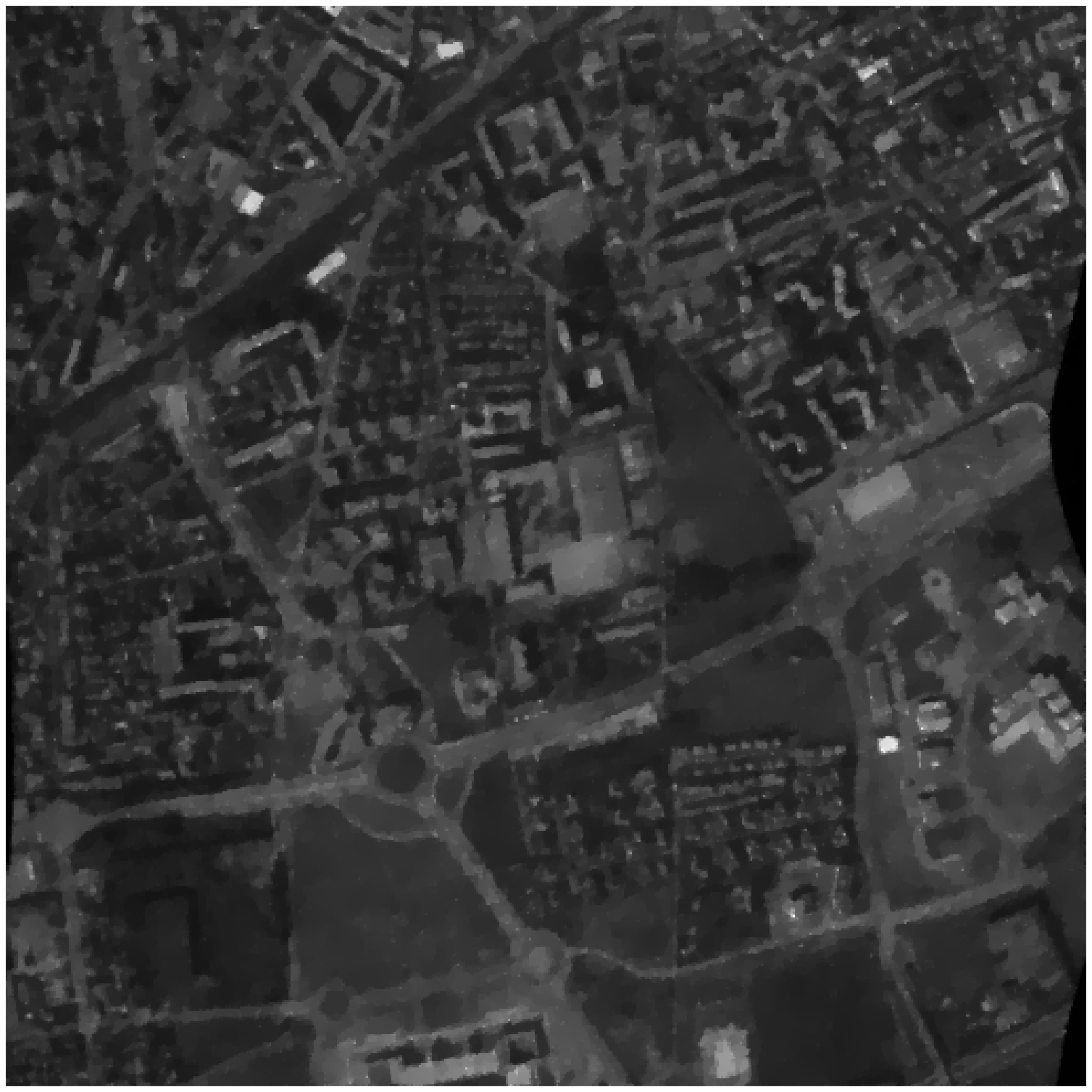}}
  \subfigure[]{
    \label{fig5.11:subfig:h} 
    \includegraphics[width=1.5in,clip]{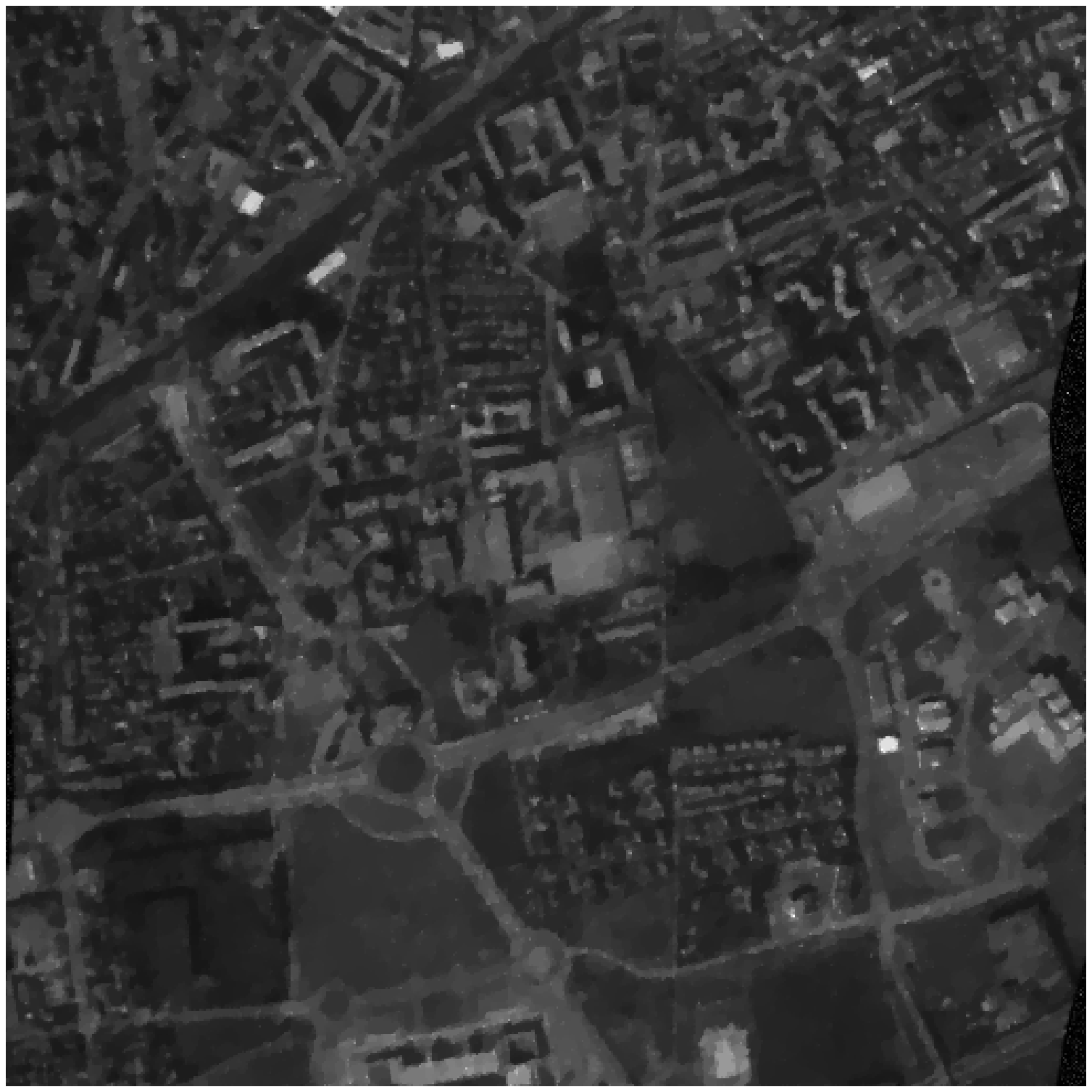}}
  \subfigure[]{
    \label{fig5.11:subfig:i} 
    \includegraphics[width=1.5in,clip]{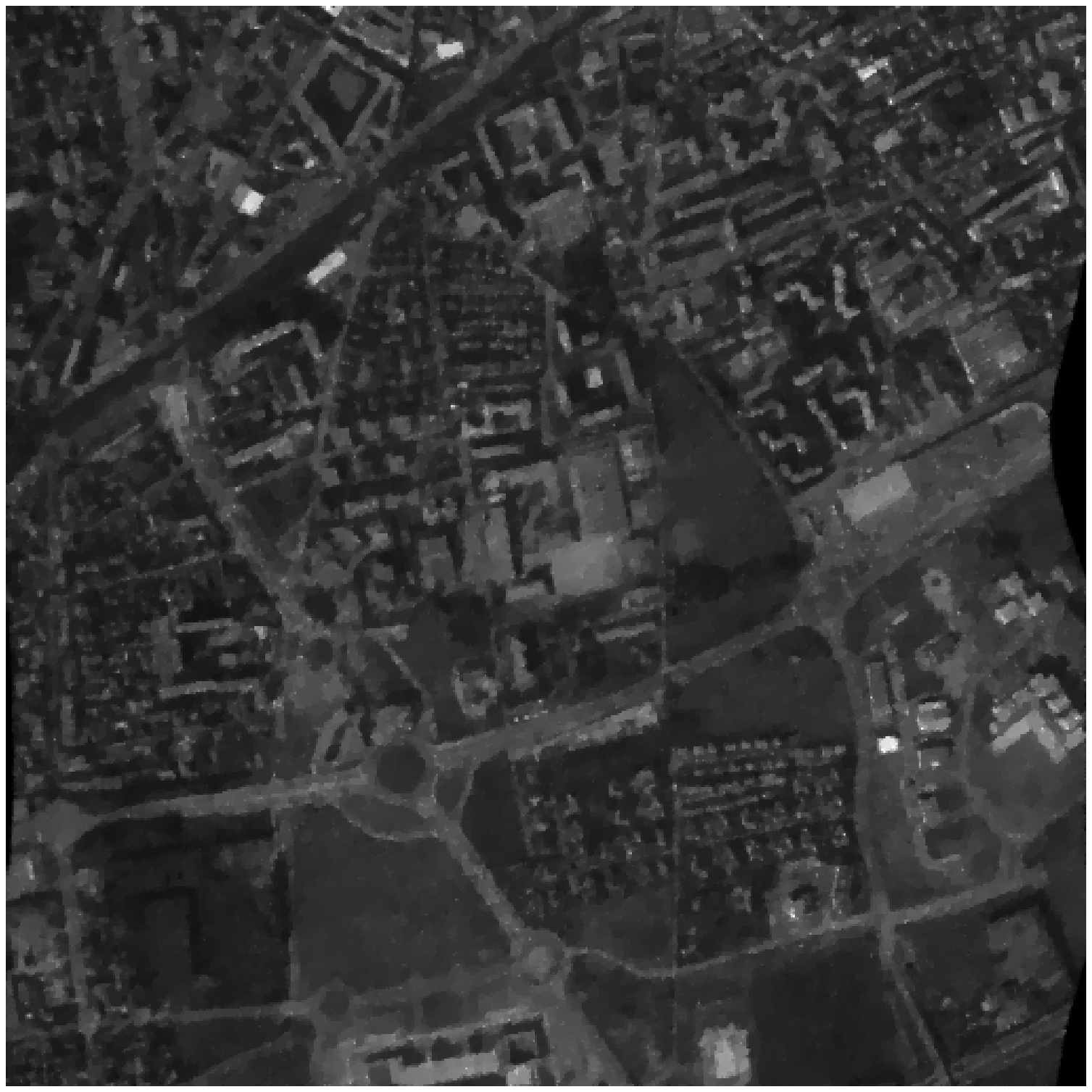}}
\caption{The first column shows the restored results of the PLAD\_EXP; the second column shows the restored results of the  PLAD\_DIV; the third column shows the restored results of Algorithm2.
}
\label{fig5.11}
\end{figure}

Similarly to the experiment results in Table \ref{tab5.2}, we observe that Algorithm 2 obtains the best effect while comparing both the PSNR values and CPU time. Figure \ref{fig5.10} and \ref{fig5.11} shows the noisy image and the restored results of PLAD algorithms and our algorithm respectively. Due to the use of the spatially varying parameter, our algorithm is able to improve the quality of the restored images compared with the PLAD algorithms. Meanwhile, since the parameter can be easily computed by the Newton method during the iteration, it takes much less time than the STV\_AL method.


\begin{table} [htbp]
\centering \caption{The comparison of different methods: the given numbers are PSNR (dB)/Iteration number/CPU time(second) }
\scalebox{0.9}{
\begin{tabular}{|c|c|c|c|c|c|c|}
  \hline
    Image & M & PLAD\_EXP \cite{SIAMJSC:Linearized} & PLAD\_DIV \cite{SIAMJSC:Linearized} & STV\_AL \cite{IEEETIP:STV} & Algorithm 1 & Algorithm 2 \\
  \hline
    & 5 & 20.75/50/4.59 & 20.76/75/4.44 & 20.47/3/28.36 & 20.75/50/5.47 & \textbf{20.81}/56/8.92 \\
  \cline{2-7}
    Remote1 & 8 & 21.17/51/4.59 & 21.23/70/4.13 & 21.56/3/26.17 & 21.62/48/4.95 & \textbf{21.78}/54/8.56 \\
  \cline{2-7}
    & 10 & 21.94/43/3.91 & 21.99/69/4.05 & 22.00/3/22.03 & 21.95/35/3.77 & \textbf{22.21}/55/8.79  \\
  \hline
    & 5 & 23.20/52/5.02 & 23.21/69/3.95 & 23.22/3/26.63 & 23.20/52/5.34 & \textbf{23.27}/32/5.55 \\
  \cline{2-7}
    Remote2 & 8 & 23.72/52/4.88 & 23.76/65/3.78 & 24.20/3/23.01 & 24.18/49/5.18 & \textbf{24.30}/57/8.85 \\
  \cline{2-7}
    & 10 & 24.46/43/3.86 & 24.50/63/3.63 & 24.58/3/22.11 & 24.55/48/4.98 & \textbf{24.79}/58/8.91 \\
  \hline
    & 5 & 25.93/48/26.95 & 25.45/110/47.55 & 26.13/3/89.13 & 25.95/47/27.80 & \textbf{26.13}/57/47.45 \\
  \cline{2-7}
    Nimes & 8 & 26.43/48/24.94 & 26.40/110/43.56 & \textbf{27.31}/3/76.85 & 26.97/45/27.50 & 27.21/51/40.25 \\
  \cline{2-7}
    & 10 & 27.19/39/24.94 & 27.14/105/41.43 & \textbf{27.90}/3/69.27 & 27.32/44/25.94 & 27.62/51/41.50 \\
  \hline
\end{tabular}}
\label{tab5.3}
\end{table}

\section{Conclusion}

In this paper, we propose two fast linearized alternating direction minimization algorithms that simultaneously estimate the regularization parameter and recover the image contaminated by Gamma noise. The new approaches are base on the statistical characteristics of some random variable with respect to the Gamma noise. By utilizing the linearized technique and the (local) discrepancy principle, we establish nonlinear equation(s) with respect to the regularization parameter. Then fast iterative algorithms, which update regularization parameter through computing the solution of the established equation(s), are proposed to remove the multiplicative noise. Numerical experiments demonstrate that the proposed methods are able to obtain a suitable value for the regularization parameter, and overall outperform those of the current state-of-the-art methods while considering both the PSNR values and CPU time.

\section*{acknowledgement}
The work was supported in part by the National Natural Science Foundation of China under Grant 61271014.

\end{document}